\documentclass{article} %
\usepackage{iclr2026_conference,times}
\iclrfinalcopy

\usepackage{amsmath,amsfonts,bm}

\def\eqref#1{equation~\ref{#1}}

\def\1{\bm{1}}

\DeclareMathAlphabet{\mathsfit}{\encodingdefault}{\sfdefault}{m}{sl}
\SetMathAlphabet{\mathsfit}{bold}{\encodingdefault}{\sfdefault}{bx}{n}

\usepackage{tikz}
\usepackage{pgfplots}
\pgfplotsset{compat=1.18}

\usepackage[table]{xcolor}
\usepackage{hyperref}
\usepackage{url}
\usepackage{graphicx}
\usepackage{subcaption}
\usepackage{wrapfig}
\usepackage{adjustbox}

\usepackage{amsmath}
\usepackage{array}
\usepackage{booktabs}
\usepackage{tabularx}
\usepackage{longtable}
\usepackage{multirow}
\usepackage{placeins}
\usepackage{float}
\usepackage{listings}

\definecolor{PromptBackground}{HTML}{F7F8FA}
\definecolor{PromptTitleBackground}{HTML}{D9DCE1}
\definecolor{PromptBorder}{HTML}{D9DCE1}

\lstdefinestyle{promptstyle}{
    basicstyle=\ttfamily\scriptsize,
    backgroundcolor=\color{PromptBackground},
    frame=single,
    rulecolor=\color{PromptBorder},
    framerule=0.3pt,
    breaklines=true,
    breakatwhitespace=false,
    breakindent=0pt,
    breakautoindent=false,
    columns=fullflexible,
    keepspaces=true,
    showstringspaces=false,
    tabsize=2,
    xleftmargin=6pt,
    xrightmargin=6pt,
    framexleftmargin=2.7pt,
    framexrightmargin=2.7pt,
    aboveskip=0pt,
    belowskip=6pt
}

\newcommand{\prompttitle}[1]{%
    \par\smallskip
    \noindent
    \colorbox{PromptTitleBackground}{%
        \parbox{\dimexpr\linewidth-2\fboxsep\relax}{%
            \strut\bfseries\small #1\strut
        }%
    }%
    \par\vspace{-1pt}\nopagebreak[4]
}

\hypersetup{hidelinks}

\usepackage{tcolorbox}
\usepackage{fontawesome5}
\usepackage{xcolor}

\newtcbox{\orangebutton}{on line,
  colback=orange,
  colframe=orange,
  coltext=white,
  arc=3pt,
  boxrule=0pt,
  left=5pt,right=5pt,top=2.5pt,bottom=2.5pt}

\newtcbox{\blackbutton}{on line,
  colback=black,
  colframe=black,
  coltext=white,
  arc=3pt,
  boxrule=0pt,
  left=5pt,right=5pt,top=2.5pt,bottom=2.5pt}

\title{Code2Games: Enabling Coding Agents for Gaming World Generation}

\author{Wei Wu$^{1*}$~~
Ziyang Xu$^{1*}$~~
Zeyu Zhang$^{1*\dag}$~~
Yang Zhao$^{2}$~~
Hao Tang$^{1\ddag}$\\[0.5em]
$^1$School of Computer Science, Peking University~~
$^2$La Trobe University\\[0.4em]
\small $^*$Equal contribution.
$^\dag$Project lead.
$^\ddag$Corresponding author: bjdxtanghao@gmail.com.
}

\begin{document}

\maketitle

\begin{center}
\vspace{-2em}
\href{https://aigeeksgroup.github.io/Code2Games}{
  \orangebutton{\faGlobe\ \textbf{Website}}
}
\hspace{8pt}
\href{https://github.com/AIGeeksGroup/Code2Games}{
  \blackbutton{\faGithub\ ~~~\textbf{Code}}
}
\end{center}

\begin{figure*}[!h]
    \centering
    \includegraphics[width=\textwidth]{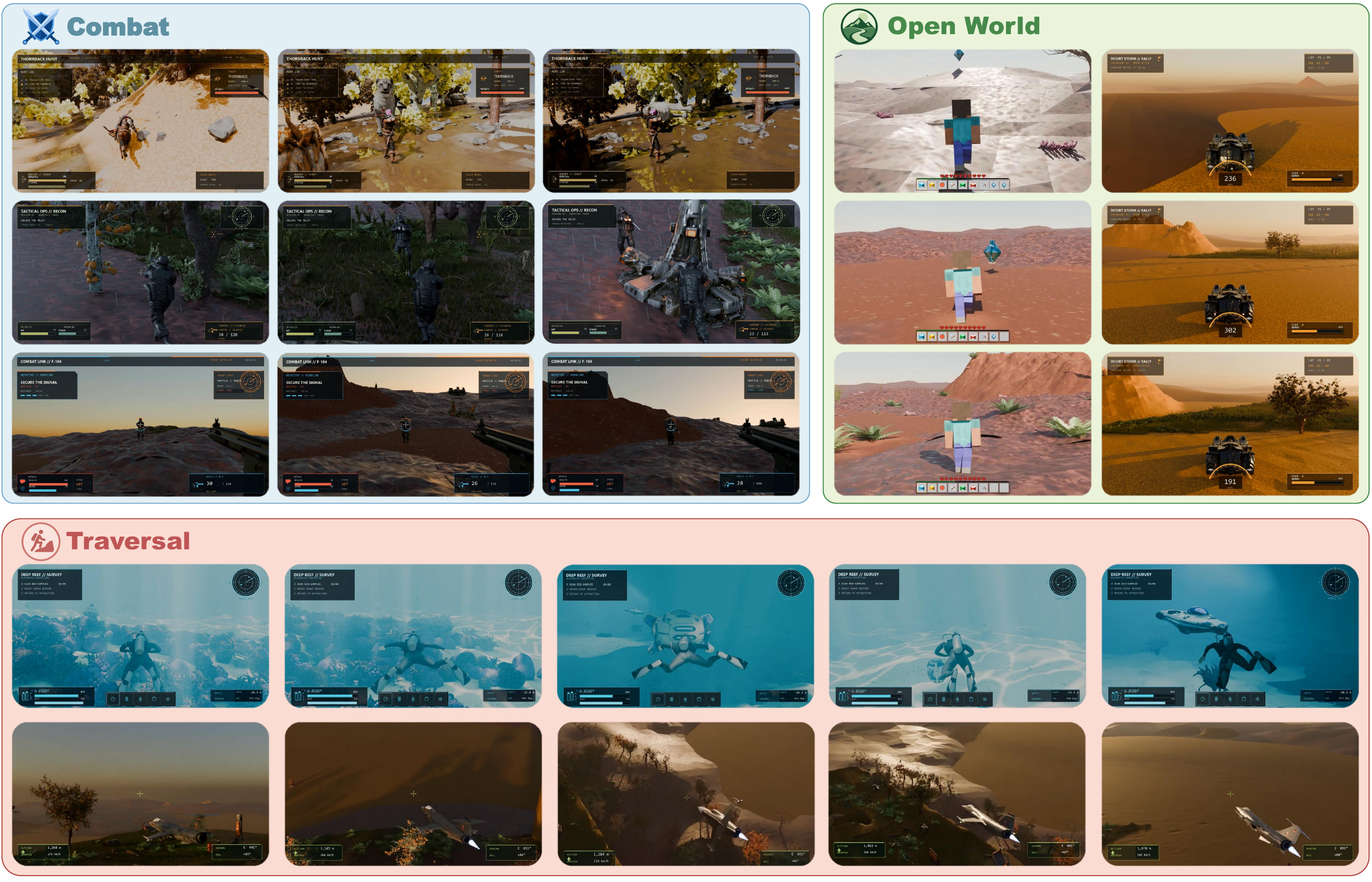}
    \caption{Overview of the gaming world scenarios generated by Code2Games.}
    \label{fig:gameplay_grid}
\end{figure*}
\begin{abstract}
  Generating a high-quality gaming world from a natural-language game intent requires joint reasoning about scene structure, spatial layout, gameplay objectives, interactive entities, and executable gameplay logic. Existing coding agents can generate individual assets, scenes, or scripts, but often struggle to maintain consistency across these components. We propose \textbf{Code2Games}, an agentic framework that builds a structured gaming world upon a base Blender world generated from the same game intent. Code2Games coordinates scene analysis, gameplay planning, constrained gaming-world generation, and gaming-engine customization through a shared scene--gameplay representation with persistent element correspondence. After world generation, Code2Games adapts the generated world to Unreal Engine 5 and employs an execution-guided reconstruction process that uses compilation diagnostics, runtime feedback, and gameplay test results to resolve inconsistencies arising during engine adaptation. To systematically evaluate gaming-world generation, we introduce the \textbf{GameCode4D benchmark}, which comprises ten fixed game prompts spanning different levels of scene and gameplay complexity. We evaluate the generated results across four dimensions: visual quality, interactive fidelity, multimodal artifact quality, and playable-game quality. Experiments demonstrate that, compared with direct gaming-world generation by coding agents and existing baseline methods, Code2Games consistently improves the visual quality and interactive fidelity of generated gaming worlds, as well as the quality of the resulting games after engine adaptation.
\end{abstract}

\section{Introduction}

\begin{wrapfigure}[20]{r}{0.52\columnwidth}
\vspace{-0.8em}
\centering
\begin{tikzpicture}
\begin{axis}[
width=0.50\columnwidth,
height=0.36\columnwidth,
xmode=log,
ymode=log,
log basis x=10,
log basis y=10,
xmin=15,
xmax=130,
ymin=5,
ymax=110,
xtick={20,50,100},
xticklabels={20,50,100},
ytick={5,10,20,50,100},
yticklabels={5,10,20,50,100},
xlabel={Scene Entity Count},
ylabel={Executable Behavior Count},
title style={font=\scriptsize,yshift=-1pt},
xlabel style={font=\scriptsize},
ylabel style={font=\scriptsize},
tick label style={font=\tiny},
grid=major,
grid style={dotted,gray!45},
axis line style={black!75},
clip=false
]

\addplot[
forget plot,
no marks,
densely dashed,
gray!48,
line width=0.25pt
] coordinates {(18,5) (18,8)};

\addplot[
forget plot,
no marks,
densely dashed,
gray!48,
line width=0.25pt
] coordinates {(15,8) (18,8)};

\addplot[
forget plot,
no marks,
densely dashed,
gray!48,
line width=0.25pt
] coordinates {(28,5) (28,12)};

\addplot[
forget plot,
no marks,
densely dashed,
gray!48,
line width=0.25pt
] coordinates {(15,12) (28,12)};

\addplot[
forget plot,
no marks,
densely dashed,
gray!48,
line width=0.25pt
] coordinates {(68,5) (68,17)};

\addplot[
forget plot,
no marks,
densely dashed,
gray!48,
line width=0.25pt
] coordinates {(15,17) (68,17)};

\addplot[
forget plot,
no marks,
densely dashed,
gray!48,
line width=0.25pt
] coordinates {(90,5) (90,23)};

\addplot[
forget plot,
no marks,
densely dashed,
gray!48,
line width=0.25pt
] coordinates {(15,23) (90,23)};

\addplot[
forget plot,
no marks,
densely dashed,
gray!48,
line width=0.25pt
] coordinates {(52,5) (52,35)};

\addplot[
forget plot,
no marks,
densely dashed,
gray!48,
line width=0.25pt
] coordinates {(15,35) (52,35)};

\addplot[
forget plot,
no marks,
densely dashed,
gray!48,
line width=0.25pt
] coordinates {(48,5) (48,60)};

\addplot[
forget plot,
no marks,
densely dashed,
gray!48,
line width=0.25pt
] coordinates {(15,60) (48,60)};

\addplot[
forget plot,
no marks,
densely dashed,
gray!48,
line width=0.25pt
] coordinates {(100,5) (100,48)};

\addplot[
forget plot,
no marks,
densely dashed,
gray!48,
line width=0.25pt
] coordinates {(15,48) (100,48)};

\addplot[
forget plot,
no marks,
densely dashed,
gray!48,
line width=0.25pt
] coordinates {(78,5) (78,72)};

\addplot[
forget plot,
no marks,
densely dashed,
gray!48,
line width=0.25pt
] coordinates {(15,72) (78,72)};

\addplot[
forget plot,
no marks,
densely dashed,
gray!48,
line width=0.25pt
] coordinates {(70,5) (70,92)};

\addplot[
forget plot,
no marks,
densely dashed,
gray!48,
line width=0.25pt
] coordinates {(15,92) (70,92)};

\addplot[
forget plot,
no marks,
densely dashed,
gray!48,
line width=0.25pt
] coordinates {(118,5) (118,64)};

\addplot[
forget plot,
no marks,
densely dashed,
gray!48,
line width=0.25pt
] coordinates {(15,64) (118,64)};

\addplot[
only marks,
mark=*,
mark size=2.4pt,
draw=blue!70!black,
fill=blue!80
] coordinates {(18,8)};

\addplot[
only marks,
mark=*,
mark size=2.6pt,
draw=blue!70!black,
fill=blue!55
] coordinates {(28,12)};

\addplot[
only marks,
mark=*,
mark size=3.0pt,
draw=blue!70!black,
fill=cyan!65
] coordinates {(68,17)};

\addplot[
only marks,
mark=*,
mark size=3.4pt,
draw=green!50!black,
fill=green!60
] coordinates {(90,23)};

\addplot[
only marks,
mark=*,
mark size=3.5pt,
draw=red!70!black,
fill=red!72
] coordinates {(52,35)};

\addplot[
only marks,
mark=*,
mark size=3.8pt,
draw=red!70!black,
fill=red!88
] coordinates {(48,60)};

\addplot[
only marks,
mark=*,
mark size=3.8pt,
draw=cyan!60!black,
fill=cyan!60
] coordinates {(100,48)};

\addplot[
only marks,
mark=*,
mark size=4.1pt,
draw=orange!70!black,
fill=orange!82
] coordinates {(78,72)};

\addplot[
only marks,
mark=*,
mark size=4.7pt,
draw=red!70!black,
fill=red!86
] coordinates {(70,92)};

\addplot[
only marks,
mark=*,
mark size=4.5pt,
draw=orange!70!black,
fill=orange!78
] coordinates {(118,64)};

\node[
font=\fontsize{4.5}{5}\selectfont,
anchor=west,
xshift=4pt,
fill=white,
fill opacity=1,
text opacity=1,
inner sep=.6pt
] at (axis cs:18,8) {Temple Run};

\node[
font=\fontsize{4.5}{5}\selectfont,
anchor=west,
xshift=4pt,
fill=white,
fill opacity=1,
text opacity=1,
inner sep=.6pt
] at (axis cs:28,12) {Skiing};

\node[
font=\fontsize{4.5}{5}\selectfont,
anchor=south east,
xshift=-2pt,
yshift=2pt,
fill=white,
fill opacity=1,
text opacity=1,
inner sep=.6pt
] at (axis cs:68,17) {Wingsuit};

\node[
font=\fontsize{4.5}{5}\selectfont,
anchor=west,
xshift=4pt,
yshift=0pt,
fill=white,
fill opacity=1,
text opacity=1,
inner sep=.6pt
] at (axis cs:90,23) {Racing};

\node[
font=\fontsize{4.5}{5}\selectfont,
anchor=east,
xshift=-4pt,
yshift=3pt,
fill=white,
fill opacity=1,
text opacity=1,
inner sep=.6pt
] at (axis cs:52,35) {TPS};

\node[
font=\fontsize{4.5}{5}\selectfont,
anchor=west,
xshift=4pt,
yshift=-4pt,
fill=white,
fill opacity=1,
text opacity=1,
inner sep=.6pt
] at (axis cs:48,60) {FPS};

\node[
font=\fontsize{4.5}{5}\selectfont,
anchor=west,
xshift=-10pt,
yshift=-6pt,
fill=white,
fill opacity=1,
text opacity=1,
inner sep=.6pt
] at (axis cs:100,48) {Underwater};

\node[
font=\fontsize{4.2}{4.8}\selectfont,
anchor=south east,
xshift=15pt,
yshift=-10pt,
fill=white,
fill opacity=1,
text opacity=1,
inner sep=.6pt
] at (axis cs:78,72) {\shortstack{Assassin's Creed}};

\node[
font=\fontsize{4.5}{5}\selectfont,
anchor=south east,
xshift=-4pt,
yshift=1pt,
fill=white,
fill opacity=1,
text opacity=1,
inner sep=.6pt
] at (axis cs:70,92) {Monster Hunt};

\node[
font=\fontsize{4.5}{5}\selectfont,
anchor=north east,
xshift=6pt,
yshift=10pt,
fill=white,
fill opacity=1,
text opacity=1,
inner sep=.6pt
] at (axis cs:118,64) {Minecraft};

\node[
anchor=north west,
draw=gray!45,
fill=white,
fill opacity=.97,
text opacity=1,
rounded corners=1pt,
inner sep=2pt,
font=\tiny,
align=left
] at (rel axis cs:0.025,0.975) {
\textbf{Game Category}\\[-1pt]
\textcolor{red!85!black}{\textbullet}\ Combat
\quad
\textcolor{orange!90!black}{\textbullet}\ Exploration\\[-1pt]
\textcolor{blue!80!black}{\textbullet}\ Traversal
\quad
\textcolor{green!55!black}{\textbullet}\ Racing
};

\node[
anchor=north west,
draw=gray!45,
fill=white,
fill opacity=.97,
text opacity=1,
rounded corners=1pt,
inner sep=2pt,
font=\tiny,
align=left
] at (rel axis cs:0.025,0.725) {
\textbf{Gameplay Elements}\\[-2pt]
\raisebox{-0.5\height}{\textcolor{gray}{\small\textbullet}}\ Fewer
\quad
\raisebox{-0.5\height}{\textcolor{gray}{\Large\textbullet}}\ More
};

\end{axis}
\end{tikzpicture}

\caption{
Complexity of gaming worlds generated by Code2Games.
The horizontal axis reports the number of scene entities,
whereas the vertical axis reports the number of executable behaviors;
both axes use logarithmic scales.
Bubble size denotes the number of gameplay elements,
and color indicates the game category.
}

\label{fig:code2games-complexity}
\vspace{-0.7em}
\end{wrapfigure}
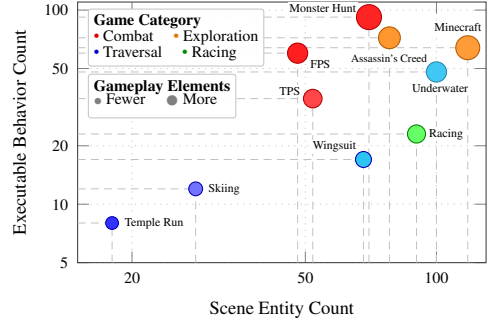

Generating a coherent gaming world from a high level design intent is a challenging problem in interactive content creation and agentic software generation. A gaming world is not merely a collection of visually plausible assets. It must combine a structured environment, spatially consistent objects, interactive entities, gameplay objectives, and executable rules that jointly reflect the intended experience. The central challenge is to maintain the correspondence between the game intent, the generated gaming world, and the interactions that users are expected to perform within it.

Recent advances in large language models (LLMs) and coding agents have enabled progress in procedural generation of 3D gaming assets, game playing, and automated game development~\cite{wang2023voyager,yao2022react,shinn2023reflexion,jimenez2024swe}. Existing systems can generate gaming assets, static scenes, gameplay scripts, or engine specific project components. However, these components are often generated and evaluated independently. An object may be visually plausible but placed in an unsuitable region, a gameplay role may not correspond to the correct entity, and an objective may be grounded in an inaccessible part of the scene. Consequently, individually reasonable components can still form a gaming world that is inconsistent with the design intent.

The difficulty of gaming world generation varies substantially across scenarios because it depends on both scene structure and gameplay logic. Scene complexity increases with the number and diversity of objects, spatial relations, regions, and navigable layouts, while gameplay complexity increases with the number of objectives, interactive elements, state transitions, and event dependencies. As shown in Figure~\ref{fig:code2games-complexity}, our scenarios cover a broad range of complexity, from relatively simple traversal and skiing settings to more complex combat, exploration, and open world environments. The high complexity of these scenarios presents a significant challenge for an agent to jointly preserve the visual organization and interactive semantics of the generated world.

To address this challenge, we propose Code2Games, an agentic framework for generating gaming worlds from a natural language game intent and a corresponding Blender base scene generated from the same intent. Code2Games first generates the gaming world through three specialized agents. The Scene Analysis Agent extracts spatial regions, objects, accessibility relations, and interaction affordances from the base scene. The Gameplay Planning Agent translates the natural language intent into a structured specification of objectives, actors, gameplay elements, and interaction rules. Based on these outputs, the Gaming World Generation Agent instantiates scene elements and gameplay components while respecting the extracted spatial and gameplay constraints. These stages are coordinated through a shared scene gameplay representation that maintains persistent correspondence between the elements of the intended world.

After the gaming world has been generated, Code2Games further adapts it to a target gaming engine through the Gaming Engine Customization Agent. This agent reconstructs the engine side actors, configuration files, and gameplay logic while preserving the element correspondence established during world generation. Since some inconsistencies only become visible after engine adaptation and execution, Code2Games introduces an execution guided reconstruction process at this stage. The adapted world is compiled, launched, and evaluated through gameplay tests, and compiler diagnostics, runtime feedback, and gameplay outcomes are used to localize and repair errors in scene realization, actor binding, configuration files, and gameplay rules. In our implementation, Unreal Engine 5 serves as the target gaming engine and the execution and validation backend.

To systematically evaluate gaming world generation, we introduce GameCode4D benchmark, a benchmark comprising ten fixed game prompts. Each prompt specifies the game genre, viewpoint, and high level objective. The benchmark covers diverse combinations of scene structural complexity and gameplay logic complexity. We evaluate the generated gaming worlds along four complementary dimensions: visual quality, interactive fidelity, multimodal artifact quality, and playable game quality. These dimensions assess the appearance of the generated world, its consistency with the specified setting and interactions, the quality of the generated artifacts, and the extent to which the intended gameplay experience is realized after engine adaptation and execution.

Our contributions are summarized as follows:

\begin{itemize}
    \item We propose Code2Games, an agentic framework that coordinates scene analysis, gameplay planning, gaming world generation, and gaming engine customization for transforming a natural language game intent and a base Blender scene into an interactive gaming world.
    \item We introduce a shared scene gameplay representation with persistent element correspondence, together with an execution guided reconstruction process that adapts generated worlds to Unreal Engine 5 and repairs inconsistencies revealed during engine adaptation and runtime execution.
    \item We introduce GameCode4D benchmark, a benchmark of ten gaming world scenarios covering diverse combinations of scene structural complexity and gameplay logic complexity, together with an evaluation protocol for scene consistency, objective grounding, interaction correctness, intent alignment, and runtime executability.
\end{itemize}

\section{Related Work}

\paragraph{Interactive video and game world models.}
Early work learned action-conditioned visual dynamics from gameplay footage
\citep{menapace2021playable,menapace2022playable}.
Recent world models improve temporal coherence, controllability, and visual fidelity.
Genie learns action-controllable environments from unlabeled videos, while DIAMOND and GameNGen model interactive game dynamics
\citep{bruce2024genie,alonso2024diamond,valevski2025gamengen}.
Later systems extend this formulation to open worlds, unseen game domains, and instruction-conditioned interaction
\citep{che2024gamegenx,yu2025gamefactory,lu2025genex,
huang2026vid2world,li2025hunyuan, tang2025hunyuan}.
These methods generate interactive observations rather than editable engine projects with explicit assets, object identities, and executable rules.
Their evaluation has accordingly expanded from perceptual video quality to physical plausibility, instruction following, temporal consistency, and action responsiveness
\citep{huang2024vbench,zheng2025vbench2,li2025hunyuan,
duan2025worldscore,ying2026wbench}.
Code2Games addresses a different stage of generation: it turns a structured world into a runnable game project while retaining the scene information needed to implement and verify gameplay.

\paragraph{Language-guided 3D world generation.}
Procedural generation and text-to-3D methods provide scalable routes to embodied environments and textured scenes
\citep{deitke2022procthor,hollein2023text2room}.
Large language and vision-language models further support compositional planning by predicting layouts, spatial relations, and scene specifications
\citep{feng2023layoutgpt,lin2023layoutprompter,yang2024holodeck}.
Other systems improve complex-scene generation through layout-conditioned synthesis, hierarchical generation, or optimization of object arrangements
\citep{zhou2024gala3d,zhang2024scenewiz3d,zhang2024scenedreamer,
yang2024scenecraft,wang2024architect,bokhovkin2025scenefactor,
sun2025layoutvlm}.
RoboGen links generated environments to task construction in simulation, while Code2Worlds uses coding LLMs to generate 4D worlds
\citep{wang2024robogen,zhang2026code2worlds}.
These approaches establish the scene as a reusable representation, but they stop at geometry, appearance, or simulation configuration.
Code2Games uses the resulting scene as the spatial substrate for gameplay: objectives, actors, elements, and interactions are grounded to scene regions, and persistent identifiers connect Blender objects and annotations to Unreal actors, trigger volumes, and executable logic.

\paragraph{Agentic game development.}
LLM-based game-development systems have progressed from instruction-driven code generation to multi-agent workflows for design, implementation, and debugging
\citep{wu2024idge,hong2025chatge,li2025cardiverse,zhang2026rpgagent}.
V-GameGym studies visual game generation by code LLMs, while AutoUE and OpenGame construct games through agentic coding and engine interaction
\citep{zhang2026v,yin2026autoue,jiang2026opengame}.
End-to-end playable-game benchmarks emphasize coherence across design, code, and runtime behavior
\citep{luo2026gamecraft}.
Code2Games differs in how the game-building process is conditioned: the agent receives both a game intent and the Code2Worlds base world generated from that intent.
It derives a game specification, places gameplay elements subject to accessibility and geometric constraints, and uses compiler, runtime, and gameplay feedback to repair the affected components.
The persistent correspondence between world representation and engine implementation makes these repairs local rather than requiring the project to be regenerated as an undifferentiated code artifact.

\section{Method}
Given a natural language game intent $u$ and a base Blender scene $W_B$ generated by Code2Worlds, Code2Games generates a runnable Unreal Engine 5 project $\mathcal{U}^{*}$. The generated project realizes the requested gameplay while respecting the spatial structure of $W_B$ and preserving the correspondence between gameplay elements, their scene realizations, Unreal representations, and executable logic.

Directly generating $\mathcal{U}^{*}$ with coding LLM is difficult because scene constraints, gameplay semantics, and engine code are represented in different forms. Errors in element placement or object binding can propagate across stages and lead to compilation failures or incorrect gameplay behavior.

\begin{figure*}[t]
  \centering
  \includegraphics[width=\textwidth]{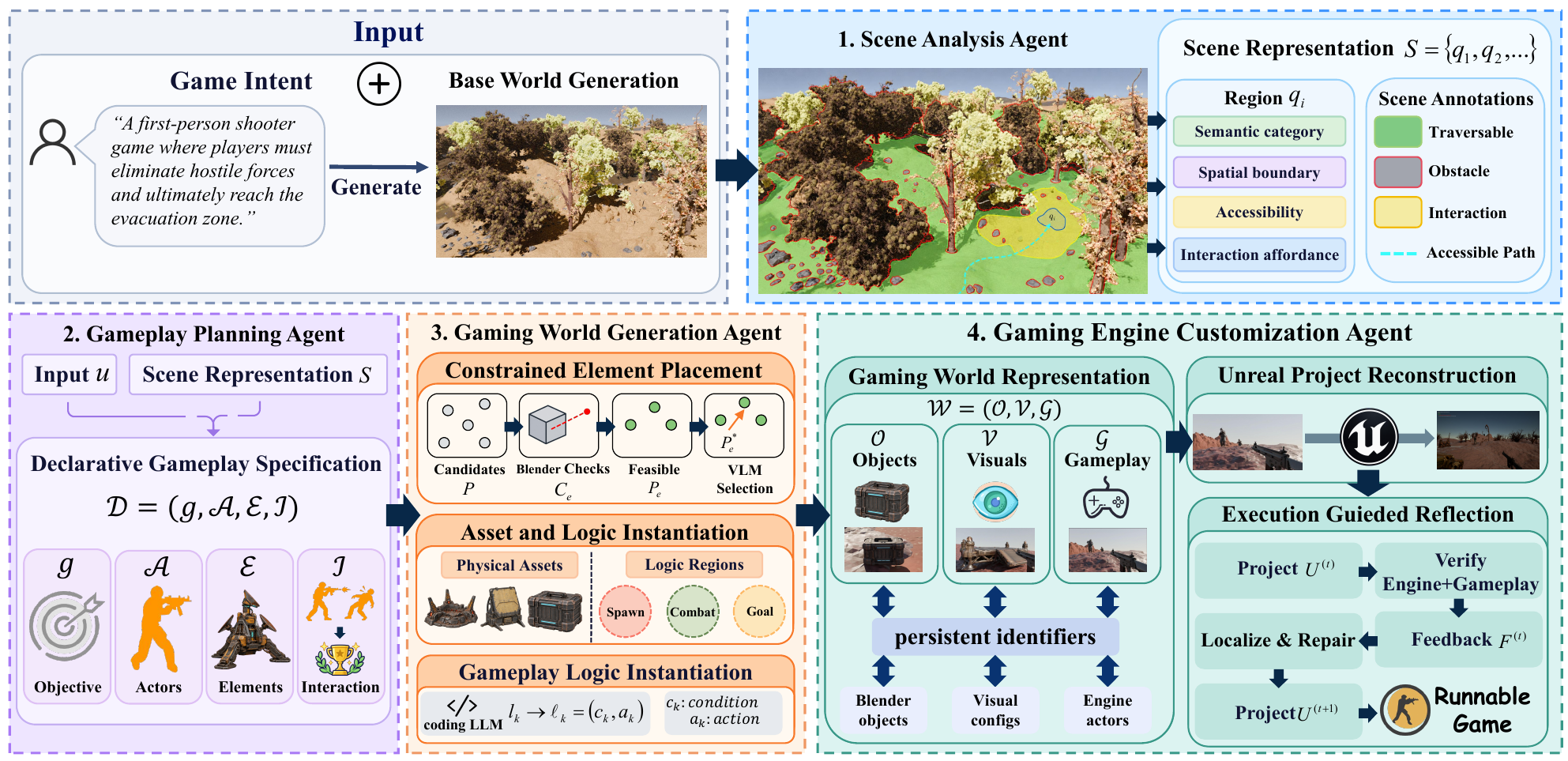}
  \caption{Overview of Code2Games, which transforms a game intent and its paired base world into a gaming world and adapts it to a game engine.}
  \label{fig:overview}
\end{figure*}

Code2Games addresses this problem with a shared scene and gameplay representation and an iterative repair procedure driven by execution feedback. The representation encodes scene constraints, gameplay requirements, and persistent element identifiers. Based on this representation, the system realizes gameplay elements in the Blender scene, reconstructs the corresponding Unreal project, and uses engine diagnostics and gameplay tests to revise failed components. These operations are implemented by four specialized agents, as illustrated in Figure ~\ref{fig:overview}.

\subsection{Scene and Gameplay Representation}
\label{sec:representation}

The scene and gameplay representation serves as a shared intermediate representation for scene analysis, gameplay planning, and game realization. It describes the spatial constraints of the input scene, the requirements of the target gameplay, and the correspondence between each gameplay element and its realizations across the generation stages.

\paragraph{Scene Constraints.} Given a base Blender scene $W_B$, the Scene Analysis Agent constructs a scene constraint representation
\begin{equation}
\mathcal{S}=\{q_j\}_{j=1}^{M}.
\end{equation}
Each region $q_j$ is represented as
\begin{equation}
q_j=(\tau_j,b_j,\rho_j,\alpha_j),
\end{equation}
where $\tau_j$ denotes the semantic category of the region, $b_j$ denotes its spatial boundary, $\rho_j$ describes its accessibility according to the scene navigation mesh, and $\alpha_j$ is the set of interaction affordances supported by the region. The semantic category determines which types of gameplay elements are compatible with the region, the spatial boundary constrains the locations at which these elements can be placed, the accessibility descriptor supports reachability checks, and the affordance set constrains the interaction types that can be assigned to the region. An affordance describes a capability offered by a scene region, such as walking, placement, or event triggering, whereas an interaction denotes a gameplay rule involving one or more actors or gameplay elements and a resulting game state update. These descriptors provide spatial constraints for gameplay planning and game realization.

\paragraph{Gameplay Planning.} Given the user intent $u$ and the scene representation $\mathcal{S}$, the Gameplay Planning Agent produces a declarative gameplay specification
\begin{equation}
\mathcal{D}=(g,\mathcal{A},\mathcal{E},\mathcal{I}).
\end{equation}
Here, $g$ defines the game objective and its success and failure conditions, $\mathcal{A}$ contains the actors, $\mathcal{E}$ contains the required gameplay elements, and $\mathcal{I}$ contains the interactions among actors, gameplay elements, and the environment. Each element $e\in\mathcal{E}$ is associated with a persistent identifier, a functional description, an element type $\kappa_e$, and a placement constraint $C_e$. The type $\kappa_e$ distinguishes physical game elements from logical gameplay regions. Each interaction $\iota_k\in\mathcal{I}$ is represented as
\begin{equation}
\iota_k=(P_k,t_k,c_k,\Delta s_k),
\end{equation}
where $P_k\subseteq\mathcal{A}\cup\mathcal{E}$ denotes the participating actors and gameplay elements, $t_k$ is the triggering event, $c_k$ is the precondition, and $\Delta s_k$ specifies the resulting game state update. For example, an interaction may specify that entering a goal region after all hostile actors have been eliminated updates the game state to success. The specification describes the required gameplay while leaving concrete locations, assets, and engine implementations to the realization stage.

\paragraph{Persistent Correspondence.}To preserve the identity of gameplay element throughout the generation process, every element $e\in\mathcal{E}$ is assigned a persistent identifier and an correspondence record
\begin{equation}
\mu(e)=
\left(
\operatorname{id}_e,\kappa_e,\beta_B(e),\beta_U(e),\mathcal{I}_e,\mathcal{L}_e
\right),
\end{equation}
where $\operatorname{id}_e$ is the persistent identifier, $\kappa_e$ is the element type, $\beta_B(e)$ and $\beta_U(e)$ denote its Blender and Unreal realizations, $\mathcal{I}_e$ is the set of interactions involving $e$, and $\mathcal{L}_e$ is the set of executable logic units implementing these interactions. For a physical game element, $\beta_B(e)$ is a Blender scene object and $\beta_U(e)$ is an Unreal Actor. For a logical gameplay region, $\beta_B(e)$ is a scene region annotation and $\beta_U(e)$ is an Unreal Trigger Volume. Given the participant set $P_k$ of an interaction $\iota_k$, the associated interaction and logic sets are defined as
\begin{equation}
\mathcal{I}_e=
\{\iota_k\in\mathcal{I}\mid e\in P_k\},
\qquad
\mathcal{L}_e=
\{\ell_k\mid \iota_k\in\mathcal{I}_e\},
\end{equation}
where $\ell_k$ denotes the executable logic unit generated from interaction $\iota_k$. This correspondence connects each declarative gameplay element to its scene realization, engine representation, interactions, and executable logic. It therefore allows placement changes, asset replacements, and logic repairs to be propagated to the affected components without regenerating the entire project.

\subsection{Game Realization with Scene Constraints}
\label{sec:realization}

Given the scene representation $\mathcal{S}$ and the gameplay specification $\mathcal{D}$, this stage maps each abstract gameplay element to a concrete realization in the Blender scene. The realization consists of three steps: generating feasible locations, selecting a location that best supports the intended gameplay function, and instantiating the corresponding assets and executable logic.

\paragraph{Feasible Placement.}
For each gameplay element $e\in\mathcal{E}$, the system first identifies scene regions that are compatible with the element type and required affordances. Let $\mathcal{J}_e$ denote the set of compatible regions:
\begin{equation}
\mathcal{J}_e=
\left\{
j\mid
\operatorname{Compat}(\tau_j,e)=1
\land
\operatorname{Afford}(\alpha_j,e)=1
\right\}.
\end{equation}
Here, $\operatorname{Compat}(\tau_j,e)$ indicates whether the semantic category of region $q_j$ is compatible with gameplay element $e$, while $\operatorname{Afford}(\alpha_j,e)$ indicates whether the interaction affordances of region $q_j$ support the placement and use of $e$.

Candidate locations are then sampled from the boundaries of these regions:
\begin{equation}
\mathcal{P}^{(0)}_e=
\operatorname{Sample}
\left(
\left\{
b_j\mid j\in\mathcal{J}_e
\right\}
\right).
\end{equation}
The candidate set is filtered using the placement constraint $C_e$ and geometric checks:
\begin{equation}
\mathcal{P}_e=
\left\{
p\in\mathcal{P}^{(0)}_e
\middle|
C_e(p)=1,\ 
\operatorname{Support}(p)=1,\ 
\operatorname{Collision}(p)=0,\ 
\operatorname{Reachable}(p)=1
\right\}.
\end{equation}
Here, $\operatorname{Support}(p)$ verifies that the candidate is placed on valid surface, $\operatorname{Collision}(p)$ checks intersection with existing objects, and $\operatorname{Reachable}(p)$ determines whether the player can reach the candidate from the start region. Surface support is checked by downward ray casting, collision is evaluated using object bounding boxes, and reachability is computed from the scene navigation mesh.

\paragraph{Semantic Location Selection.}
The geometric checks determine whether a candidate is feasible, but they do not determine whether the candidate is suitable for the intended gameplay. For each element $e$, the system constructs a top-down scene observation $\mathcal{T}_e$ that marks the candidates in $\mathcal{P}_e$. A vision-language model ranks the feasible candidates according to their compatibility with the functional description of $e$:
\begin{equation}
p_e^{*}
=
\arg\max_{p\in\mathcal{P}_e}
s_{\mathrm{VLM}}(e,p,\mathcal{T}_e),
\end{equation}
where $s_{\mathrm{VLM}}$ measures the semantic suitability of candidate $p$ for element $e$. Thus, geometric validation eliminates invalid locations, while semantic selection chooses among the remaining candidates. If $\mathcal{P}_e$ is empty, the unsatisfied placement constraint is reported to the subsequent repair procedure.

\paragraph{Asset and Logic Instantiation.}
The selected location $p_e^{*}$ determines the concrete realization of each gameplay element $e\in\mathcal{E}$. For a physical game element, the system generates or retrieves a compatible asset, adjusts its scale and orientation, constructs its collision geometry, and instantiates the corresponding Blender object at $p_e^{*}$. For a logical gameplay region, no visible asset is required. Instead, the system records the region annotation and its spatial extent for subsequent reconstruction as a gameplay volume. In both cases, the realization is associated with the persistent identifier $\operatorname{id}_e$ and the correspondence record $\mu(e)$.

For each interaction $\iota_k=(P_k,t_k,c_k,\Delta s_k)\in\mathcal{I}$, the code generation model maps the declarative interaction into an executable logic unit:
\begin{equation}
\ell_k=(\widehat{c}_k,\widehat{a}_k),
\end{equation}
where $\widehat{c}_k$ detects the triggering event $t_k$ and evaluates the precondition $c_k$, while $\widehat{a}_k$ executes the game state update $\Delta s_k$. The resulting logic units form
\begin{equation}
\mathcal{L}=
\left\{
\ell_k
\right\}_{k=1}^{K}.
\end{equation}
Each logic unit is bound to the actors and gameplay elements in $P_k$ through their persistent identifiers. This binding connects the declarative interaction specification to the concrete scene elements and enables the same interaction logic to be reconstructed in the target game engine. The resulting realization contains the instantiated scene elements, their spatial arrangement, and the executable logic required to implement the gameplay specification.

\subsection{Unreal Project Reconstruction}
\label{sec:reconstruction}

Given the realized scene and gameplay logic, the system reconstructs an Unreal Engine project while preserving spatial organization and persistent correspondences. The realized world is represented as
\begin{equation}
\mathcal{W}=(\mathcal{O},\mathcal{V},\mathcal{G}),
\end{equation}
where $\mathcal{O}$ stores the realized scene objects, asset references, transformations, and persistent identifiers, $\mathcal{V}$ stores materials, lighting, and environmental settings, and $\mathcal{G}$ stores executable logic, game state variables, and bindings between logic units and gameplay elements. The correspondence record $\mu(e)$ connects each element in $\mathcal{D}$ to its realization in $\mathcal{W}$.

\paragraph{Scene and Visual Reconstruction.}
Conditioned on $\mathcal{W}$, the system generates Unreal construction code to instantiate actors and gameplay volumes. Each physical element restores its asset, transform, collision configuration, and persistent identifier, while logical regions are reconstructed as Unreal gameplay volumes. The visual configuration in $\mathcal{V}$ is translated into Unreal materials, lights, and environmental settings, while asset references and transformations are resolved using the identifiers and coordinate conventions shared by the Blender and Unreal representations.

\paragraph{Gameplay Binding.}
For each logic unit $\ell_k\in\mathcal{L}$, the system creates the runtime condition and action in Unreal Engine. The runtime condition detects the triggering event and evaluates the precondition specified by $\iota_k$, while the action updates the game state according to $\Delta s_k$. The logic unit is bound to the actors and gameplay elements in $P_k$ through their persistent identifiers, preserving the correspondence between the declarative gameplay specification and its engine implementation.

\paragraph{Project Assembly.}
The generated construction code initializes the game state, assembles the reconstructed actors and gameplay volumes, restores their visual and collision configurations, and registers the executable logic units. The resulting initial project is denoted by
\begin{equation}
\mathcal{U}^{(0)}
=
\operatorname{Assemble}
\left(
\mathcal{W},\mathcal{D}
\right).
\end{equation}

\paragraph{Execution Feedback and Repair.}
The initial project is compiled and executed to collect engine and gameplay feedback:
\begin{equation}
\mathcal{F}^{(t)}
=
\left(
\mathcal{F}_{\mathrm{engine}}^{(t)},
\mathcal{F}_{\mathrm{game}}^{(t)}
\right),
\end{equation}
where $\mathcal{F}_{\mathrm{engine}}^{(t)}$ contains compilation and runtime diagnostics, and $\mathcal{F}_{\mathrm{game}}^{(t)}$ records failures detected by objective assertions and scripted interaction traces derived from $g$ and $\mathcal{I}$. Using the correspondence records, the system localizes the affected components as
\begin{equation}
\mathcal{C}^{(t)}
=
\operatorname{Localize}
\left(
\mathcal{F}^{(t)},
\{\mu(e)\}_{e\in\mathcal{E}}
\right)
\end{equation}
and updates only these components:
\begin{equation}
\left(
\mathcal{W}^{(t+1)},\mathcal{U}^{(t+1)}
\right)
=
\operatorname{Repair}
\left(
\mathcal{W}^{(t)},\mathcal{U}^{(t)},
\mathcal{C}^{(t)},\mathcal{F}^{(t)},\mathcal{S},\mathcal{D}
\right).
\end{equation}
The repaired project is executed again until the specified gameplay objectives are satisfied or the repair budget is exhausted.

\section{Experiments}
We evaluate whether Code2Games generates gaming worlds that remain faithful to the intended design and whether these worlds can be successfully adapted and executed in a game engine.

\subsection{Benchmark and Metrics}
\label{sec:benchmark_metrics}

\paragraph{GameCode4D Benchmark.}

We construct \textbf{GameCode4D Benchmark}, a prompt-to-gaming world test set comprising ten
frozen prompts $\mathcal{P}=\{p_i\}_{i=1}^{10}$. Each prompt provides only the genre, viewpoint, and high-level objective. Given this description, each method must generate the corresponding gaming world, including its scene, mechanics, assets, spatial layout, and interactions.

\begin{wrapfigure}{r}{0.44\textwidth}
    \centering
    \vspace{-3.5em}
    \includegraphics[width=\linewidth]{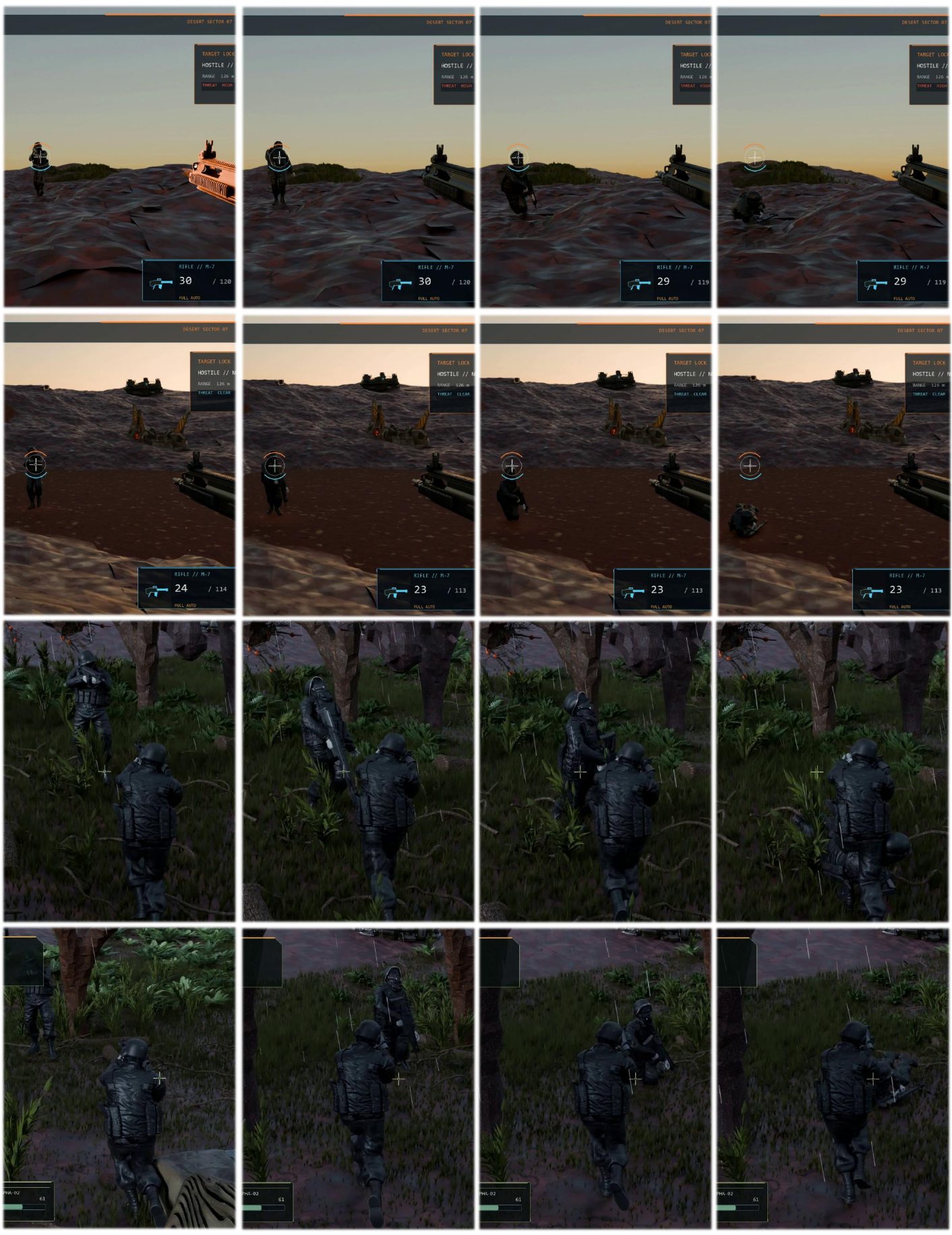}
    \caption{Qualitative results of gaming worlds generated by Code2Games.}
    \label{fig:qualitative_results}
    \vspace{-2em}
\end{wrapfigure}

For each generated game, we record two unedited, approximately 50-second rollouts, yielding 20 videos per method. The two rollouts execute distinct action routes on the same artifact and expose different regions and interactions. Scores are averaged across all rollouts and games. Appendix~\ref{app:experimental_protocols} specifies the recording and aggregation protocol. Figure~\ref{fig:qualitative_results} shows representative gaming worlds
generated by Code2Games across different game scenarios.
\paragraph{Evaluation metrics.}
Our evaluation covers visual quality, interactive fidelity, multimodal artifact quality, and playability. We compute image quality (IQ), dynamic degree (Dynamic), and motion smoothness (Motion) with the VBench implementations
\citep{huang2024vbench,huang2025vbench++}. Following WBench \citep{ying2026wbench}, setting adherence (Setting), interaction adherence (Inter.), and physics compliance (Physics) are evaluated against the game prompt, controller trace, and resulting rollout. The code, image, and video scores follow the modality-specific evaluation of V-GameGym \citep{zhang2026v}. Following GameCraft-Bench\citep{luo2026gamecraft}, we assess playable-game quality through build reliability (Build), core mechanics ($M$), content depth ($D$), functional visuals ($V$), and art and presentation ($A$). Build measures autonomous build-and-run reliability rather than final project executability. It is the average pass rate of standardized engine checks completed without human intervention, compilation failure, or runtime crash. In our experiments, all projects were successfully repaired within five iterations at most. The Human score is obtained from a user study, where participants rate generated games under the same playable-game criteria on a 0--100 scale. Appendix~\ref{app:experimental_protocols} defines each measure and provides the LLM-judge configuration.
\begin{table}[!h]
\caption{
Results on the GameCode4D Benchmark using the same ten prompts. The top panel reports visual quality and interactive fidelity, while the bottom panel reports multimodal artifact quality and
playable-game quality. Bold values indicate the best results in each column.}
\label{tab:main_results}
\centering
\scriptsize
\setlength{\tabcolsep}{1.8pt}
\renewcommand{\arraystretch}{0.98}
\begin{tabular*}{\columnwidth}{@{\extracolsep{\fill}}lcccccc@{}}
\toprule
\multirow{2}{*}{Method}
& \multicolumn{3}{c}{Visual Quality}
& \multicolumn{3}{c}{Interactive Fidelity} \\
\cmidrule(lr){2-4}\cmidrule(lr){5-7}
& IQ $\uparrow$ & Dynamic $\uparrow$ & Motion $\uparrow$
& Setting $\uparrow$ & Inter. $\uparrow$ & Physics $\uparrow$ \\
\midrule
OpenGame + Qwen3.8-Max & 62.8 & 80.5 & 94.1 & 78.5 & 65.0 & 67.0 \\
AutoUE + Qwen3.8-Max & 63.1 & 82.2 & 93.8 & 80.3 & 65.8 & 68.7 \\
Claude Code + Claude Opus 5 & 63.6 & 81.8 & 95.1 & 80.0 & 66.0 & 68.4 \\
Codex + GPT-5.6 Sol & 63.3 & 83.2 & 94.9 & 80.7 & 67.8 & 69.2 \\
Qwen Code + Qwen3.8-Max & 61.7 & 77.8 & 93.5 & 75.2 & 60.8 & 64.8 \\
\midrule
\textbf{Code2Games (ours) + Qwen3.8-Max} & 64.3 & 84.8 & 94.9 & 83.1 & 70.6 & 70.3 \\
\textbf{Code2Games (ours) + GPT-5.6 Sol} & 65.4 & 87.2 & 95.8 & 86.3 & \textbf{76.2} & \textbf{74.4} \\
\textbf{Code2Games (ours) + Claude Opus 5} & \textbf{66.1} & \textbf{88.2} & \textbf{96.1} & \textbf{87.5} & 75.8 & 73.9 \\
\midrule
\end{tabular*}
\vspace{-1.5pt}
\begin{tabular*}{\columnwidth}{@{\extracolsep{\fill}}lccccccccc@{}}
\multirow{2}{*}{Method}
& \multicolumn{3}{c}{Multimodal Artifact}
& \multicolumn{6}{c}{Playable Game} \\
\cmidrule(lr){2-4}\cmidrule(lr){5-10}
& Code $\uparrow$ & Image $\uparrow$ & Video $\uparrow$
& Build $\uparrow$ & M $\uparrow$ & D $\uparrow$ & V $\uparrow$ & A $\uparrow$ & Human $\uparrow$ \\
\midrule
OpenGame + Qwen3.8-Max & 94.2 & 20.6 & 22.9 & 92.6 & 53.5 & 39.8 & 43.6 & 39.7 & 60.8 \\
AutoUE + Qwen3.8-Max & 93.8 & 20.3 & 22.7 & 91.8 & 54.5 & 38.9 & 44.3 & 39.0 & 61.3 \\
Claude Code + Claude Opus 5 & 96.5 & 20.7 & 23.4 & \textbf{94.5} & 56.8 & 42.2 & 46.0 & 41.5 & 63.8 \\
Codex + GPT-5.6 Sol & 97.3 & 20.3 & 24.3 & 94.2 & 58.1 & 42.9 & 45.4 & 40.4 & 64.2 \\
Qwen Code + Qwen3.8-Max & 92.5 & 17.0 & 19.3 & 93.3 & 43.5 & 31.0 & 35.8 & 30.4 & 51.5 \\
\midrule
\textbf{Code2Games (ours) + Qwen3.8-Max} & 96.1 & 24.7 & 28.8 & 94.1 & 58.2 & 46.0 & 51.5 & 45.5 & 68.0 \\
\textbf{Code2Games (ours) + GPT-5.6 Sol} & \textbf{98.5} & 28.7 & 33.2 & 95.9 & 67.8 & 54.8 & 59.5 & 54.3 & 75.5 \\
\textbf{Code2Games (ours) + Claude Opus 5} & 98.1 & \textbf{30.3} & \textbf{34.9} & \textbf{97.7} & \textbf{68.8} & \textbf{56.7} & \textbf{61.0} & \textbf{57.0} & \textbf{77.5} \\
\bottomrule
\end{tabular*}
\end{table}

\label{sec:main_results}

\subsection{Main Results}

\begin{figure*}[h]
    \centering
    \includegraphics[width=\textwidth]{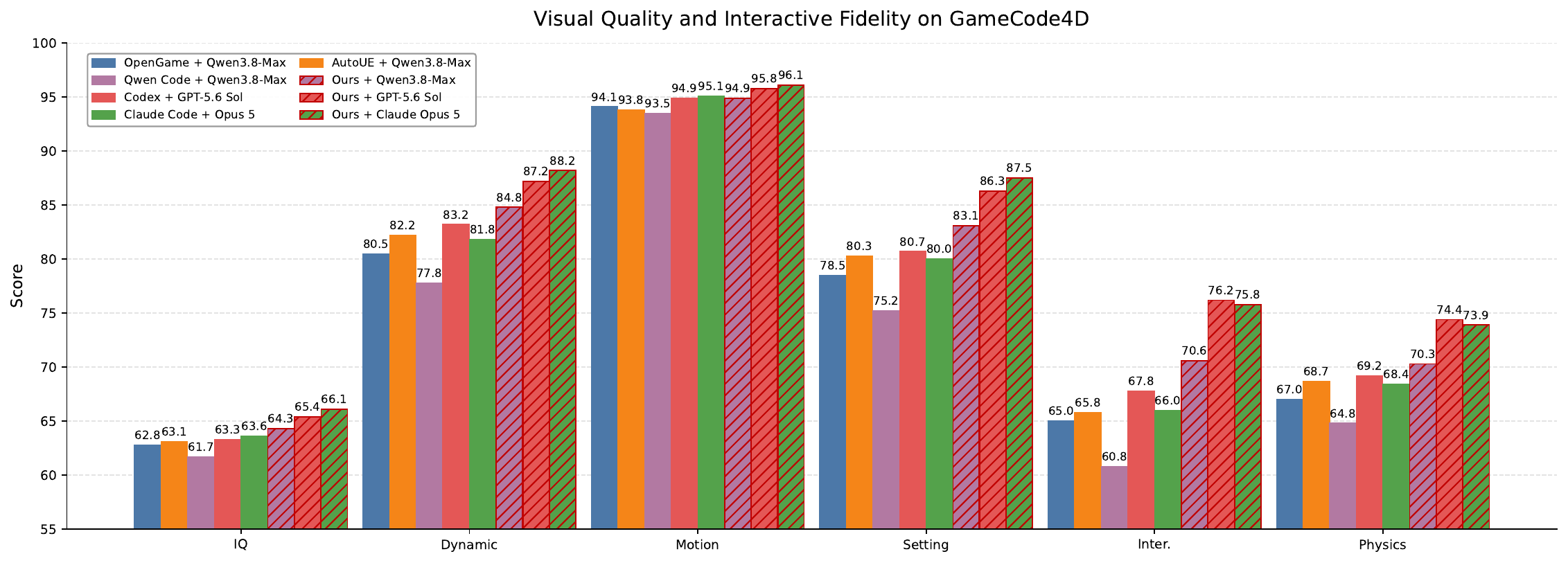}
    \caption{Visual quality and interactive fidelity on the GameCode4D benchmark.}
    \label{fig:visual_interactive}
\end{figure*}

Table~\ref{tab:main_results} compares OpenGame and AutoUE 
\citep{jiang2026opengame,yin2026autoue}, three coding-agent configurations, and
three Code2Games configurations under the same ten prompts. The two panels cover the quality of gaming worlds, interactive fidelity, multimodal artifact quality, and playable game quality. The comparison of visual quality and interactive fidelity is visualized in Figure~\ref{fig:visual_interactive}.

Across matched backbones, Code2Games consistently improves the quality of generated gaming worlds over the corresponding coding-agent baselines while maintaining high build reliability across all three backbone configurations. The strongest configuration with Claude Opus 5 achieves a build reliability of 97.7\%. These results demonstrate the effectiveness of the execution-oriented generation pipeline in producing high-quality and reliable playable games. Even the Qwen3.8-Max configuration exceeds the strongest baseline on 12 of 15 measures and trails only in motion, code quality, and build reliability. The Claude Opus 5 configuration achieves the best results on 12 of the 15 evaluated metrics, while the GPT-5.6 Sol configuration leads interaction adherence, physics compliance and code quality. Relative to the strongest baseline in each column, the largest gains occur in art and presentation (+15.5), functional visuals (+15.0), content depth (+13.8), and Human (+13.3), compared with +9.6 for image quality and +1.0 for motion. This pattern suggests that the main benefit lies in scene-grounded gameplay rather than visual polish alone. Appendix~\ref{app:experimental_protocols} provides the study protocol and diagnostic rubric.

\subsection{Ablation Studies}

To better understand how different components contribute to Code2Games, we conduct component-wise ablation studies by selectively removing or incrementally introducing individual components while keeping the remaining generation pipeline unchanged. Specifically, we analyze four aspects: scene representation, gameplay specification, scene constraints, and execution feedback. All variants use the same backbone model and evaluation protocols.

\paragraph{Scene and Gameplay Component Analysis.} To evaluate the contribution of scene-grounding components, we remove scene representation, gameplay specification, and scene constraint individually from the full Code2Games framework. All other modules remain unchanged, allowing us to measure the specific impact of each component on generated game quality.

\begin{wraptable}[11]{R}{0.48\columnwidth}
\vspace{-1.5em}
\centering

\scriptsize
\setlength{\tabcolsep}{2pt}

\caption{Ablation of Scene-grounding.}
\label{tab:ablation_scene}
\vspace{-1.5em}
\resizebox{\linewidth}{!}{
\begin{tabular}{lcccccc}
\toprule
Variant
& Setting $\uparrow$
& Inter. $\uparrow$
& Physics $\uparrow$
& M $\uparrow$
& D $\uparrow$
& V $\uparrow$
\\
\midrule

w/o Scene Representation
&78.2&71.5&72.8&60.5&49.8&55.0
\\

w/o Gameplay Specification
&82.1&69.8&73.5&62.0&51.5&56.2
\\

w/o Scene Constraints
&83.4&74.1&66.2&64.5&53.0&57.8
\\

\midrule

\textbf{Full Code2Games}
&\textbf{87.5}
&\textbf{75.8}
&\textbf{73.9}
&\textbf{68.8}
&\textbf{56.7}
&\textbf{61.0}
\\

\bottomrule
\end{tabular}
}

\vspace{0.8em}

\caption{Ablation of Execution feedback.}
\label{tab:ablation_feedback}
\vspace{-1.5em}
\resizebox{\linewidth}{!}{
\begin{tabular}{lcccccc}
\toprule
Variant
& Build $\uparrow$
& M $\uparrow$
& D $\uparrow$
& V $\uparrow$
& Inter. $\uparrow$
& Physics $\uparrow$
\\
\midrule

Initial
&78.5&64.0&53.0&58.0&58.0&54.0
\\

+ Engine Feedback Repair
&88.7&65.5&53.5&59.0&59.0&55.0
\\

+ Gameplay Feedback Repair
&\textbf{97.7}
&\textbf{68.8}
&\textbf{56.7}
&\textbf{61.0}
&\textbf{75.8}
&\textbf{73.9}
\\

\bottomrule
\end{tabular}
}

\vspace{-3em}
\end{wraptable}

As shown in Table~\ref{tab:ablation_scene}, removing any scene-grounding component degrades both interactive fidelity and playable-game quality. Removing scene representation mainly reduces Setting (87.5$\rightarrow$78.2), as the agent loses scene organization information. Removing gameplay specification causes the largest drop in Inter. (75.8$\rightarrow$69.8), while removing scene constraints most affects Physics (73.9$\rightarrow$66.2), demonstrating the importance of spatial validity during game realization.

\paragraph{Progressive Execution Feedback and Repair.} To evaluate the contribution of execution feedback and repair, we progressively introduce engine-feedback repair and gameplay-feedback repair from an initial Unreal reconstruction without execution feedback. As shown in Table~\ref{tab:ablation_feedback}, progressively adding feedback-driven repair improves both execution reliability and gameplay quality. Build increases from 78.5 to 97.7, while Inter. improves from 58.0 to 75.8. Engine-feedback repair mainly resolves engine-level failures, while gameplay-feedback repair further improves the correctness of interactive behaviors through gameplay validation.

\section{Conclusion}
We presented Code2Games, a framework that transforms a game intent and its paired Code2Worlds scene into a gaming world and adapts it to Unreal Engine 5. Code2Games connects scene understanding, gameplay planning, constrained world realization, and engine reconstruction through a shared scene-and-gameplay representation, while persistent identifiers and execution-guided repair preserve consistency across generation stages. Evaluation on GameCode4D Benchmark shows that the framework improves visual quality, interactive fidelity, multimodal artifact quality, and playable-game quality over matched coding agents across multiple backbone models. The ablation results further demonstrate the complementary contributions of structured game design, scene-aware constraints, and execution-guided reflection, with reflection also improving build reliability. These results show that grounding gameplay generation in explicit scene structure and closing the generation loop with engine feedback provides an effective route from generated worlds to playable games.

\bibliography{iclr2026_conference}

@inproceedings{huang2024vbench,
  title={Vbench: Comprehensive benchmark suite for video generative models},
  author={Huang, Ziqi and He, Yinan and Yu, Jiashuo and Zhang, Fan and Si, Chenyang and Jiang, Yuming and Zhang, Yuanhan and Wu, Tianxing and Jin, Qingyang and Chanpaisit, Nattapol and others},
  booktitle={2024 IEEE/CVF Conference on Computer Vision and Pattern Recognition (CVPR)},
  pages={21807--21818},
  year={2024},
  organization={IEEE}
}

@article{huang2025vbench++,
  title={Vbench++: Comprehensive and versatile benchmark suite for video generative models},
  author={Huang, Ziqi and Zhang, Fan and Xu, Xiaojie and He, Yinan and Yu, Jiashuo and Dong, Ziyue and Ma, Qianli and Chanpaisit, Nattapol and Si, Chenyang and Jiang, Yuming and others},
  journal={IEEE Transactions on Pattern Analysis and Machine Intelligence},
  year={2025},
  publisher={IEEE}
}

@article{ying2026wbench,
  title={Wbench: A comprehensive multi-turn benchmark for interactive video world model evaluation},
  author={Ying, Kaining and Hu, Hengrui and Ren, Siyu and Li, Jiamu and Chen, Fengjiao and Wang, Ziwen and Cao, Xuezhi and Cai, Xunliang and Ding, Henghui},
  journal={arXiv preprint arXiv:2605.25874},
  year={2026}
}

@inproceedings{zhang2026v,
  title={V-gamegym: Visual game generation for code large language models},
  author={Zhang, Wei and Yang, Jian and Tao, Renshuai and Chai, Linzheng and Guo, Shuyue and Wu, Jiajun and Chen, Xiaoming and Cui, Ganqu and Ding, Ning and Xu, Xander and others},
  booktitle={Findings of the Association for Computational Linguistics: ACL 2026},
  pages={5613--5641},
  year={2026}
}

@article{luo2026gamecraft,
  title={GameCraft-Bench: Can Agents Build Playable Games End-to-End in a Real Game Engine?},
  author={Luo, Tongxu and Wang, Rongsheng and Bi, Jiaxi and Xu, Chenming and Tang, Zhengyang and Chen, Jianlong and Liang, Juhao and Ji, Ke and Guo, Shuqi and Du, Yuhao and others},
  journal={arXiv preprint arXiv:2606.17861},
  year={2026}
}

@inproceedings{yin2026autoue,
  title={AutoUE: Automated Generation of 3D Games in Unreal Engine via Multi-Agent Systems},
  author={Yin, Lei and Cheng, Wentao and Qin, Zhida and Huang, Tianyu and Li, Yidong and Ding, Gangyi},
  booktitle={Findings of the Association for Computational Linguistics: ACL 2026},
  pages={2341--2364},
  year={2026}
}

@article{jiang2026opengame,
  title={Opengame: Open agentic coding for games},
  author={Jiang, Yilei and Hu, Jinyuan and Xiao, Qianyin and Zheng, Yaozhi and Ma, Ruize and Feng, Kaituo and Han, Jiaming and Peng, Tianshuo and Fan, Kaixuan and Zhang, Manyuan and others},
  journal={arXiv preprint arXiv:2604.18394},
  year={2026}
}

@inproceedings{duan2025worldscore,
  title={Worldscore: A unified evaluation benchmark for world generation},
  author={Duan, Haoyi and Yu, Hong-Xing and Chen, Sirui and Fei-Fei, Li and Wu, Jiajun},
  booktitle={Proceedings of the IEEE/CVF international conference on computer vision},
  pages={27713--27724},
  year={2025}
}

@article{zheng2025vbench2,
  title={Vbench-2.0: Advancing video generation benchmark suite for intrinsic faithfulness},
  author={Zheng, Dian and Huang, Ziqi and Liu, Hongbo and Zou, Kai and He, Yinan and Zhang, Fan and Gu, Lulu and Zhang, Yuanhan and He, Jingwen and Zheng, Wei-Shi and others},
  journal={arXiv preprint arXiv:2503.21755},
  year={2025}
}

@inproceedings{che2024gamegenx,
  title={Gamegen-x: Interactive open-world game video generation},
  author={Che, Haoxuan and He, Xuanhua and Liu, Quande and Jin, Cheng and Chen, Hao},
  booktitle={International Conference on Learning Representations},
  volume={2025},
  pages={37546--37593},
  year={2025}
}

@inproceedings{yu2025gamefactory,
  title={GameFactorly: Creating New Games with Generative Interactive Videos},
  author={Yu, Jiwen and Qin, Yiran and Wang, Xintao and Wan, Pengfei and Zhang, Di and Liu, Xihui},
  booktitle={2025 IEEE/CVF International Conference on Computer Vision (ICCV)},
  pages={11590--11599},
  year={2025},
  organization={IEEE}
}

@article{zhang2026code2worlds,
  title={Code2worlds: Empowering coding llms for 4d world generation},
  author={Zhang, Yi and Wang, Yunshuang and Zhang, Zeyu and Tang, Hao},
  journal={arXiv preprint arXiv:2602.11757},
  year={2026}
}

@article{menapace2021playable,
  title={Playable video generation},
  author={Menapace, Willi and Lathuiliere, Stephane and Tulyakov, Sergey and Siarohin, Aliaksandr and Ricci, Elisa},
  journal={arXiv preprint arXiv:2101.12195},
  year={2021}
}

@inproceedings{menapace2022playable,
  title={Playable environments: Video manipulation in space and time},
  author={Menapace, Willi and Lathuiliere, St{\'e}phane and Siarohin, Aliaksandr and Theobalt, Christian and Tulyakov, Sergey and Golyanik, Vladislav and Ricci, Elisa},
  booktitle={2022 IEEE/CVF Conference on Computer Vision and Pattern Recognition (CVPR)},
  pages={3574--3583},
  year={2022},
  organization={IEEE}
}

@inproceedings{bruce2024genie,
  title={Genie: Generative interactive environments},
  author={Bruce, Jake and Dennis, Michael D and Edwards, Ashley and Parker-Holder, Jack and Shi, Yuge and Hughes, Edward and Lai, Matthew and Mavalankar, Aditi and Steigerwald, Richie and Apps, Chris and others},
  booktitle={Forty-first international conference on machine learning},
  year={2024}
}

@article{alonso2024diamond,
  title={Diffusion for world modeling: Visual details matter in atari},
  author={Alonso, Eloi and Jelley, Adam and Micheli, Vincent and Kanervisto, Anssi and Storkey, Amos and Pearce, Tim and Fleuret, Fran{\c{c}}ois},
  journal={Advances in Neural Information Processing Systems},
  volume={37},
  pages={58757--58791},
  year={2024}
}

@inproceedings{valevski2025gamengen,
  title={Diffusion models are real-time game engines},
  author={Valevski, Dani and Leviathan, Yaniv and Arar, Moab and Fruchter, Shlomi},
  booktitle={International Conference on Learning Representations},
  volume={2025},
  pages={73754--73776},
  year={2025}
}

@article{huang2026vid2world,
  title={Vid2world: Crafting video diffusion models to interactive world models},
  author={Huang, Siqiao and Wu, Jialong and Zhou, Qixing and Miao, Shangchen and Long, Mingsheng},
  journal={arXiv preprint arXiv:2505.14357},
  year={2025}
}

@inproceedings{lu2025genex,
  title={Genex: Generating an explorable world},
  author={Lu, Taiming and Shu, Tianmin and Yuille, Alan and Khashabi, Daniel and Chen, Jieneng},
  booktitle={International Conference on Learning Representations},
  volume={2025},
  pages={52310--52335},
  year={2025}
}

@article{li2025hunyuan,
  title={Hunyuan-gamecraft: High-dynamic interactive game video generation with hybrid history condition},
  author={Li, Jiaqi and Tang, Junshu and Xu, Zhiyong and Wu, Longhuang and Zhou, Yuan and Shao, Shuai and Yu, Tianbao and Cao, Zhiguo and Lu, Qinglin},
  journal={arXiv preprint arXiv:2506.17201},
  volume={2},
  number={3},
  pages={6},
  year={2025}
}

@article{tang2025hunyuan,
  title={Hunyuan-gamecraft-2: Instruction-following interactive game world model},
  author={Tang, Junshu and Liu, Jiacheng and Li, Jiaqi and Wu, Longhuang and Yang, Haoyu and Zhao, Penghao and Gong, Siruis and Yuan, Xiang and Shao, Shuai and Zhang, Linfeng and others},
  journal={arXiv preprint arXiv:2511.23429},
  year={2025}
}

@article{deitke2022procthor,
  title={ProcTHOR: Large-Scale Embodied AI Using Procedural Generation},
  author={Deitke, Matt and VanderBilt, Eli and Herrasti, Alvaro and Weihs, Luca and Ehsani, Kiana and Salvador, Jordi and Han, Winson and Kolve, Eric and Kembhavi, Aniruddha and Mottaghi, Roozbeh},
  journal={Advances in neural information processing systems},
  volume={35},
  pages={5982--5994},
  year={2022}
}

@inproceedings{hollein2023text2room,
  title={Text2room: Extracting textured 3d meshes from 2d text-to-image models},
  author={H{\"o}llein, Lukas and Cao, Ang and Owens, Andrew and Johnson, Justin and Nie{\ss}ner, Matthias},
  booktitle={2023 IEEE/CVF International Conference on Computer Vision (ICCV)},
  pages={7875--7886},
  year={2023},
  organization={IEEE}
}

@article{feng2023layoutgpt,
  title={Layoutgpt: Compositional visual planning and generation with large language models},
  author={Feng, Weixi and Zhu, Wanrong and Fu, Tsu-jui and Jampani, Varun and Akula, Arjun and He, Xuehai and Basu, Sugato and Wang, Xin Eric and Wang, William Yang},
  journal={Advances in neural information processing systems},
  volume={36},
  pages={18225--18250},
  year={2023}
}

@article{lin2023layoutprompter,
  title={Layoutprompter: Awaken the design ability of large language models},
  author={Lin, Jiawei and Guo, Jiaqi and Sun, Shizhao and Yang, Zijiang and Lou, Jian-Guang and Zhang, Dongmei},
  journal={Advances in Neural Information Processing Systems},
  volume={36},
  pages={43852--43879},
  year={2023}
}

@inproceedings{yang2024holodeck,
  title={Holodeck: Language guided generation of 3d embodied ai environments},
  author={Yang, Yue and Sun, Fan-Yun and Weihs, Luca and VanderBilt, Eli and Herrasti, Alvaro and Han, Winson and Wu, Jiajun and Haber, Nick and Krishna, Ranjay and Liu, Lingjie and others},
  booktitle={2024 IEEE/CVF Conference on Computer Vision and Pattern Recognition (CVPR)},
  pages={16277--16287},
  year={2024},
  organization={IEEE}
}

@article{wang2023voyager,
  title={Voyager: An open-ended embodied agent with large language models},
  author={Wang, Guanzhi and Xie, Yuqi and Jiang, Yunfan and Mandlekar, Ajay and Xiao, Chaowei and Zhu, Yuke and Fan, Linxi and Anandkumar, Anima},
  journal={arXiv preprint arXiv:2305.16291},
  year={2023}
}

@article{yao2022react,
  title={React: Synergizing reasoning and acting in language models},
  author={Yao, Shunyu and Zhao, Jeffrey and Yu, Dian and Du, Nan and Shafran, Izhak and Narasimhan, Karthik and Cao, Yuan},
  journal={arXiv preprint arXiv:2210.03629},
  year={2022}
}

@article{zhou2024gala3d,
  title={Gala3d: Towards text-to-3d complex scene generation via layout-guided generative gaussian splatting},
  author={Zhou, Xiaoyu and Ran, Xingjian and Xiong, Yajiao and He, Jinlin and Lin, Zhiwei and Wang, Yongtao and Sun, Deqing and Yang, Ming-Hsuan},
  journal={arXiv preprint arXiv:2402.07207},
  year={2024}
}

@inproceedings{zhang2024scenewiz3d,
  title={Towards text-guided 3d scene composition},
  author={Zhang, Qihang and Wang, Chaoyang and Siarohin, Aliaksandr and Zhuang, Peiye and Xu, Yinghao and Yang, Ceyuan and Lin, Dahua and Zhou, Bolei and Tulyakov, Sergey and Lee, Hsin-Ying},
  booktitle={2024 IEEE/CVF Conference on Computer Vision and Pattern Recognition (CVPR)},
  pages={6829--6838},
  year={2024},
  organization={IEEE}
}

@inproceedings{zhang2024scenedreamer,
  title={3d-scenedreamer: Text-driven 3d-consistent scene generation},
  author={Zhang, Songchun and Zhang, Yibo and Zheng, Quan and Ma, Rui and Hua, Wei and Bao, Hujun and Xu, Weiwei and Zou, Changqing},
  booktitle={2024 IEEE/CVF Conference on Computer Vision and Pattern Recognition (CVPR)},
  pages={10170--10180},
  year={2024},
  organization={IEEE}
}

@article{wang2024robogen,
  title={Robogen: Towards unleashing infinite data for automated robot learning via generative simulation},
  author={Wang, Yufei and Xian, Zhou and Chen, Feng and Wang, Tsun-Hsuan and Wang, Yian and Fragkiadaki, Katerina and Erickson, Zackory and Held, David and Gan, Chuang},
  journal={arXiv preprint arXiv:2311.01455},
  year={2023}
}

@article{yang2024scenecraft,
  title={Scenecraft: Layout-guided 3d scene generation},
  author={Yang, Xiuyu and Man, Yunze and Chen, Jun-Kun and Wang, Yu-Xiong},
  journal={Advances in Neural Information Processing Systems},
  volume={37},
  pages={82060--82084},
  year={2024}
}

@inproceedings{jimenez2024swe,
  title={Swe-bench: Can language models resolve real-world github issues?},
  author={Jimenez, Carlos E and Yang, John and Wettig, Alexander and Yao, Shunyu and Pei, Kexin and Press, Ofir and Narasimhan, Karthik},
  booktitle={International Conference on Learning Representations},
  volume={2024},
  pages={54107--54157},
  year={2024}
}

@article{shinn2023reflexion,
  title={Reflexion: Language agents with verbal reinforcement learning},
  author={Shinn, Noah and Cassano, Federico and Gopinath, Ashwin and Narasimhan, Karthik and Yao, Shunyu},
  journal={Advances in neural information processing systems},
  volume={36},
  pages={8634--8652},
  year={2023}
}

@article{wang2024architect,
  title={Architect: Generating vivid and interactive 3d scenes with hierarchical 2d inpainting},
  author={Wang, Yian and Qiu, Xiaowen and Liu, Jiageng and Chen, Zhehuan and Cai, Jiting and Wang, Yufei and Wang, Tsun-Hsuan and Xian, Zhou and Gan, Chuang},
  journal={Advances in Neural Information Processing Systems},
  volume={37},
  pages={67575--67603},
  year={2024}
}

@inproceedings{bokhovkin2025scenefactor,
  title={Scenefactor: Factored latent 3d diffusion for controllable 3d scene generation},
  author={Bokhovkin, Aleksey and Meng, Quan and Tulsiani, Shubham and Dai, Angela},
  booktitle={Proceedings of the IEEE/CVF Conference on Computer Vision and Pattern Recognition},
  pages={628--639},
  year={2025}
}

@inproceedings{sun2025layoutvlm,
  title={Layoutvlm: Differentiable optimization of 3d layout via vision-language models},
  author={Sun, Fan-Yun and Liu, Weiyu and Gu, Siyi and Lim, Dylan and Bhat, Goutam and Tombari, Federico and Li, Manling and Haber, Nick and Wu, Jiajun},
  booktitle={2025 IEEE/CVF Conference on Computer Vision and Pattern Recognition (CVPR)},
  pages={29469--29478},
  year={2025},
  organization={IEEE}
}

@inproceedings{zhang2026rpgagent,
  title={RPGAgent: Driving Coherent Story-to-Play Generation with an LLM-Based Multi-Agent System},
  author={Zhang, Shunan and Xiao, Yi and Ma, Ruoxuan and Leung, Chi-sing},
  booktitle={Proceedings of the 2026 CHI Conference on Human Factors in Computing Systems},
  pages={1--22},
  year={2026}
}

@inproceedings{hong2025chatge,
  title={Game development as human-llm interaction},
  author={Hong, Jiale and Wu, Hongqiu and Zhao, Hai},
  booktitle={Proceedings of the 63rd Annual Meeting of the Association for Computational Linguistics (Volume 1: Long Papers)},
  pages={4333--4354},
  year={2025}
}

@inproceedings{wu2024idge,
  title={Instruction-driven game engine: A poker case study},
  author={Wu, Hongqiu and Liu, Xingyuan and Wang, Yan and Zhao, Hai},
  booktitle={Proceedings of the 2024 Conference on Empirical Methods in Natural Language Processing: System Demonstrations},
  pages={507--519},
  year={2024}
}

@inproceedings{li2025cardiverse,
  title={Cardiverse: Harnessing llms for novel card game prototyping},
  author={Li, Danrui and Zhang, Sen and Sohn, Samuel S and Hu, Kaidong and Usman, Muhammad and Kapadia, Mubbasir},
  booktitle={Proceedings of the 2025 Conference on Empirical Methods in Natural Language Processing},
  pages={29723--29750},
  year={2025}
}
\bibliographystyle{iclr2026_conference}

\clearpage
\appendix
\section*{Appendix}
\section{Limitation and Future Work}
\label{app:limitations}

Code2Games currently encounters a visual fidelity gap when adapting Blender
scenes to a game engine. Differences in material representations, procedural
shader nodes, lighting, and rendering may alter the appearance of the
reconstructed scene, including color, surface detail, transparency, and
atmosphere. Consequently, the Unreal reconstruction may not fully preserve the
visual appearance of the source Blender scene. Future work will investigate
improved material translation and rendering calibration to reduce these
discrepancies and better preserve visual consistency across engines.

\section{Ablation Study}
\label{app:ablation_studies}

The main paper reports component-wise ablations of scene representation,
gameplay specification, scene constraints, and execution-guided repair.
To avoid duplicating these results, the supplementary material provides only
a finer-grained decomposition of scene-constrained game realization,
reported in Table~\ref{tab:ablation_constraint_progressive}.

\section*{Additional Ablation Results}
\label{app:additional_ablation}

The main paper evaluates the contribution of scene representation, gameplay
specification, scene constraints, and execution-guided repair. Here, we provide
an additional fine-grained analysis of the scene-constrained realization
module by decomposing the constraints used during gameplay-element placement.
All variants use the same backbone model, game-intent prompts, assets, and
evaluation protocol as the full Code2Games framework.

\paragraph{Progressive Scene-Constrained Game Realization.}
We begin with an unconstrained realization strategy that directly instantiates
gameplay elements according to the gameplay specification without considering
scene-region compatibility or geometric validity. We then progressively
introduce region compatibility and geometric feasibility checks. Region
compatibility restricts gameplay elements to semantically appropriate scene
regions, while geometric feasibility verifies valid surface support,
collision-free placement, and accessibility. The full variant combines these
constraints with scene-constrained candidate selection.

\begin{table}[h]
\centering
\small
\setlength{\tabcolsep}{7pt}
\renewcommand{\arraystretch}{1.12}
\caption{Fine-grained ablation of scene-constrained game realization.}
\label{tab:ablation_constraint_progressive}
\begin{tabular}{lcccc}
\toprule
Variant &
M $\uparrow$ &
D $\uparrow$ &
V $\uparrow$ &
A $\uparrow$ \\
\midrule
Unconstrained Realization
& 58.0 & 45.0 & 52.0 & 45.0 \\

+ Region Compatibility
& 62.0 & 50.0 & 55.5 & 50.0 \\

+ Geometric Feasibility Checks
& 65.0 & 53.0 & 58.0 & 53.0 \\

+ Full Scene-Constrained Selection
& \textbf{68.8} & \textbf{56.7} & \textbf{61.0} & \textbf{57.0} \\
\bottomrule
\end{tabular}
\end{table}

As shown in Table~\ref{tab:ablation_constraint_progressive}, performance
improves consistently as additional scene constraints are introduced.
Region compatibility first improves the semantic alignment between gameplay
elements and their surrounding environments. Geometric feasibility checks
provide further gains by filtering unsupported, colliding, or inaccessible
placements. Combining both forms of constraint with scene-constrained
selection achieves the strongest results, improving M from 58.0 to 68.8,
D from 45.0 to 56.7, V from 52.0 to 61.0, and A from 45.0 to 57.0.
These results show that semantic compatibility and geometric validity address
complementary failure modes during gameplay-element realization.

\FloatBarrier

\section{Implementation Details}
\label{app:implementation}

Code2Games uses models from the Qwen family for language and visual reasoning,
while gameplay assets are generated using Hunyuan3D. Scene understanding,
gameplay-element placement, and visual inspection are performed in Blender
4.2.0 through its Python API, and the resulting gaming worlds are reconstructed
as Unreal Engine 5 projects for executable gameplay. Additional implementation
and evaluation settings used throughout the experiments are provided in the
detailed supplementary material below.

\section{Benchmark Details}
\label{app:benchmark_details}

GameCode4D evaluates the generation of playable gaming worlds across diverse
game scenarios, including reconnaissance, combat, monster hunting, relic
running, racing, skiing, underwater exploration, flight, and resource
collection. Each game-intent prompt specifies the environment, player role,
objective, and required interactions, while several scenarios further include
variants with different environments, weather conditions, or routes. The
complete game-intent prompts used in our qualitative evaluation are reported
in Table~\ref{tab:game_intents}.

\section*{Benchmark Prompts}
\label{app:benchmark_prompts}

Table~\ref{tab:game_intents} lists the game-intent prompts used for the
qualitative examples. The prompts cover diverse gameplay settings, objectives,
and interaction types, including combat, exploration, traversal, racing,
underwater exploration, flight, and resource collection.

\footnotesize
\setlength{\tabcolsep}{4pt}
\renewcommand{\arraystretch}{1.08}

\begin{longtable}{
@{}
p{0.035\textwidth}
p{0.61\textwidth}
>{\raggedright\arraybackslash}p{0.13\textwidth}
>{\raggedright\arraybackslash}p{0.17\textwidth}
@{}
}

\caption{Game-intent prompts for the qualitative examples. IDs match the qualitative figure sequence.}
\label{tab:game_intents}
\\

\toprule
ID & Prompt Content & Game Scenario & Core Interactions \\
\midrule
\endfirsthead

\toprule
ID & Prompt Content & Game Scenario & Core Interactions \\
\midrule
\endhead

\endfoot

\bottomrule
\endlastfoot

01 &
Create a first-person sci-fi assault on a red desert ridgeline. Eliminate three hostile sentries and secure the remote signal relay. &
FPS &
Shoot; survive; secure relay \\
\midrule

02 &
Create a first-person sci-fi combat sweep on a red-desert ridgeline. Track and eliminate three hostile sentries, then secure the remote signal uplink. &
FPS &
Track; eliminate; secure uplink \\
\midrule

03 &
Create a third-person monster hunt in a dark forest. Track the Thornback, break its neck armor, defeat it, and claim the trophy. &
Monster hunt &
Track; break armor; claim trophy \\
\midrule

04 &
Create a storm-weather Thornback hunt through a wet forest. Track the creature in low visibility, defeat it, and retrieve the trophy. &
Monster hunt &
Track; fight; retrieve trophy \\
\midrule

05 &
Create a high-speed off-road rally through a desert oasis. Drive across uneven terrain, use nitro, pass eight checkpoints, and reach the finish. &
Racing &
Drive; boost; pass checkpoints \\
\midrule

06 &
Create a desert-storm rally across layered dunes. Control the vehicle over changing slopes, pass eight rally gates, and complete the stage. &
Racing &
Steer; boost; pass gates \\
\midrule

07 &
Create a third-person alpine slalom on a bright snow course. Control speed and direction, clear eight gates, and reach the finish arch. &
Skiing &
Steer; clear gates; finish \\
\midrule

08 &
Create a golden-hour alpine slalom route. Descend through warm-lit snow, clear eight ordered gates, and reach the finish without penalties. &
Skiing &
Descend; clear gates; finish \\
\midrule

09 &
Create a third-person reconnaissance game on a rocky ridgeline. Explore landmarks, locate a hidden signal, inspect a shrine, and reach extraction. &
Temple run &
Navigate; inspect; extract \\
\midrule

10 &
Create a third-person relic runner in a rain-soaked jungle temple. Dodge ruin obstacles, collect five relics, and reach the escape shrine. &
Temple run &
Dodge; collect; escape \\
\midrule

11 &
Create a third-person tactical mission in a dark rainy jungle. Follow tracks, avoid or eliminate patrols, and secure the target uplink. &
Third-person shooter &
Track; evade; secure uplink \\
\midrule

12 &
Create a third-person coastal-jungle infiltration. Scout through vegetation, flank hostile patrols, and secure the marked relay with a silenced weapon. &
Third-person shooter &
Scout; flank; secure relay \\
\midrule

13 &
Create a free-swimming scientific survey in a coral reef. Collect three samples, investigate two marked sites, and return to extraction. &
Underwater exploration &
Swim; collect; investigate \\
\midrule

14 &
Create a blackwater anomaly survey under low visibility. Collect three samples, scan two anomalous sites, monitor sonar, and extract after completion. &
Underwater exploration &
Collect; scan; extract \\
\midrule

15 &
Create an F-104 low-altitude flight mission at dawn. Follow the safe corridor, activate four flight cells, and lock the summit mast. &
Wingsuit &
Fly; activate cells; lock mast \\
\midrule

16 &
Create a summit-intercept flight over foggy mountains. Follow the designated corridor, activate four terrain-link cells, and lock the summit mast. &
Wingsuit &
Hold corridor; activate; lock \\
\midrule

17 &
Create a third-person fantasy archer assassination in a rocky desert. Scout the arena, take cover, locate the target, and eliminate it with a bow. &
Assassin's Creed-style &
Scout; take cover; shoot \\
\midrule

18 &
Create a third-person archer hunt in a dead-tree wasteland. Position using sparse terrain, aim at distant targets, and eliminate them with a bow. &
Assassin's Creed-style &
Position; aim; eliminate \\
\midrule

19 &
Create a third-person block-style scavenging challenge in a red desert canyon. Explore markers, collect resources, avoid hazards, and complete the required set. &
Minecraft-style &
Explore; collect; manage inventory \\
\midrule

20 &
Create a block-style resource expedition across rocky clearings and forest terrain. Traverse multiple regions, gather crafting materials, and complete the resource set. &
Minecraft-style &
Traverse; collect; complete \\

\end{longtable}

\normalsize

\normalsize

\section{Experimental Protocols}
\label{app:experimental_protocols}

We use a unified experimental protocol for recording, aggregating, and
evaluating the generated gaming worlds. The protocol covers gameplay rollout
recording, automatic metric computation, LLM-as-a-Judge evaluation, and the
human user study, with identical settings applied to all compared methods
unless otherwise specified. Detailed metric definitions, scoring procedures,
and evaluation configurations are provided in the supplementary material
below.

\section*{Detailed Experimental Protocols}

\subsection*{Recording and Aggregation Protocol}

For each generated game, we record two unedited gameplay rollouts of
approximately 50 seconds on the same generated artifact. The two rollouts
follow predefined but distinct action routes so that they cover different
scene regions and interactions. Each route contains the initial game state,
navigation, at least one core interaction, a visible gameplay-state transition,
and a goal or progress signal. All compared methods use the same recording
configuration and controller budget.

Each rollout is evaluated independently. For every game, we first average the
scores of its two rollouts and then average the resulting game-level scores
across the ten benchmark games. This produces 20 gameplay videos per method
while ensuring that each game contributes equally to the final benchmark
score.

\subsection*{Metric Implementation}

\paragraph{Visual Quality.}
We adopt three automated computer-vision metrics from
VBench~\citep{huang2024vbench,huang2025vbench++}.
\textit{Image quality (IQ)} is computed per frame using the MUSIQ
multi-scale image quality transformer.
\textit{Dynamic degree} is measured via RAFT optical flow:
frames are extracted at 8\,fps; a video is classified as dynamic when the count of
frames exceeding a motion-magnitude threshold meets a count threshold
proportional to clip length.
\textit{Motion smoothness} uses AMT video frame interpolation:
every other frame is removed, AMT reconstructs the intermediate frames.
Higher scores indicate less temporal jitter.

\paragraph{Interactive Fidelity.}
Inspired by WBench~\citep{ying2026wbench}, we evaluate three
interactive-fidelity dimensions using LLM-as-a-Judge with Qwen3.8-Max at
temperature~$0$.
\textit{Setting adherence (Setting)} evaluates whether the generated
environment matches the requested setting.
\textit{Interaction adherence (Inter.)} evaluates whether recorded player
actions produce the specified interactions.
\textit{Physics compliance (Physics)} evaluates physical plausibility of
motions and state transitions across the rollout.

\paragraph{Multimodal Artifact Quality.}
Inspired by V-GameGym~\citep{zhang2026v}, we evaluate generated game artifacts
across three modalities using LLM-as-a-Judge with Qwen3.8-Max at temperature~$0$.
Each modality is scored on four sub-dimensions ($0$--$25$ each, summing to
$0$--$100$).
\textit{Code} uses the source files and game description as input.
\textit{Image} uses screenshots and game description.
\textit{Video} uses the gameplay recording and game description.

\paragraph{Playable-Game Quality.}
Inspired by GameCraft-Bench~\citep{luo2026gamecraft}, we evaluate four
dimensions using LLM-as-a-Judge: core mechanics ($M$), content depth ($D$),
functional visuals ($V$), and art and presentation ($A$).
We use criterion-level rubrics that apply uniformly across the ten diverse
GameCode4D games, with scoring calibration and conditional caps.
The overall playable-game score follows the formula
$\text{Build}/100 \times (0.15M + 0.35D + 0.15V + 0.35A)$,
where $M,D,V,A$ are each normalised to $[0,100]$.

\textit{Build} measures autonomous build-and-run reliability: it is the
average pass rate of standardised engine checks (compilation, project
launch, target-level loading, and predefined-route execution) without human
intervention. All projects reached the executable state within at most five
repair iterations.

\subsection*{LLM-as-a-Judge Protocol}

All LLM-judged metrics use Qwen3.8-Max with temperature set to 0. We use the
same judge configuration for all compared methods. Depending on the evaluated
criterion, the judge receives the game-intent description, ordered
controller-action trace, gameplay video, and the corresponding code or
screenshots. The judge is instructed to score only evidence observable in
the supplied inputs and not to infer unobserved mechanics or interactions.

For each evaluated criterion, we use the following fixed prompt template:

\prompttitle{Evaluation Prompt}
\begin{lstlisting}[style=promptstyle]
You are evaluating a game generated from a short description.

Inputs:
1. Game description
2. Ordered controller-action trace
3. Gameplay video(s)
4. Code/screenshots, if required by the criterion

Criterion: [criterion name]
Rubric: [criterion-specific definition]

Score only evidence visible in the supplied inputs. Do not infer
unobserved mechanics or interactions. Return a score from 0 to 100
and a brief justification. Identify the relevant event, frame, or
artifact. If the required evidence is unavailable, return N/A.

Output:
score: [0--100 or N/A]
evidence: [brief justification]
\end{lstlisting}

\subsubsection*{Interactive Fidelity Prompts}

\prompttitle{Setting Adherence --- User Prompt}
\begin{lstlisting}[style=promptstyle]
Inputs:
1. Game description
2. Gameplay video

Criterion: Setting Adherence

Evaluate the video on two aspects:

[Aspect 1: Environment Maintenance]
How well does the generated game environment match the setting described
in the game description throughout the entire rollout? Consider:
  - Scene type (e.g. forest, urban, underwater, snowy mountain)
  - Lighting conditions and time of day
  - Weather effects and atmospheric properties
  - Overall visual style consistency with the described setting
Anchor: 100 = all described setting elements fully realised and
maintained; 50 = setting partially matches with noticeable deviations
in some elements; 0 = environment is entirely inconsistent with the
described setting.

[Aspect 2: Scene Content Completeness]
As the camera moves through the scene during gameplay, do setting
elements described as part of the game world appear plausibly?
(E.g. if the description mentions forests, enemies, or structures,
do they appear as gameplay unfolds?)
Anchor: this aspect rewards the presence of expected scene content
across the rollout, not just the initial frame.

Combine both aspects proportionally into a single 0-100 score.
Base your score solely on what is directly observable in the video.

Output:
score: [0-100]
evidence: [one or two sentences naming specific frames or elements
that support the score, for both aspects if relevant]
\end{lstlisting}

\prompttitle{Interaction Adherence --- User Prompt}
\begin{lstlisting}[style=promptstyle]
Inputs:
1. Game description
2. Ordered controller-action trace
3. Gameplay video

Criterion: Interaction Adherence

Evaluate whether the player actions in the controller-action trace
produce the interactions specified in the game description.

Answer the following four questions based on the video and trace:

[Q1] Does a core interaction described in the game (e.g. collecting an
item, attacking an enemy, triggering a checkpoint, activating an event)
visually occur in the video?
Anchor: Yes = the interaction type is recognisably present; No = no
described interaction type is observable.

[Q2] When the player performs an input corresponding to a described
interaction, does the game respond with the expected consequence?
(Use the trace to identify intended inputs.)
Anchor: Yes = input-to-consequence link is visible; No = inputs produce
no observable response or the wrong consequence.

[Q3] Are the key entities and mechanics involved in the interaction
correct? (E.g. correct enemy type, correct item, correct trigger zone.)
Anchor: Yes = entities and mechanics match the description;
No = clearly wrong entities or mechanics are used.

[Q4] Across the full rollout, what fraction of the described
interaction types are successfully realised?
Anchor: All = 100; More than half = ~70; About half = ~50;
Less than half = ~30; None = 0.

Combine all four answers into a single 0-100 score.

Output:
score: [0-100]
evidence: [one or two sentences identifying the specific interaction
events observed and any failures noted]
\end{lstlisting}

\prompttitle{Physics Compliance --- User Prompt}
\begin{lstlisting}[style=promptstyle]
Inputs:
1. Gameplay video

Criterion: Physics Compliance

You are evaluating the physical plausibility and causal consistency
of the game video. Scan ALL frames for violations of the following:

[A. Rendering Physics]
  1. Solid body integrity: no character or object clipping through
     surfaces, walls, or other objects
  2. Gravity and fall behaviour: objects and characters fall and
     land naturally; no floating without explanation
  3. Motion continuity: no teleportation or discontinuous jumps
     in position between frames
  4. Object permanence: objects do not randomly appear or vanish
     without a gameplay cause
  5. Character locomotion: movement is biomechanically plausible;
     no jerky or puppet-like animation

[B. Causal Consistency]
  Note: instructed player actions and their direct consequences are
  expected and should NOT be penalised. Only penalise unintended
  changes.
  6. No effect without a visible cause for non-instructed changes
  7. No cause without a visible effect for player-initiated actions

Scoring anchors:
  100 = physics and causality are natural and consistent throughout;
        no violations detected in any frame
   67 = 1-2 noticeable violations (brief clipping, minor float, etc.)
   33 = multiple violations or one severe breakdown in a major sequence
    0 = physics or causality mostly broken throughout the video

ANY violation, however brief or localised, should lower the score.
Do NOT penalise camera behaviour (shake, pan, zoom).

Output:
score: [0-100]
evidence: [name the specific violation type, frame/moment, and
severity; if no violations found, state that explicitly]
\end{lstlisting}

\subsubsection*{Multimodal Artifact Quality Prompts}

\vspace{4pt}

\prompttitle{Code Quality --- User Prompt}
\begin{lstlisting}[style=promptstyle]
Inputs:
1. Game description
2. Generated source code

Criterion: Code Quality

Evaluate whether the generated source code meets the game description.
Score each sub-dimension from 0 to 25:

  1. Functionality (0-25): Are all mechanics described in the
     requirements structurally implemented? (game loop, input
     handling, core interactions, win/lose conditions)

  2. Code Quality (0-25): Code structure, readability, and
     consistency. Are identifiers meaningful? Is logic well
     organised? Are there obvious structural gaps?

  3. Game Logic (0-25): Is the game logic complete and reasonable?
     Does state management, collision, scoring, and event handling
     behave as the description intends?

  4. Technical Implementation (0-25): Are the game engine APIs
     used correctly? Are the key algorithms and data structures
     appropriate for the described game?

Return JSON in exactly this shape (no extra keys, no markdown):
{
  "functionality_score": <0-25>,
  "code_quality_score": <0-25>,
  "game_logic_score": <0-25>,
  "technical_score": <0-25>,
  "total_score": <0-100>,
  "evidence": "<one sentence summary of the main strengths or gaps>"
}
\end{lstlisting}

\prompttitle{Image Quality --- User Prompt}
\begin{lstlisting}[style=promptstyle]
Inputs:
1. Game description
2. Screenshots captured from the generated game (up to 20 images)

Criterion: Image Quality

Evaluate whether the screenshots demonstrate that the generated game
meets the description. Score each sub-dimension from 0 to 25:

  1. Visual Completeness (0-25): Is the game screen clear and
     complete? No missing textures, black regions, broken meshes,
     or rendering errors.

  2. UI Design (0-25): UI layout, readability, and visual quality.
     Are HUD elements legible and coherently designed?

  3. Functionality Display (0-25): Do the screenshots show the key
     features and interactions described in the requirements? Do
     they evidence that the described game type is present?

  4. Overall Visual Quality (0-25): Overall visual polish and
     completion of the game as seen in these screenshots.

Return JSON in exactly this shape (no extra keys, no markdown):
{
  "visual_completeness_score": <0-25>,
  "ui_design_score": <0-25>,
  "functionality_display_score": <0-25>,
  "overall_quality_score": <0-25>,
  "total_score": <0-100>,
  "evidence": "<one sentence summary of the main visual strengths
               or gaps observed across the screenshots>"
}
\end{lstlisting}

\prompttitle{Video Quality --- User Prompt}
\begin{lstlisting}[style=promptstyle]
Inputs:
1. Game description
2. Gameplay video recording

Criterion: Video Quality

Evaluate whether the gameplay video demonstrates that the generated
game meets the description. Score each sub-dimension from 0 to 25:

  1. Animation Effect (0-25): Are character and object animations
     smooth and natural? Are transitions between states fluid?

  2. Interaction Logic (0-25): Do player interactions produce
     correct and timely responses in the video? (e.g. character
     moves on input, enemies react when hit, items are picked up)

  3. Game Flow (0-25): Is the overall gameplay flow complete and
     coherent? Does the session progress logically from start
     through core interactions toward a goal state?

  4. Dynamic Quality (0-25): Overall temporal quality: no
     flickering, abrupt scene changes, frame-rate drops, or
     rendering instability that disrupts the play experience.

Return JSON in exactly this shape (no extra keys, no markdown):
{
  "animation_score": <0-25>,
  "interaction_score": <0-25>,
  "gameplay_flow_score": <0-25>,
  "dynamic_quality_score": <0-25>,
  "total_score": <0-100>,
  "evidence": "<one sentence summary of the main dynamic strengths
               or issues observed in the video>"
}
\end{lstlisting}

\subsubsection*{Playable-Game Quality Prompts}

\noindent\textbf{Scoring calibration (all four dimensions):}
\begin{itemize}
\item \textbf{100}: Feature fully implemented and demonstrated with
publishable quality; a player would not notice anything missing.
\item \textbf{50}: Feature exists and functions but uses placeholder
assets, minimal content, or lacks polish; it works but would not ship.
\item \textbf{0}: Feature is absent, broken, or purely decorative with
no gameplay effect.
\end{itemize}

\noindent\textbf{General cap conditions:}
\begin{itemize}
\item Score at most 50 if visible gameplay relies on untextured geometry,
missing meshes, or default engine widgets with no authored art.
\item Score at most 50 if the feature is demonstrated only in a scripted
or static moment with no sustained interactive gameplay.
\item Score 100 only if a visible player action causes a state change
that persists and is reflected in the game world.
\end{itemize}

\vspace{4pt}

\prompttitle{Core Mechanics ($M$) --- User Prompt}
\begin{lstlisting}[style=promptstyle]
Inputs:
1. Game description
2. Ordered controller-action trace
3. Gameplay video

Criterion: Core Mechanics

The core mechanic is the irreducible gameplay loop described in the
game description: the player action, the state change it causes, the
rules governing it, and the resulting consequence.

Evaluate:
  - Is the core mechanic present and triggerable by the player?
  - Does a visible player action cause a persistent state change
    (e.g. collecting an item removes it and updates inventory,
    hitting an enemy reduces its health, crossing a checkpoint
    records progress)?
  - Is the mechanic consistently functional across the rollout?
  - Is the feedback (visual, HUD update, or audio cue) clear and
    unambiguous after each interaction?

Cap conditions:
  - Score at most 50 if the mechanic works but is demonstrated only
    in a scripted sequence, not via player input.
  - Score at most 50 if the mechanic triggers but produces no
    persistent state change observable in the video.
  - Score 100 only if the full loop (action -> state change ->
    consequence -> feedback) is clearly demonstrated in the rollout.

Output:
score: [0-100]
evidence: [identify the specific mechanic, the player action that
triggers it, and the state change observed in the video]
\end{lstlisting}

\prompttitle{Content Depth ($D$) --- User Prompt}
\begin{lstlisting}[style=promptstyle]
Inputs:
1. Game description
2. Ordered controller-action trace
3. Gameplay video

Criterion: Content Depth

Evaluate whether the game provides meaningful objectives, progression,
and interactive content beyond a single interaction loop.

Assess the following dimensions:
  - Variety: are there multiple content types (enemies, items,
    obstacles, zones, or level segments) that are visually and
    functionally distinct, not merely colour or name variations?
  - Progression: are visible progress indicators present (score,
    timer, checkpoints, inventory, health, stage counter)?
  - Objectives: are one or more objectives communicated to the
    player and verifiably pursued during the rollout?
  - Late-game or escalation content: does the rollout show any
    change in difficulty, environment, or content variety over time?

Cap conditions:
  - Score at most 50 if content variants differ only by colour,
    name, or stat number with identical visual appearance.
  - Score at most 50 if only a single objective or interaction
    type is present throughout the rollout.
  - Score 100 only if at least two functionally distinct content
    types are demonstrated with clear progression indicators.

Output:
score: [0-100]
evidence: [list the distinct content types, objectives, and
progression signals observed in the video]
\end{lstlisting}

\prompttitle{Functional Visuals ($V$) --- User Prompt}
\begin{lstlisting}[style=promptstyle]
Inputs:
1. Game description
2. Gameplay video

Criterion: Functional Visuals

Evaluate whether visual assets, interface elements, and feedback
signals effectively communicate gameplay state at all times.

Assess at 1280x720 or native resolution:
  - HUD readability: are health, score, timer, inventory, and
    other essential state indicators legible and stably positioned?
  - Interactive object clarity: can enemies, collectibles, hazards,
    and interaction targets be identified clearly from the video?
  - Action feedback: after player actions, is there visible feedback
    (hit effects, pickup animations, state-change indicators)?
  - Absence of UI issues: no overlapping panels, missing textures
    on key objects, invisible geometry, or unreadable text.

Cap conditions:
  - Score at most 50 if HUD elements are present but only as
    plain text labels with no visual design or spatial separation.
  - Score at most 50 if interactive objects are indistinguishable
    from background geometry without authoring cues.
  - Score 100 only if a player can understand their resources,
    current state, and next action without guessing.

Output:
score: [0-100]
evidence: [identify specific HUD elements, feedback events, or
clarity issues observed in the video]
\end{lstlisting}

\prompttitle{Art and Presentation ($A$) --- User Prompt}
\begin{lstlisting}[style=promptstyle]
Inputs:
1. Gameplay video 

Criterion: Art and Presentation

Evaluate the overall visual coherence, completeness, and presentation
quality of the game as a playable artifact. The bar is a game that
reads as intentionally designed rather than debug-like.

Assess:
  - Asset consistency: do all visible assets (characters, props,
    terrain, UI) share a coherent visual style?
  - Lighting quality: is the scene lighting intentional,
    atmospheric, and free from unlit patches or default lighting?
  - Absence of placeholder content: no raw untextured geometry,
    default engine shapes, mismatched asset packs, or debug labels
    visible during normal gameplay.
  - Scene composition: does the environment read as an authored
    scene with deliberate placement of elements?

Cap conditions:
  - Score at most 50 if any significant portion of the visible
    scene uses solid-colour fills, untextured meshes, or clearly
    default engine assets.
  - Score at most 50 if visual style is inconsistent across
    different parts of the scene (e.g. realistic terrain with
    cartoon props, mixed resolutions).
  - Score 100 only if the game reads as a publishable vertical
    slice with no obvious placeholder elements.

Output:
score: [0-100]
evidence: [describe specific assets, scenes, or visual elements
that support the score, noting any placeholder content observed]
\end{lstlisting}

\subsection*{User-Study Protocol}

We conduct a human evaluation using the same playable-game dimensions used
for automatic evaluation. Method identities are hidden from participants,
and gameplay videos are presented in randomized order. The two recordings
from the same generated game are placed in different rating blocks to reduce
direct comparison and ordering effects.

A total of 15 raters participate in the study. For each evaluated game,
participants assign scores from 0 to 100 for core mechanics ($M$), content
depth ($D$), functional visuals ($V$), and art and presentation ($A$).
The Human score reported in the main results aggregates these ratings across
participants and benchmark games.

\begin{table}[ht]
\caption{User-study rubric. Each dimension is scored from 0 to 100, using the
same playable-game criteria as the automatic evaluation.}
\label{tab:user_rubric}
\centering
\footnotesize
\setlength{\tabcolsep}{3pt}
\renewcommand{\arraystretch}{1.05}

\begin{tabularx}{\columnwidth}{
>{\raggedright\arraybackslash}p{0.27\columnwidth}X
}
\toprule
Dimension & Rating statement \\
\midrule

Core mechanics ($M$) &
The central player action and its consequence are clear and functional. \\

Content depth ($D$) &
The world, objectives, and interactions provide meaningful progression. \\

Functional visuals ($V$) &
Visual assets, interface, and feedback support readable gameplay. \\

Art and presentation ($A$) &
The game appears coherent, complete, and engaging as a demonstration. \\

\bottomrule
\end{tabularx}
\end{table}

\FloatBarrier

\section{Additional Qualitative Results}
\label{app:qualitative_results}

We provide additional qualitative results across diverse GameCode4D scenarios.
Representative gameplay frames are shown in Figures~\ref{fig:gameplay_01}--\ref{fig:gameplay_20}.

\newcommand{\keyframepanel}[3]{%
  \begin{figure}[H]
    \centering
    \includegraphics[
      width=\linewidth,
      height=0.38\textheight,
      keepaspectratio
    ]{figure/appendix/#1}%
    \caption{#2}
    \label{fig:gameplay_#3}
    \vspace{-6pt}
  \end{figure}%
}

\keyframepanel{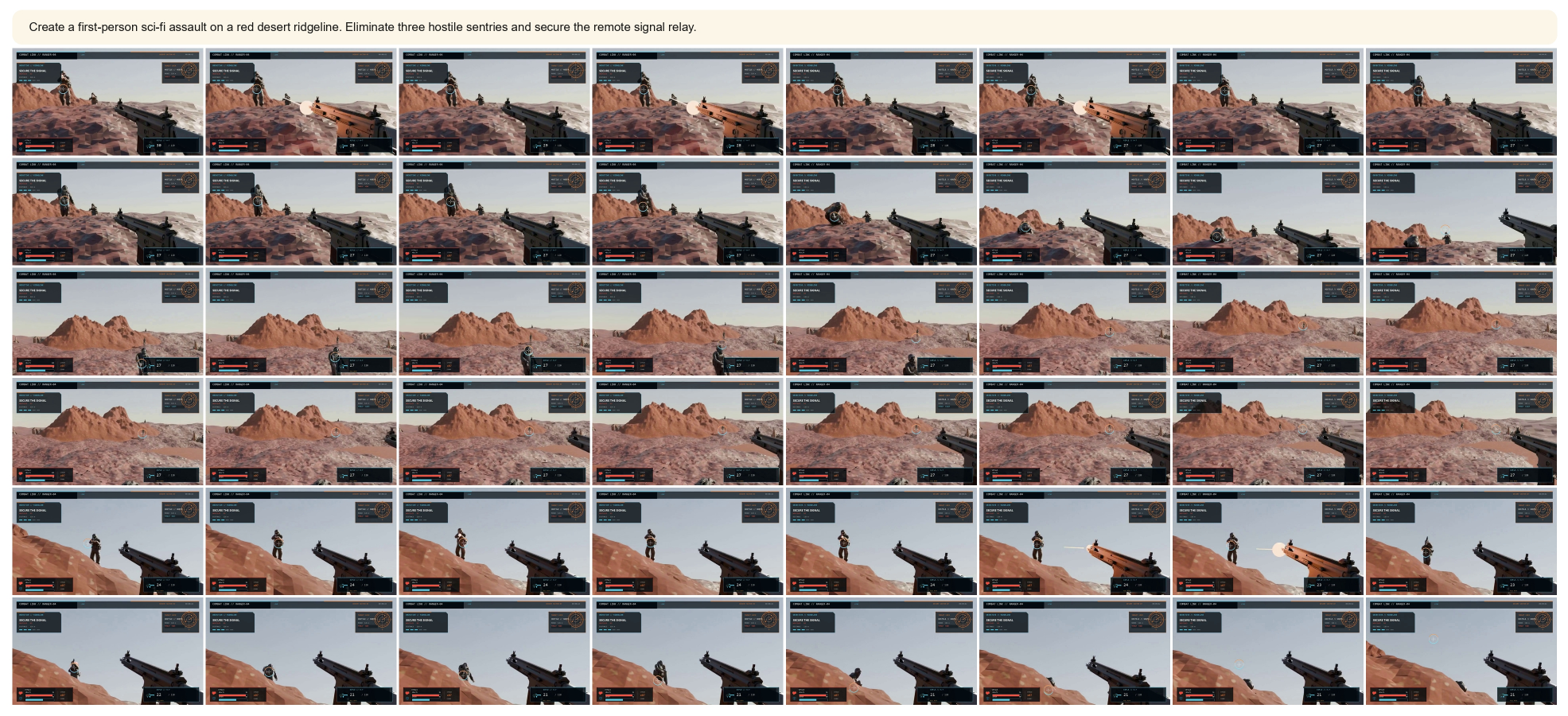}{FPS gameplay example 1.}{01}
\keyframepanel{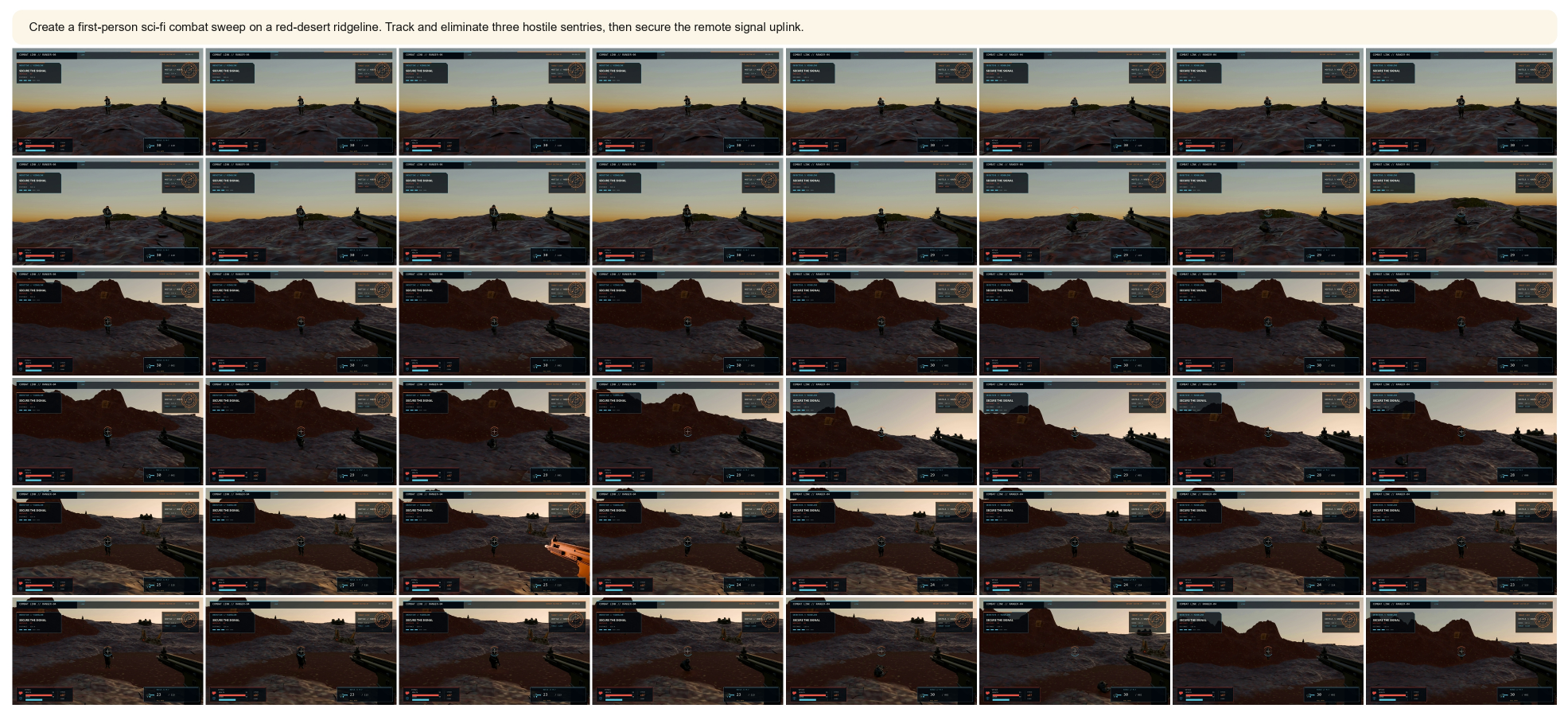}{FPS gameplay example 2.}{02}
\keyframepanel{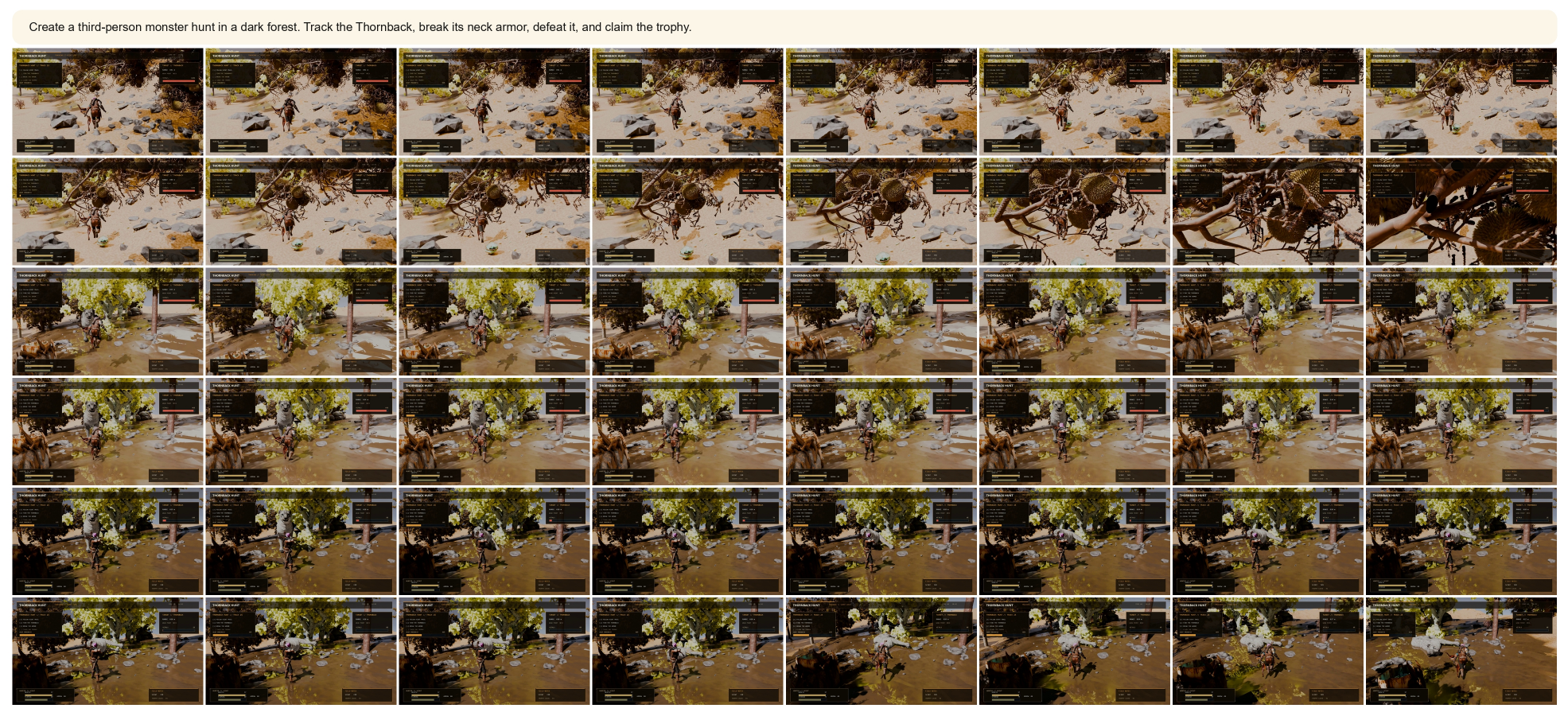}{Monster-hunt gameplay example 1.}{03}
\keyframepanel{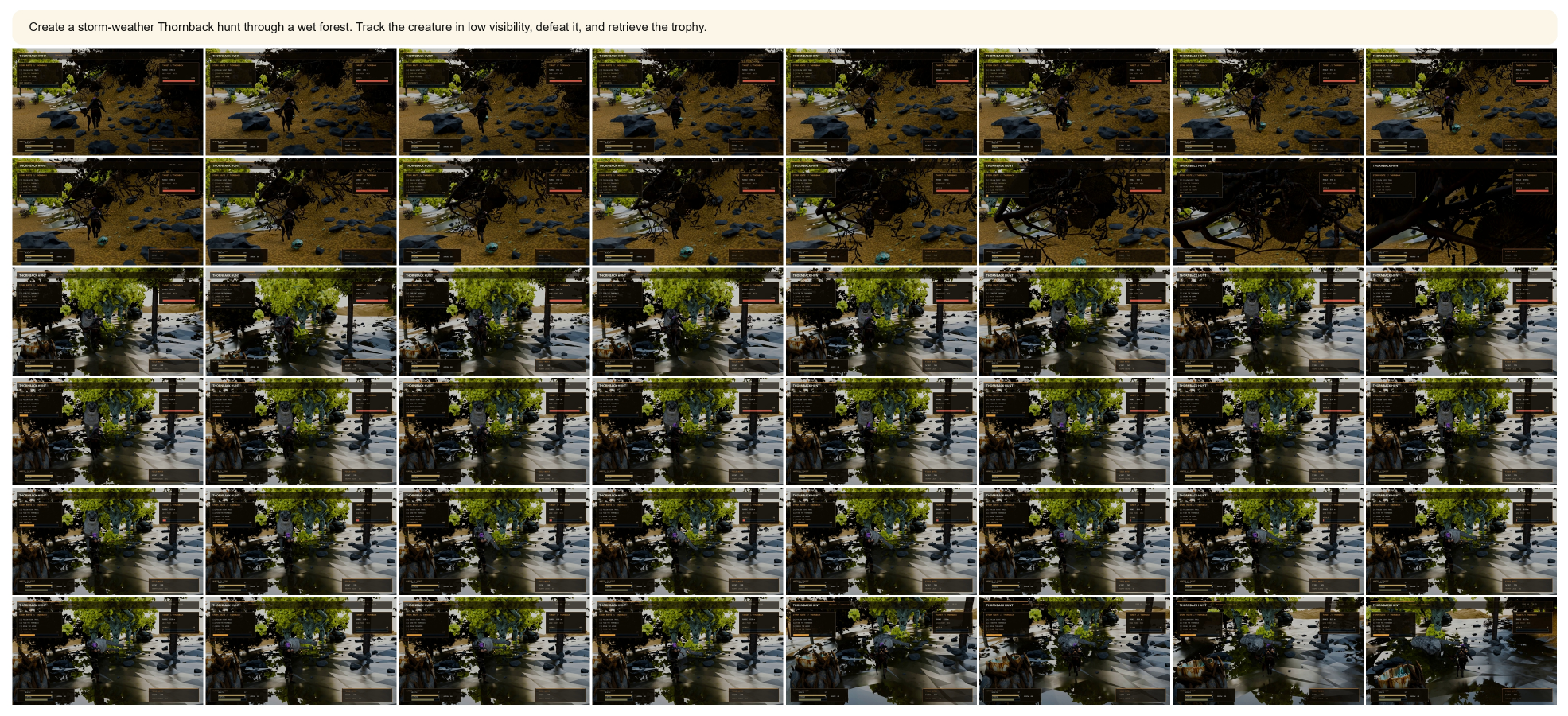}{Monster-hunt gameplay example 2.}{04}
\keyframepanel{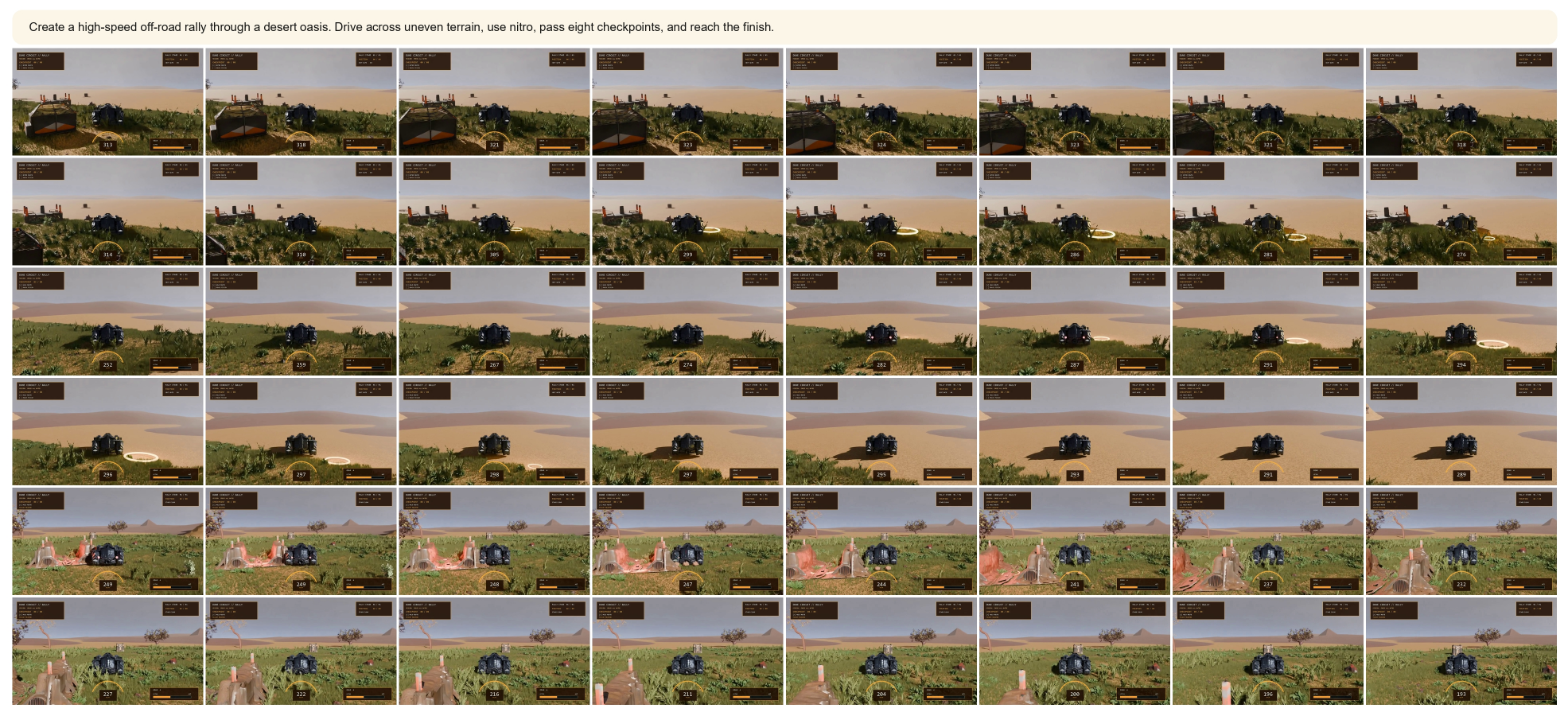}{Racing gameplay example 1.}{05}
\keyframepanel{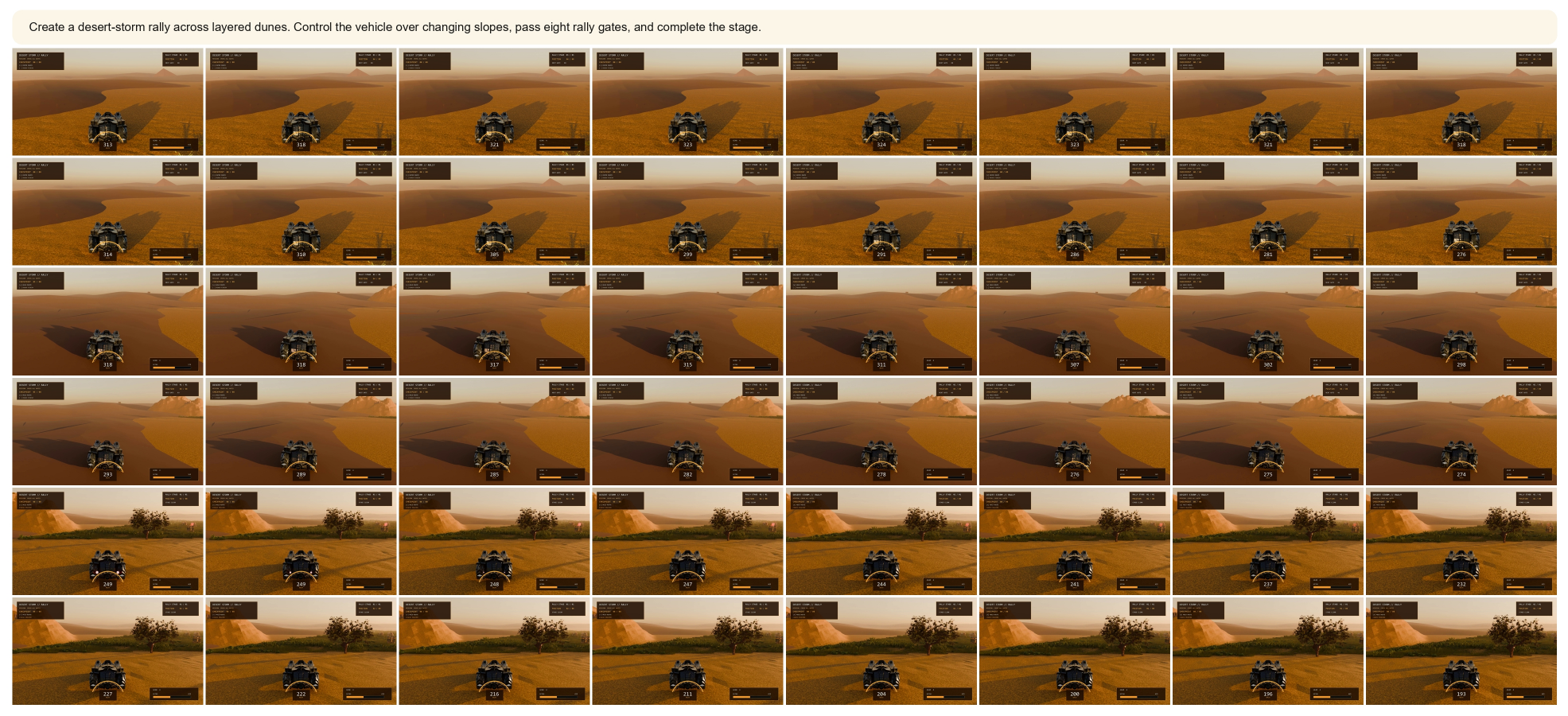}{Racing gameplay example 2.}{06}
\keyframepanel{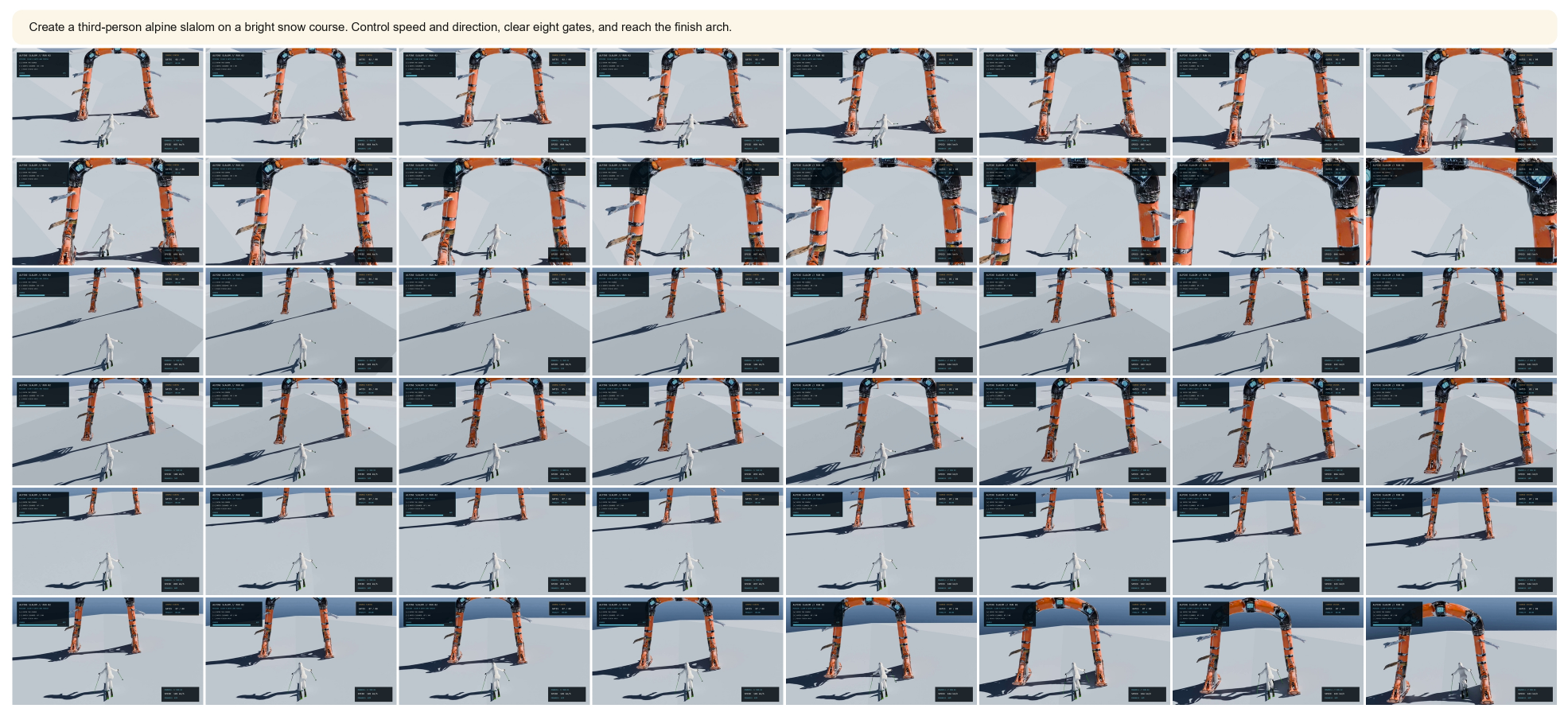}{Skiing gameplay example 1.}{07}
\keyframepanel{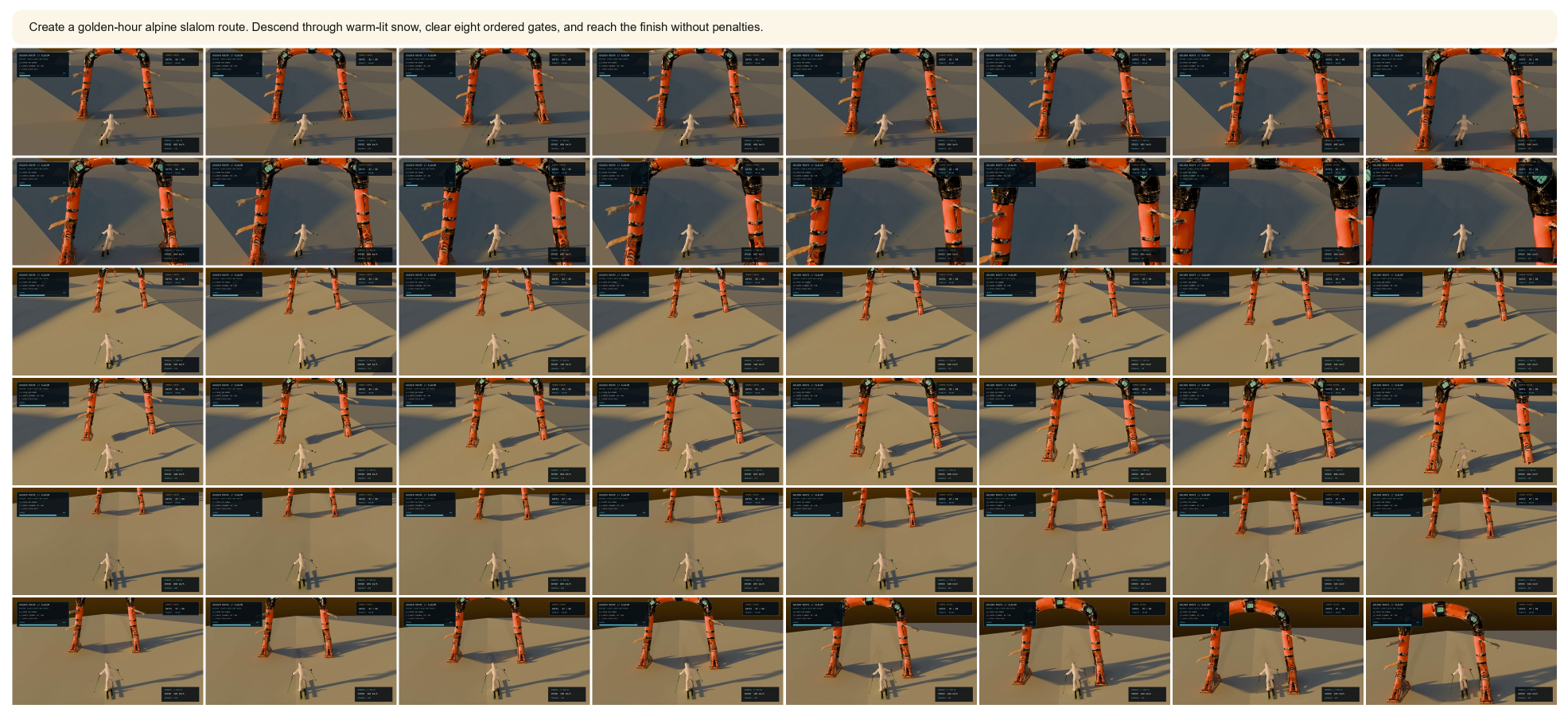}{Skiing gameplay example 2.}{08}
\keyframepanel{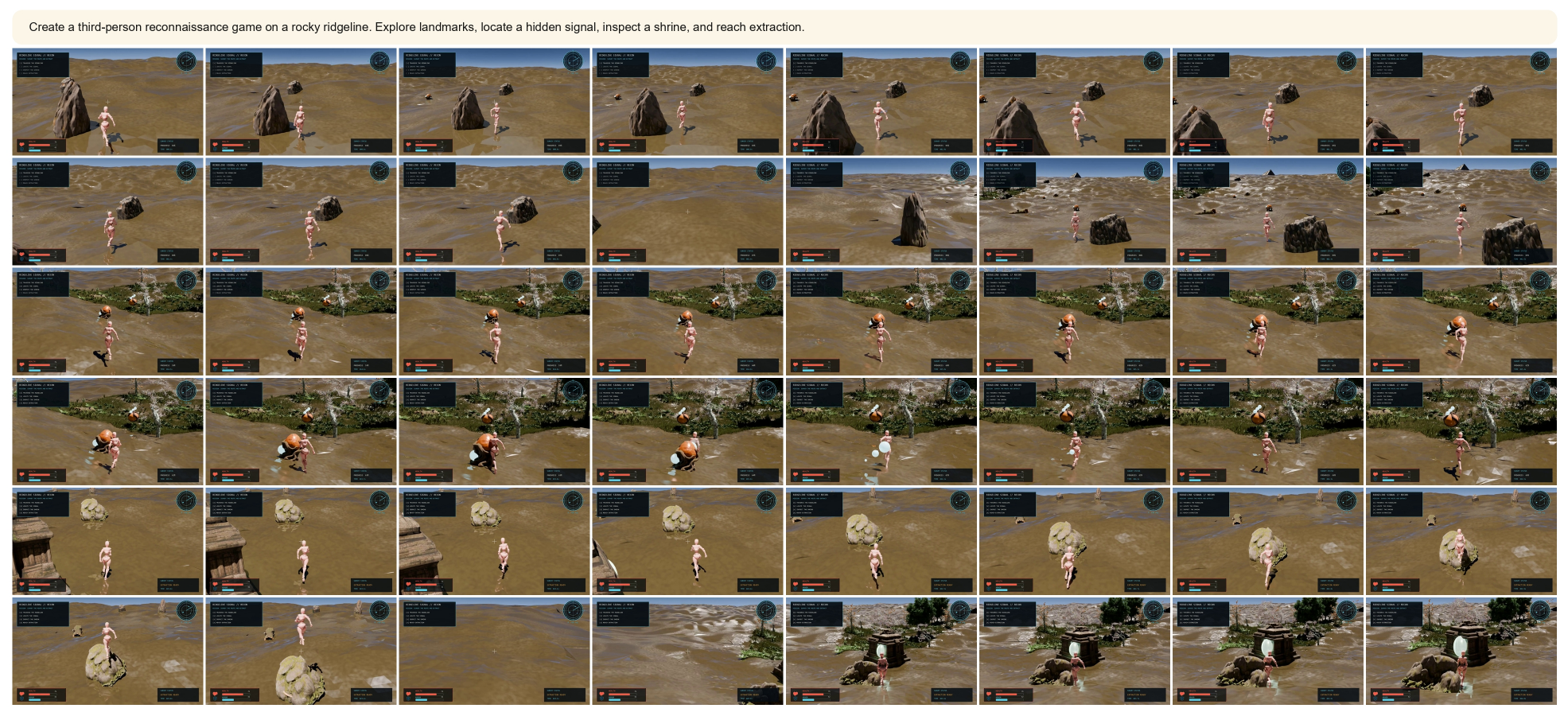}{Temple-run gameplay example 1.}{09}
\keyframepanel{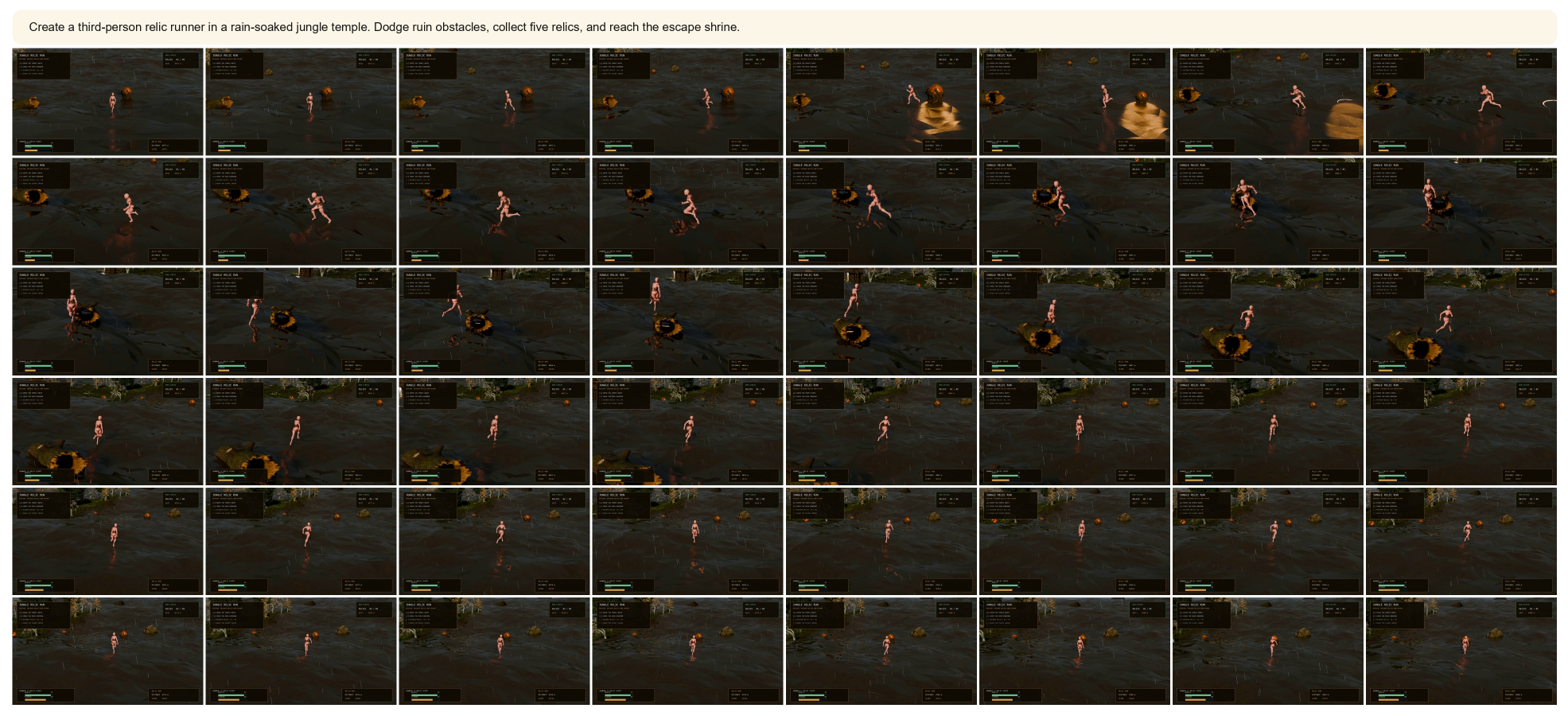}{Temple-run gameplay example 2.}{10}
\keyframepanel{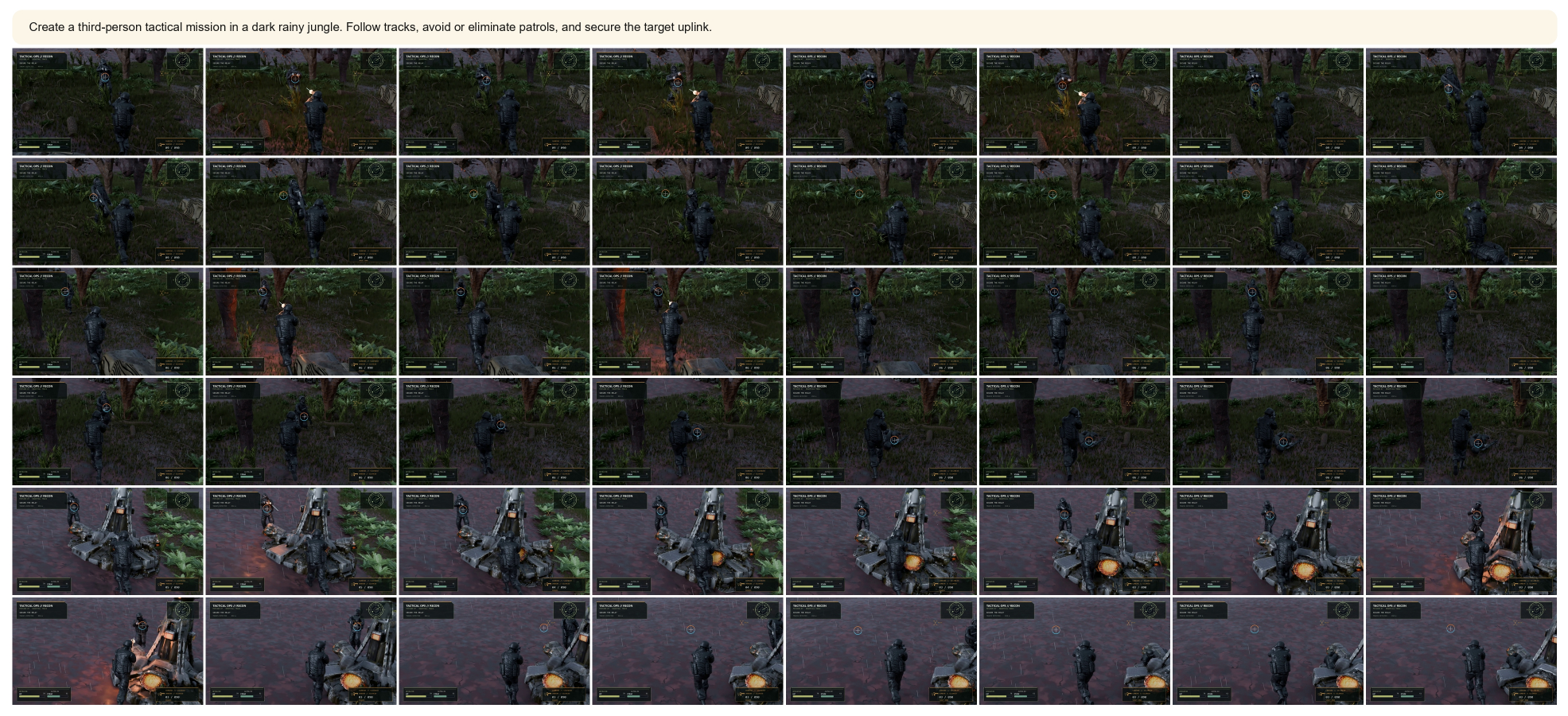}{Third-person shooter gameplay example 1.}{11}
\keyframepanel{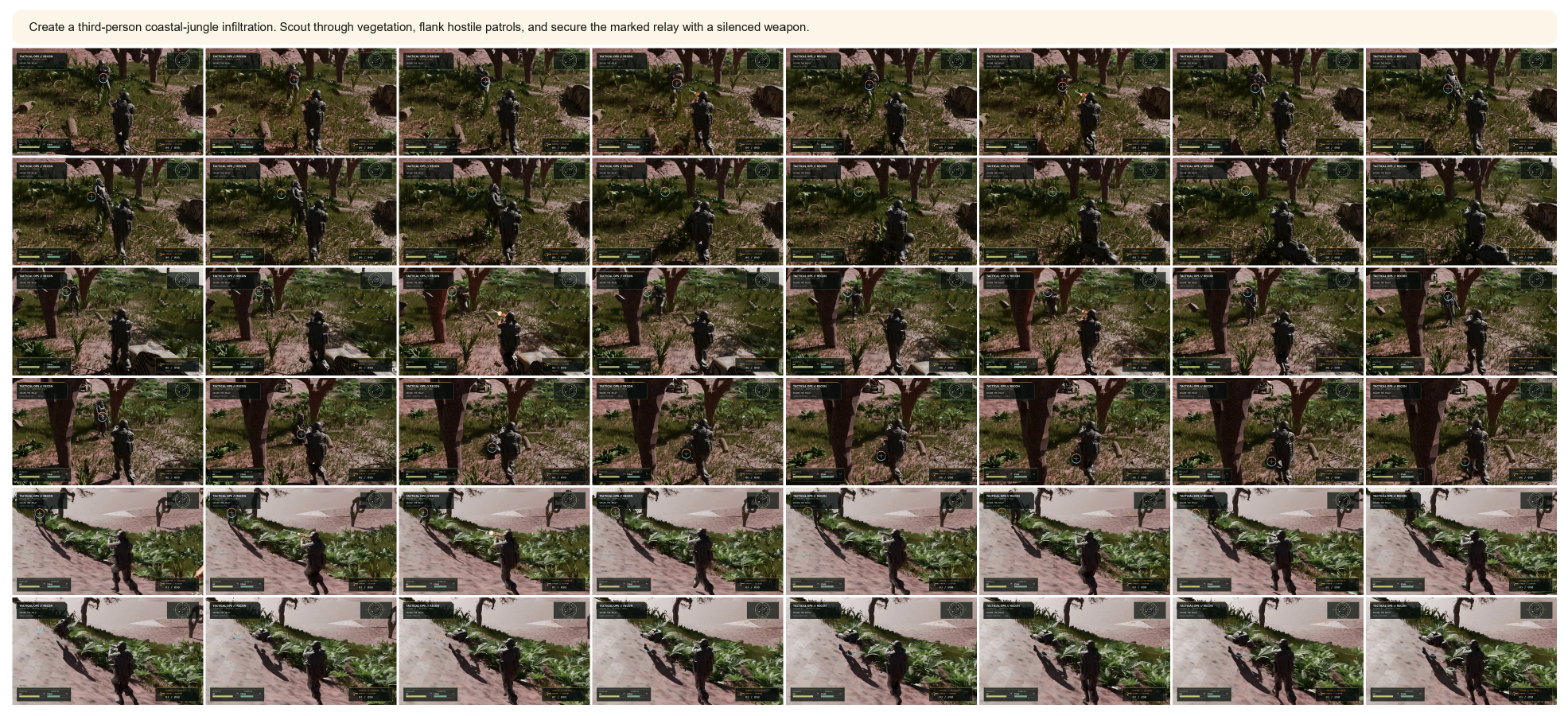}{Third-person shooter gameplay example 2.}{12}
\keyframepanel{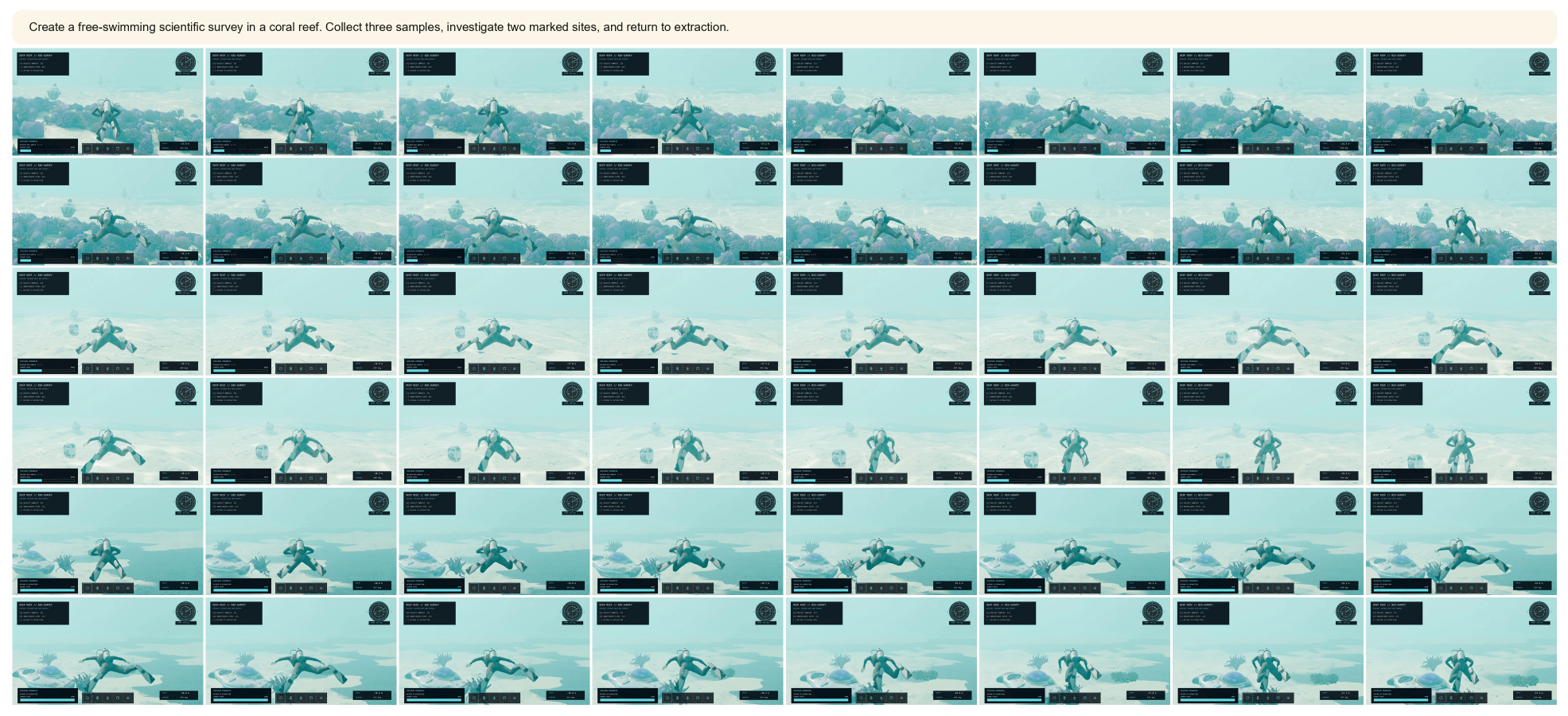}{Underwater exploration gameplay example 1.}{13}
\keyframepanel{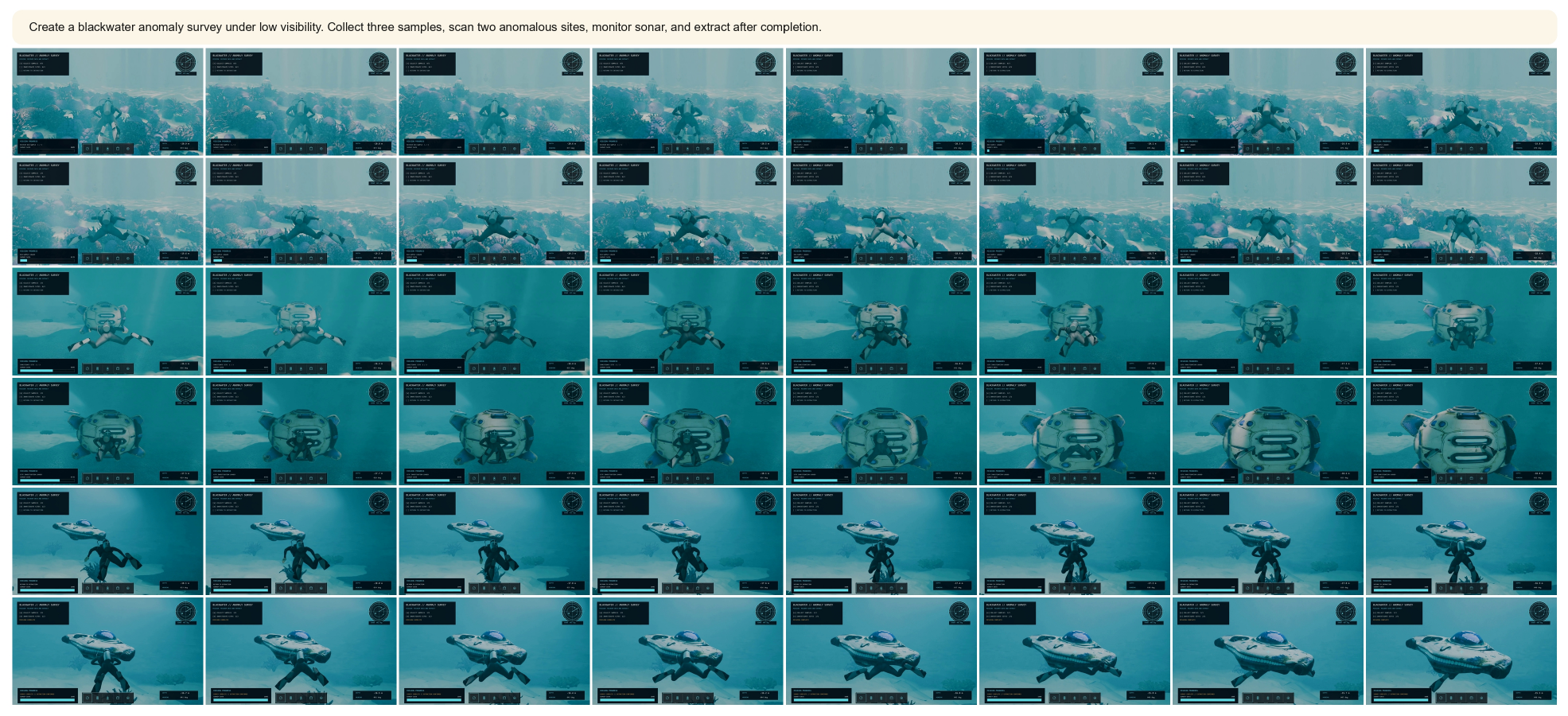}{Underwater exploration gameplay example 2.}{14}
\keyframepanel{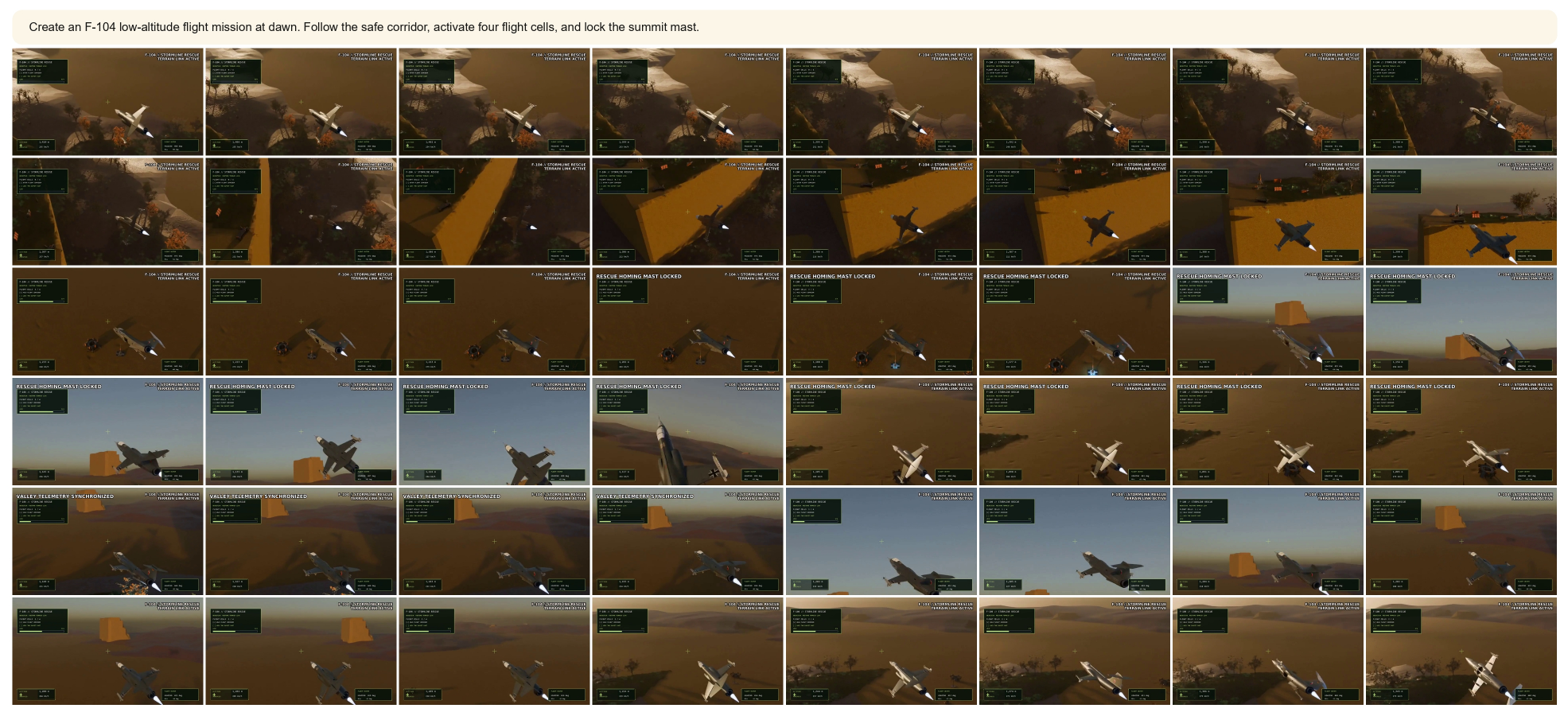}{Wingsuit gameplay example 1.}{15}
\keyframepanel{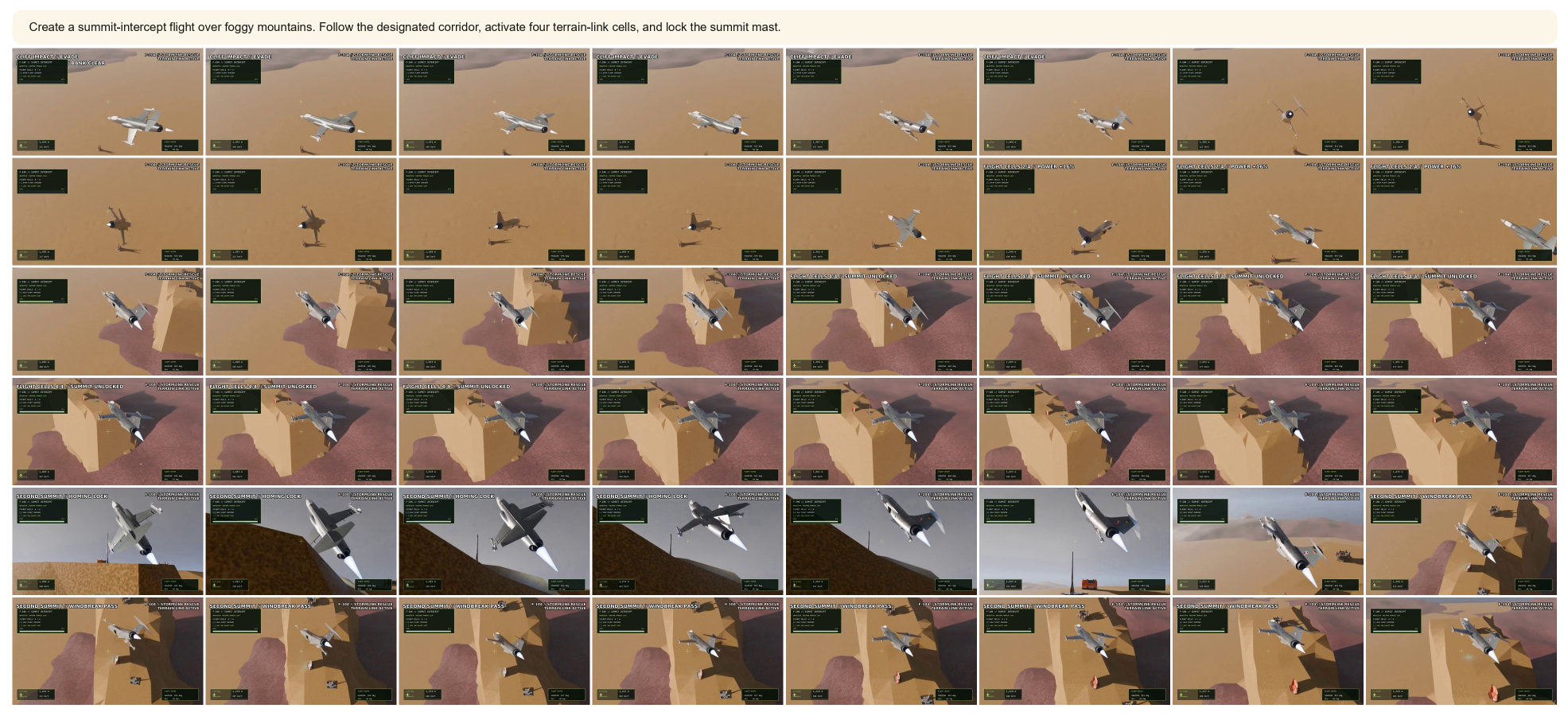}{Wingsuit gameplay example 2.}{16}
\keyframepanel{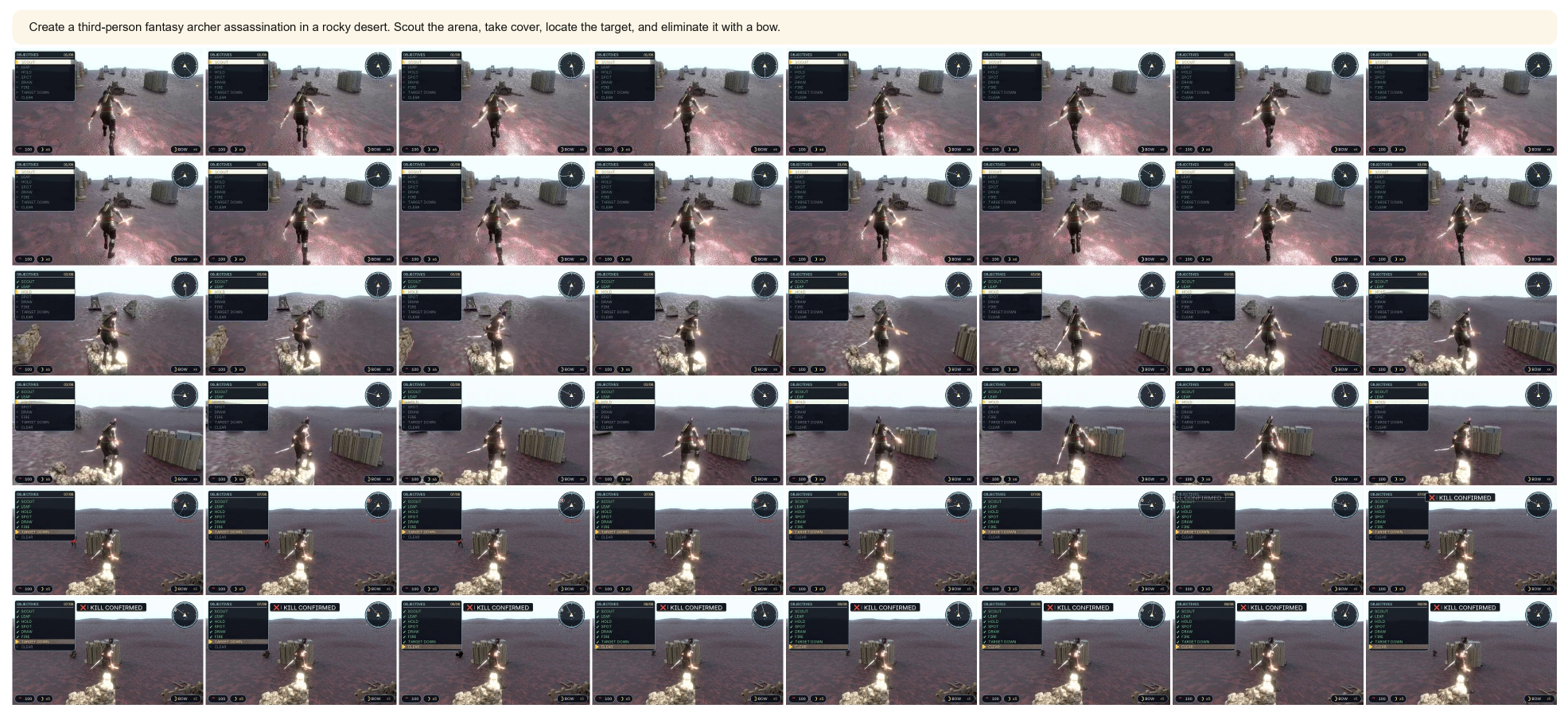}{Assassin's Creed-style gameplay example 1.}{17}
\keyframepanel{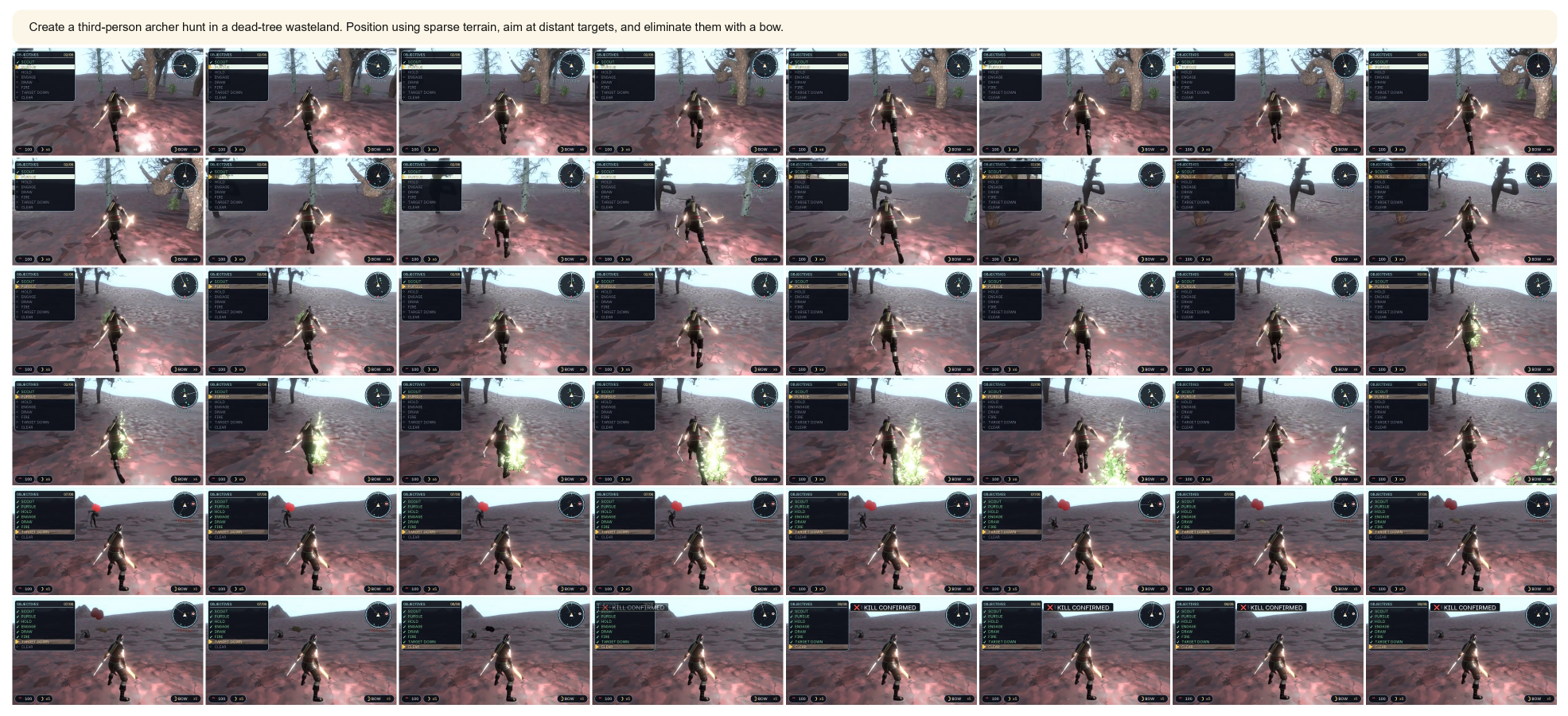}{Assassin's Creed-style gameplay example 2.}{18}
\keyframepanel{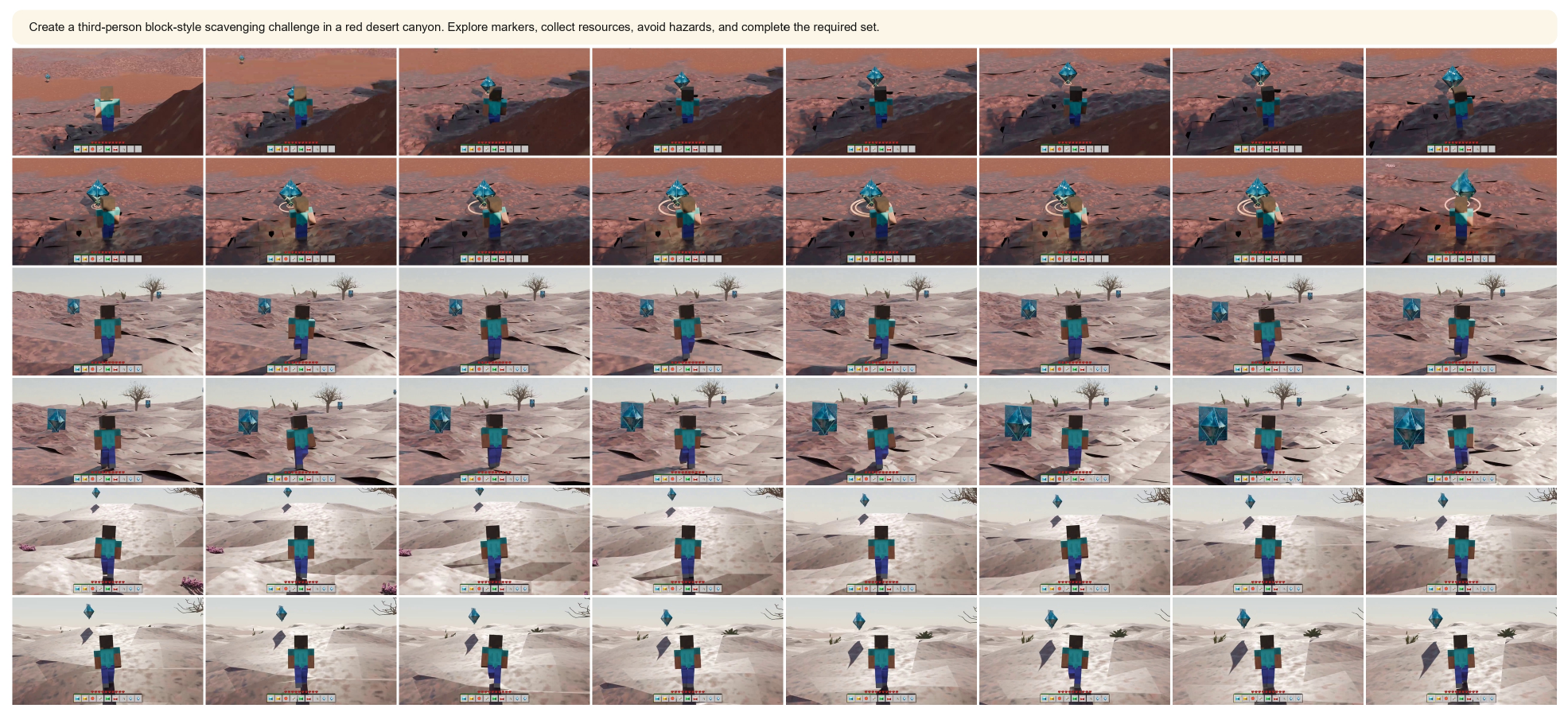}{Minecraft-style gameplay example 1.}{19}
\keyframepanel{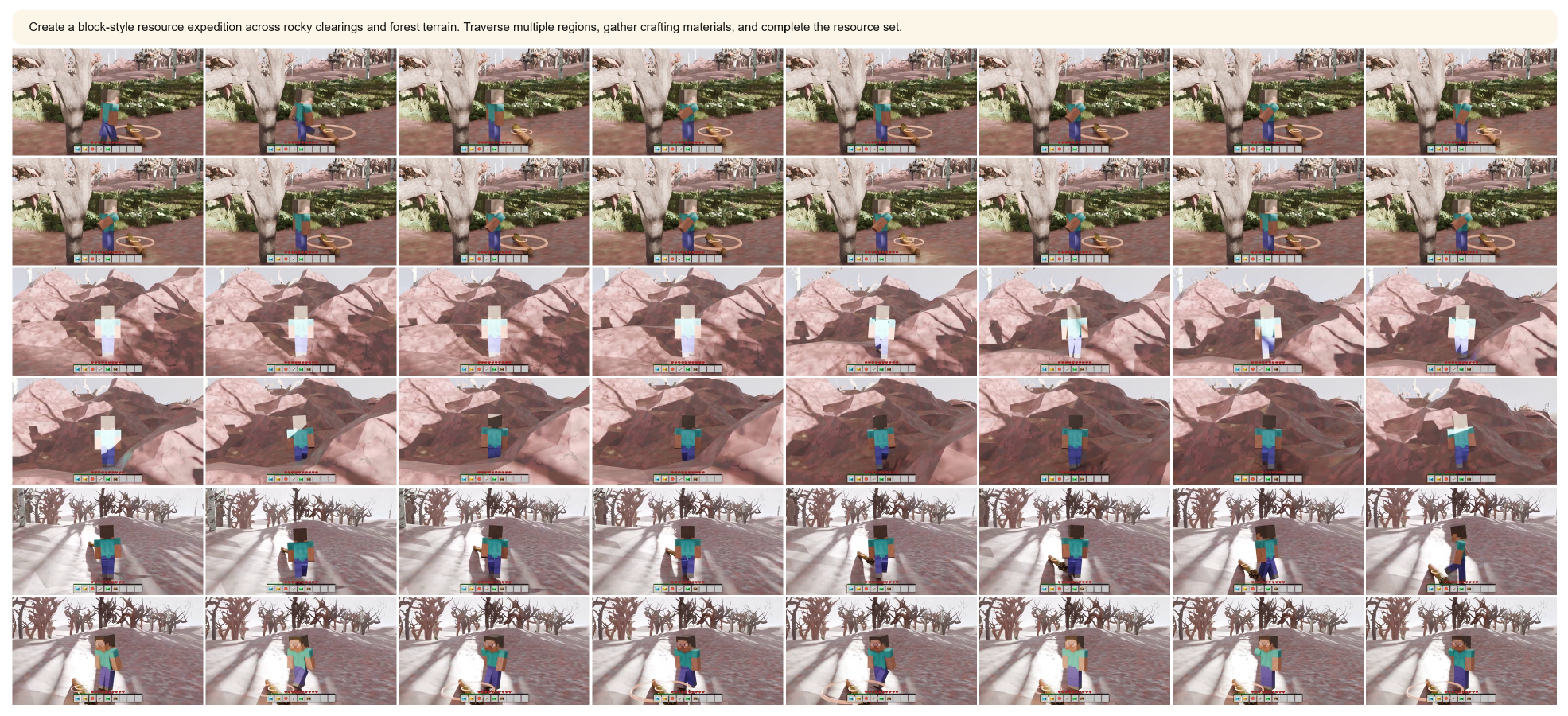}{Minecraft-style gameplay example 2.}{20}
\FloatBarrier

\section{System Prompt Design}
\label{app:system_prompt_design}

We provide the complete system prompts used by the Code2Games agents to support
reproducibility and clarify the responsibilities of different stages in the
generation pipeline. These prompts cover scene understanding, gameplay design,
route planning and repair, scene-constrained candidate selection, asset
realization, reference-image generation, and LLM-based evaluation. The complete
prompt templates are provided in the supplementary material below.

\section*{System Prompts}

\subsection*{Objective Scene Analysis}

\prompttitle{System Prompt}
\begin{lstlisting}[style=promptstyle]
You are a strict JSON DEFAULT_CAMERA objective visual scene understanding agent. Return only valid JSON. Describe visible scene content only. Do not design games, rules, routes, coordinate points, timing, scripts, or assets.
\end{lstlisting}

\prompttitle{User Prompt}
\begin{lstlisting}[style=promptstyle]
# Role
You are the DEFAULT_CAMERA objective visual scene understanding agent for Code2Games / Code2Worlds.
# Project Boundary
1. The current stage only does objective visual scene understanding.
2. Describe only what is actually visible in the current DEFAULT_CAMERA image, supported by metadata.
3. Do not design a game, rules, tasks, routes, points, trajectories, placement, scripting, or follow-up actions.
4. Do not output coordinate point fields or world-space target points.
5. Do not split the image into fixed near/middle/far narrative fields.
6. Describe visual elements, surface types, spatial relations, visible continuity, discontinuity, and uncertainty.
# User Prompt
1. The user prompt is only a goal/style reference. It is not evidence of what is visible.
2. Prioritize the current image and metadata over the prompt. If the image shows water, mud, rock, vegetation, or slopes, describe those real visual contents.
<USER_PROMPT>
# Duration Seconds
This value is passed for downstream context only. Do not use it to design motion or timing.
<DURATION_SECONDS>
# Inputs
1. default_camera_view_grid.png
   The DEFAULT_CAMERA render with a light 3x3 screen grid.
2. default_camera_metadata.json
   Camera parameters, 3x3 grid raycast hints, and coarse local scene hints.
# Task
1. Produce a detailed but compact objective visual understanding of the DEFAULT_CAMERA view.
2. State whether the image is usable as a visual-understanding input.
Identify the overall visual environment type, terrain structure, visible elements, surface distribution, spatial relations, visual openness, visual blockage or complexity, visual discontinuity, guidance features, and uncertainty.
# Important Visual Reasoning Constraints
1. Metadata is only auxiliary. If the image clearly shows water, breaks, occlusion, or complex terrain, do not treat a whole region as clear just because a grid-cell center raycast hits ground.
2. Use screen grid area names when describing regions: upper_left, upper_center, upper_right, middle_left, middle_center, middle_right, lower_left, lower_center, lower_right.
3. Do not say the character should run forward. Do not infer the user's requested style as visible content.
# Output Rules
1. Return only strict JSON. Do not output markdown.
2. selected_view must be DEFAULT_CAMERA.
3. confidence must be a number from 0.0 to 1.0.
4. The top-level JSON must contain only view_status and scene_visual_context.
# Required JSON Schema
```json
{
  "view_status": {
    "selected_view": "DEFAULT_CAMERA",
    "usable": true,
    "confidence": 0.85,
    "reason": ""
  },
  "scene_visual_context": {
    "overall_summary": "",
    "environment_visual_type": "",
    "terrain_structure": "",
    "visible_elements": [],
    "surface_distribution": {
      "ground_regions": [],
      "water_regions": [],
      "vegetation_regions": [],
      "rock_or_slope_regions": [],
      "uncertain_regions": []
    },
    "spatial_relations": [],
    "traversability_visual_observation": {
      "visually_open_regions": [],
      "visually_blocked_or_complex_regions": [],
      "visually_discontinuous_regions": []
    },
    "visual_guidance_features": [],
    "uncertainty_notes": []
    }
}
```
# default_camera_metadata.json
```json
<DEFAULT_CAMERA_METADATA_JSON>
```
\end{lstlisting}

\prompttitle{Image-Input Prompt}
\begin{lstlisting}[style=promptstyle]
Image: default_camera_view_grid.png
\end{lstlisting}

\subsection*{Rule-Relevant Visual Reading}

\prompttitle{System Prompt}
\begin{lstlisting}[style=promptstyle]
You are a strict JSON rule_visual_reader. Return visual notes only. Do not design rules, routes, points, coordinates, or code.
\end{lstlisting}

\prompttitle{User Prompt}
\begin{lstlisting}[style=promptstyle]
# Role
You are rule_visual_reader for Code2Games.
You are a visual model. Your only job is to extract rule-relevant visual notes from the DEFAULT_CAMERA image.
You are not the game rule designer, not placement_agent, not a path planner, and not a script writer.

# Inputs
1. default_camera_view_grid.png
2. default_camera_metadata.json
3. scene_understanding_default_camera.json
4. user_prompt as broad context only

# Task
1. Summarize only visual information that may be useful for later game rule design.
2. Do not design game goals, win conditions, fail conditions, scoring, task points, routes, or placement.
3. Do not output coordinates, point locations, fixed paths, waypoints, or Blender code.
Do not interpret visual gaps as routes, corridors, lanes, or intended movement directions.
4. If there is an open gap between objects, describe it neutrally as visible open space, lower vegetation density, reduced occlusion, or clearer ground visibility.
5. movement_space_observation may only describe ground visibility, occlusion level, vegetation density, openness, and visible terrain continuity.
6. movement_space_observation must not describe where a player should go, a default direction, route, path, lane, or progression plan.
7. notes_for_rule_designer must not recommend turning a gap into a route, aligning camera behavior, or reinforcing one-way movement.

# Output Rules
Return only strict JSON. Do not output markdown.

# Required JSON Schema
```json
{
  "visual_rule_notes_status": {
    "usable": true,
    "confidence": 0.85,
    "reason": ""
  },
  "visual_rule_notes": {
    "rule_relevant_scene_summary": "",
    "scene_features_useful_for_rules": [],
    "movement_space_observation": "",
    "visual_landmarks_for_rules": [],
    "rule_design_risks_from_visuals": [],
    "notes_for_rule_designer": []
  }
}
```

# Text Inputs
```json
<TEXT_INPUTS_JSON>
```
\end{lstlisting}

\prompttitle{Image-Input Prompt}
\begin{lstlisting}[style=promptstyle]
Image: default_camera_view_grid.png
\end{lstlisting}

\subsection*{Declarative Gameplay Planning}

\prompttitle{System Prompt}
\begin{lstlisting}[style=promptstyle]
You are a strict JSON game_rule_agent. Design playable game rules from text inputs only. Do not output coordinates, paths, waypoints, keyframes, Blender code, or asset prompts.
\end{lstlisting}

\prompttitle{User Prompt}
\begin{lstlisting}[style=promptstyle]
# Role
1. You are game_rule_agent for Code2Games.
2. You are not scene_understanding, not placement_agent, not a path planner, and not a script writer.
3. You are a text/coding LLM call and you do not receive images. Use only the text inputs below.

# Task
1. Design a playable game rule plan that fits the current DEFAULT_CAMERA scene.
2. You design how the game works. You do not choose exact placement points and you do not write Blender code.
3. The user prompt is a goal direction, not something to copy blindly. Combine it with scene understanding and visual notes.
4. The phrase temple-run-like is a style reference only. It does not mean the game must use endless-runner mechanics.
5. Do not design forced automatic forward motion unless the user explicitly requests that exact mechanic.
6. Do not make the core loop depend on forced one-way movement.
7. The game should be a task-based free-navigation challenge.

# Design Scope
1. Include game concept, core loop, player/NPC role, objective, win condition, fail condition, time limit, scoring rules, allowed actions, scene element usage, randomness/free-navigation rules, and gameplay element categories needed by a later placement_agent.
2. Do not design the game as only 'NPC runs forward'. The player or NPC should support free movement and random choices.
3. allowed_actions must include move_forward, move_backward, move_left, move_right. Add jump, dodge, collect, interact, dive, or swim_or_wade when appropriate.
4. Recommended rule direction: free-navigation exploration, required item collection, interaction with a goal or event category, obstacle management, and completion within time_limit_seconds.
5. The existing scene is the immutable visual world base. Existing trees, bushes, rocks, water, and terrain may inspire the rules and provide navigation context, but they are not automatically gameplay obstacles, collectibles, goals, landmarks, or newly placed assets.
6. Gameplay elements listed for later placement describe new game content to add on top of that base. For example, observing an existing bush may justify vegetation-compatible gameplay, but must not define that bush itself as a placed obstacle or collision trigger.

# Time And Failure
1. duration_seconds is a reference budget. The rule time_limit_seconds may be larger, for example 45 or 60 when duration_seconds is 30.
2. The default fail condition should be time-based: if the win_condition is not achieved within time_limit_seconds, the player fails.
3. win_condition must require completing a task objective. It must not only be surviving until the timer ends.
4. Keep fail_condition simple. Light extra fail conditions are allowed, such as a collision penalty count exceeding a limit or health reaching zero.
5. Do not make the first collision an immediate failure unless the user explicitly asks for harsh one-hit failure rules.

# Forbidden Output
1. Do not output concrete coordinates.
2. Do not output exact spawn points, item points, event points, route points, fixed paths, waypoints, trajectories, or keyframes.
3. Do not output Blender code, Python code, bpy calls, or asset-generation prompts.
4. If later placement needs locations, placement_agent will choose them from this game_rule_plan.

# Placement Categories Only
1. required_gameplay_elements_for_placement must contain categories and purposes only.
2. Allowed element_type values: collectible, obstacle, event_trigger, goal_area, hazard_zone, safe_zone, landmark.
3. required_gameplay_elements_for_placement should describe only the element category, gameplay purpose, coarse scene-based placement intent, and required_count.
4. placement_hint_from_scene may request a new element near useful environment context for composition, but must never request reusing, converting, or attaching gameplay behavior to a specific existing scene object.
5. For obstacle requirements, describe the intended player response such as jump, dodge, or route-around. Leave the concrete new obstacle asset type to the later asset planner and do not default all obstacles to existing vegetation.
6. Do not output coordinates, exact locations, paths, or world coordinate fields. The later placement_agent will decide exact world positions.

# Output Rules
Return only strict JSON. Do not output markdown.

# Required JSON Schema
```json
{
  "game_rule_status": {
    "usable": true,
    "confidence": 0.85,
    "reason": ""
  },
  "game_rule_plan": {
    "game_concept": {
      "title": "",
      "description": "",
      "why_it_fits_this_scene": ""
    },
    "core_loop": "",
    "player_or_npc_role": "",
    "objective": "",
    "win_condition": "",
    "fail_condition": "",
    "time_limit_seconds": 60,
    "scoring_rules": [],
    "allowed_actions": [],
    "scene_element_usage": [],
    "randomness_policy": {
      "free_navigation": true,
      "fixed_path_required": false,
      "random_choice_allowed": true,
      "description": ""
    },
    "required_gameplay_elements_for_placement": [
      {
        "element_type": "",
        "purpose": "",
        "placement_hint_from_scene": "",
        "required_count": 1
      }
    ]
  }
}
```

# Text Inputs
```json
<TEXT_INPUTS_JSON>
```
\end{lstlisting}

\subsection*{Top-Down Visual Region and Route Planning}

\prompttitle{System Prompt}
\begin{lstlisting}[style=promptstyle]
Return only valid JSON. Do not output markdown. Do not output text outside the JSON object. Do not output Blender world coordinates. The path_cells field is a strict 16x16 grid route and must be valid before you answer. Every consecutive pair must be adjacent by edge only. Mathematical rule: abs(row difference) + abs(column difference) must equal 1 for every step. Diagonal corner-touching moves are invalid because abs(row difference) == 1 and abs(column difference) == 1 gives a sum of 2. Physical rule: the runner cannot cut through a grid-cell corner or jump diagonally; it must move through a horizontal or vertical neighbor first. Never jump over cells, skip intermediate cells, or teleport between regions. Never repeat any grid cell anywhere in path_cells. Never backtrack, revisit a previous cell, create a loop, or output a route like H7 -> H6 -> H5 -> H6 -> H7. If a scenic route would require moving from D4 to D8, output D4, D5, D6, D7, D8. If a scenic route would require moving from D4 to E5, output D4, D5, E5 or D4, E4, E5. If a scenic route would require moving diagonally across multiple rows/columns, output all intermediate edge-neighbor cells. If the path is not continuous and non-repeating, your answer is invalid.
\end{lstlisting}

\prompttitle{User Prompt}
\begin{lstlisting}[style=promptstyle]
# Role

You are the visual route planner for Code2Games / Code2Worlds.

# Project Goal

We already have a Blender scene generated by Code2Worlds/Infinigen.
We need to choose one path.
Later scripts will add game-like GLB assets along this path and render a high-quality, cinematic, game-like gameplay video.

# Inputs

You will see three grid-aligned images:

1. 01_topdown_rgb.png
   A real topdown RGB scene image with grid labels.

2. 02_topdown_object_density.png
   An object, vegetation, rock, asset, and scene-content density map.
   High density is not bad by default; it often means the image is visually richer.

3. 03_topdown_geometry.png
   Terrain height, slope, roughness, and continuity evidence.

You also receive layout_hierarchy.json as compact cell-level evidence.

# Step 1: Analyze the Images

First analyze the three images in detail in the same VLM call, and output analysis fields in JSON:
- rgb_scene_analysis
- object_density_analysis
- geometry_analysis
- cross_image_summary
- path_planning_implications

For 01_topdown_rgb.png, analyze:
- visible scene structure,
- open areas,
- dense areas,
- apparent corridors,
- terrain regions,
- visually meaningful spatial patterns,
- where the camera route would look beautiful.
Use grid cell ids from A1 to P16.

For 02_topdown_object_density.png, analyze:
- object, vegetation, rock, asset, and scene-content density patterns,
- low-density regions,
- high-density visually rich regions,
- scenic boundaries,
- sparse gaps inside or near dense regions,
- transitions between empty and dense regions,
- which dense regions are visually valuable for the final video.
Do not treat dense regions mainly as danger or obstacles.
Use grid cell ids from A1 to P16.

For 03_topdown_geometry.png, analyze:
- terrain height patterns,
- slope patterns,
- roughness,
- continuity,
- stable terrain regions,
- visually interesting terrain variation,
- regions that give camera depth or route progression.
Use grid cell ids from A1 to P16.

Then compare the three images:
- where RGB appearance, density evidence, and geometry evidence agree,
- where they disagree,
- which regions are most valuable for a scenic route,
- which regions are empty and visually boring,
- which regions help camera-visible scenery.

Then explain path planning implications:
- which regions should the path try to pass through or near,
- which regions should be used as scenic corridors,
- which regions provide turns, depth, and visual progression,
- which regions are good for later GLB asset staging.

Do not output world_xyz.

# Step 2: Choose the Path

Based on the image analysis above, directly choose the final path.

Core objective:
- More scenery is better.
- Prefer passing through or near visually rich regions.
- High object density does not mean avoid; it often means more trees, rocks, vegetation, assets, and terrain detail.
- Do not choose an empty boring outer route.
- Do not make safe obstacle avoidance the main objective.
- This is not true playable-game collision planning.
- This is for Blender gameplay video staging, so choose the path with the strongest visual feeling.
- The path only needs to be continuous, visually plausible, and basically on terrain.
- Prefer a path with turns, spatial variation, camera depth, and scenic progression.

Path construction rules:
- Build path_cells as a single forward-progressing route.
- Every next cell must be an edge-neighbor of the previous cell.
- Mathematical rule: for every consecutive pair, abs(row difference) + abs(column difference) must equal 1.
- Diagonal moves are invalid: abs(row difference) == 1 and abs(column difference) == 1 is a corner-cutting jump, not adjacency for this task.
- Physical interpretation: the runner cannot cut through a grid-cell corner; a diagonal-looking move must be decomposed into one horizontal step and one vertical step.
- The next cell cannot be the same as the previous cell.
- Do not jump over cells.
- Do not skip intermediate cells.
- Do not teleport between scenic regions.
- Do not revisit any previous cell.
- Do not repeat a cell anywhere in path_cells.
- Do not create loops.
- Do not create a U-turn route.
- Do not go forward and then backtrack through the same cells.
- Do not use a route like H7 -> H6 -> H5 -> H6 -> H7.
- If you enter a scenic basin or turn region, exit through a new adjacent cell instead of going back through already used cells.
- The path should have a clear start, middle, and end.
- The route should feel like one continuous gameplay run, not a patrol path.
- Before finalizing, internally check every pair:
  1. Every consecutive pair is adjacent.
  2. No cell appears more than once.
  3. The path does not backtrack through previous cells.
  4. The path has a clear forward progression.
- If you want to move from D4 to D8, you must output D4, D5, D6, D7, D8.
- If you want to move from D4 to E5, you must output either D4, D5, E5 or D4, E4, E5.
- If you want to move from D4 to H8, you must output a sequence of edge-neighbor cells; do not use diagonal shortcuts.
- Never output a path with disconnected segments.
- Scenic richness cannot override continuity and no-repeat rules.

# Output

Return only strict JSON:
{
  "rgb_scene_analysis": "...",
  "object_density_analysis": "...",
  "geometry_analysis": "...",
  "cross_image_summary": "...",
  "path_planning_implications": "...",
  "visual_route_strategy": "...",
  "path_cells": ["A1", "A2"],
  "confidence": 0.0,
  "reason": "Explain why this path has more scenery and is suitable for the final high-quality gameplay video.",
  "key_visual_regions": [
    {
      "grid_cell": "H5",
      "reason": "Why this region is visually important."
    }
  ]
}

Rules:
- Return only JSON.
- Do not output markdown.
- Do not output text outside the JSON object.
- path_cells may only use A1 to P16.
- path_cells must be continuous using edge-neighbor moves only.
- Consecutive cells must satisfy abs(row difference) + abs(column difference) == 1.
- Diagonal steps are not allowed.
- Do not explain the continuity check outside JSON.
- The final path_cells must already be corrected.
- Do not output Blender world coordinates.
- Do not output world_xyz.

Cell-level evidence JSON:
```json
<LAYOUT_HIERARCHY_JSON>
```
\end{lstlisting}

\prompttitle{Image-Input Prompts}
\begin{lstlisting}[style=promptstyle]
Image 1 is 01_topdown_rgb.png: a real topdown view of the scene with grid cell labels.
Image 2 is 02_topdown_object_density.png: a grid-aligned evidence map showing object, vegetation, rock, asset, and scene-content density. High density often means visually richer scenery, not something to avoid by default.
Image 3 is 03_topdown_geometry.png: a grid-aligned evidence map showing terrain height, slope, roughness, and continuity cues.
\end{lstlisting}

\subsection*{Top-Down Route Repair}

\prompttitle{System Prompt}
\begin{lstlisting}[style=promptstyle]
Return only valid JSON. Do not output markdown. Do not output text outside the JSON object. Do not output Blender world coordinates. The path_cells field is a strict 16x16 grid route and must be valid before you answer. Every consecutive pair must be adjacent by edge only. Mathematical rule: abs(row difference) + abs(column difference) must equal 1 for every step. Diagonal corner-touching moves are invalid because abs(row difference) == 1 and abs(column difference) == 1 gives a sum of 2. Physical rule: the runner cannot cut through a grid-cell corner or jump diagonally; it must move through a horizontal or vertical neighbor first. Never jump over cells, skip intermediate cells, or teleport between regions. Never repeat any grid cell anywhere in path_cells. Never backtrack, revisit a previous cell, create a loop, or output a route like H7 -> H6 -> H5 -> H6 -> H7. If a scenic route would require moving from D4 to D8, output D4, D5, D6, D7, D8. If a scenic route would require moving from D4 to E5, output D4, D5, E5 or D4, E4, E5. If a scenic route would require moving diagonally across multiple rows/columns, output all intermediate edge-neighbor cells. If the path is not continuous and non-repeating, your answer is invalid.
\end{lstlisting}

\prompttitle{User Prompt}
\begin{lstlisting}[style=promptstyle]
You previously returned an invalid path for a 16x16 grid.

Validation errors:
<VALIDATION_ERRORS_JSON>

Previous JSON:
```json
<PREVIOUS_JSON>
```

The path must be repaired. Please repair path_cells only, and update confidence/reason if needed.

Rules:
- Return only valid JSON.
- Do not output markdown.
- Use the same JSON schema.
- Keep the scenic route intention.
- Keep the route visually rich.
- But the path_cells must be continuous.
- Do not repeat any cell.
- Do not revisit any previous cell.
- Do not backtrack.
- Do not create loops.
- Do not use a path like H7 -> H6 -> H5 -> H6 -> H7.
- If the previous route entered H5 from H6, it must leave through a different unused adjacent cell, or choose another scenic branch.
- Every consecutive pair must be adjacent by edge only.
- Mathematical rule: abs(row difference) + abs(column difference) must equal 1.
- Diagonal moves are invalid even if the cells touch at a corner.
- Physical interpretation: the runner cannot jump through a grid corner; replace every diagonal shortcut with two edge-neighbor steps.
- The same cell cannot appear twice in a row.
- Do not jump from one region to another.
- Insert only explicit edge-neighbor intermediate cells where needed.
- All cells must be within A1 to P16.
- Do not output Blender world coordinates.
- Do not output world_xyz.

Required JSON schema:
{
  "rgb_scene_analysis": "...",
  "object_density_analysis": "...",
  "geometry_analysis": "...",
  "cross_image_summary": "...",
  "path_planning_implications": "...",
  "visual_route_strategy": "...",
  "path_cells": ["A1", "A2"],
  "confidence": 0.0,
  "reason": "...",
  "key_visual_regions": [
    {
      "grid_cell": "H5",
      "reason": "..."
    }
  ]
}
\end{lstlisting}

\prompttitle{Image-Input Prompts}
\begin{lstlisting}[style=promptstyle]
Image 1 is 01_topdown_rgb.png: a real topdown view of the scene with grid cell labels.
Image 2 is 02_topdown_object_density.png: a grid-aligned evidence map showing object, vegetation, rock, asset, and scene-content density. High density often means visually richer scenery, not something to avoid by default.
Image 3 is 03_topdown_geometry.png: a grid-aligned evidence map showing terrain height, slope, roughness, and continuity cues.
\end{lstlisting}

\subsection*{Scene-Constrained Candidate Selection}

\prompttitle{System Prompt}
\begin{lstlisting}[style=promptstyle]
Return only valid JSON. Never output coordinates, code, paths, or waypoints.
\end{lstlisting}

\prompttitle{User Prompt}
\begin{lstlisting}[style=promptstyle]
# Code2Games single-pass placement
Choose one existing candidate_id for every required gameplay element in game_rule_plan.required_gameplay_elements_for_placement.
Use only candidate ids and allowed_placement_modes from the input. Never invent coordinates, paths, code, or candidate ids.
The existing scene is the visual base, not the gameplay asset set. Trees, bushes, rocks, and terrain remain background environment. Every selected candidate is the actual location for a newly added gameplay element or logical zone.
nearest_tree and nearest_bush are composition context: prefer suitable ground near environment objects so the final shot has depth and visual integration, but leave enough clearance for a new standalone object. Near does not mean on, inside, or reused.
Prefer clearings inside or between the existing scenic vegetation, where trees, bushes, rocks, or other scene features remain nearby around the gameplay space. Do not place most gameplay elements on empty ground outside the scenic cluster merely because it has maximum clearance. The intended result is gameplay embedded in the scenery, not a separate prop field beside it.
Balance integration and fit: choose a genuinely empty patch within the scenic area whose local clearance can contain the later asset. A point touching vegetation is not a clearing, but a moderately open interior gap is usually compositionally better than a very distant exterior point. Use the complete scene understanding and distribute elements through multiple natural interior clearings rather than forming one isolated open-field group.
Visible assets must not form accidental prop piles. Spread them across different rays, screen grid cells, and depth buckets wherever the game rules permit. The deterministic staging step enforces both a minimum center distance and a positive footprint-to-footprint gap and rejects the plan instead of silently moving assets.
Interpret nearest_tree.distance_xy and nearest_bush.distance_xy as horizontal clearance in scene meters from the candidate to the vegetation object's XY bounding box: 0 means the point overlaps that projected bounding box, below 1 meter is crowded, 1 to below 3 meters has limited clearance, 3 to below 5 meters is open, and 5 meters or more is very open. Never call a 0.16-meter clearance open.
The DEFAULT_CAMERA context is only an initial scene reference, not the final gameplay camera. Do not optimize placement only for that view; the later director may create a first-person, third-person, vehicle, cinematic, or spectator camera.
Candidate facts override narrative convenience. Every placement_reason must agree with that candidate's candidate_kind, surface_type, walkability, vegetation distances, screen grid cell, and hit_object. Never describe a vegetation-crowded candidate as open or clear.
Supported element types: camera_anchor: first-person, third-person, vehicle, cinematic, or spectator camera anchor; capture_zone: logical or visibly marked defend, contest, or capture area; checkpoint: logical or visibly marked progression/race checkpoint; collectible: visible item collected for score, progression, or objectives; cover_point: tactical position for player or AI cover behavior; destructible: visible breakable object with a later damage response; enemy_spawn: hostile actor spawn or reinforcement anchor; event_trigger: logical trigger that starts a scripted gameplay event; goal_area: visible destination marker plus completion trigger; hazard_zone: visible dangerous surface or volume plus hazard trigger; interaction_point: logical or visibly represented use, dialogue, switch, door, or terminal point; item_pickup: weapon, ammunition, health, power-up, key, or inventory pickup; landmark: visible orientation or spectacle asset; moving_platform: visible platform or carrier whose motion is added by the director/runtime; navigation_anchor: unordered AI or gameplay navigation anchor for later path planning; npc_spawn: friendly, neutral, civilian, companion, or quest NPC anchor; objective_object: visible mission-critical object to protect, carry, activate, or destroy; obstacle: visible object that blocks, redirects, or tests movement; player_spawn: initial player or respawn anchor; puzzle_element: visible switch, mechanism, movable piece, or puzzle target; safe_zone: logical or visibly marked area where danger is reduced; traversal_point: climb, vault, jump, grapple, launch, landing, or transition anchor; vehicle_spawn: player, opponent, traffic, or ambient vehicle anchor.
Collectibles should usually use on_ground. Only a small minority may use above_ground when they intentionally guide a reachable jump or form a clear gameplay cue. Do not make every collectible airborne and do not mechanically reuse one height for all airborne collectibles.
Player, enemy, NPC, and vehicle spawns are logical anchors. Put ground actors on walkable, non-overlapping ground with clearance suitable for their role; the later director supplies rigs, AI, vehicles, and animation.
Cover points need tactically useful nearby occlusion and reachable ground. Camera anchors and traversal points may use above_ground only when the requested height has a clear purpose.
Checkpoints, capture zones, event triggers, objectives, and navigation anchors must form a coherent layout, but do not invent an ordered path or reuse one candidate for multiple elements.
Item pickups must be reachable and readable. Destructibles, moving platforms, puzzle elements, and objective objects need enough physical clearance for their later visible asset.
A safe_zone must use a ground_surface candidate with good or medium walkability, with both nearest_tree.distance_xy and nearest_bush.distance_xy at least 3 meters. Do not use vegetation-crowded points for safe zones.
Obstacles are newly added game obstacles, not existing vegetation. Select walkable ground near useful visual context, never an existing object surface. Their later asset should clearly communicate jump, dodge, or route-around gameplay.
Landmarks and goal markers are also newly added visible assets. Select ground beside useful scene context rather than using an existing tree, bush, or rock as the landmark or goal.
Choose the height for the specific object, not from a fixed element-type rule. Use placement_mode=above_ground and a positive numeric height_offset only when the gameplay object should be airborne and reachable. For every other mode use height_offset=0.
Each placement may optionally specify surface_alignment_mode=auto|upright|terrain_normal|full_surface_normal plus max_tilt_degrees, surface_clearance_m, vegetation_clearance_radius_m, and yaw_degrees. Prefer auto: Python deterministically uses terrain normals on ground/slopes and full sampled normals on cliff_or_wall candidates. Use upright only for objects whose gameplay meaning requires vertical orientation.
Asset dimensions, fog or water volume height, interaction size, collision size, and exact mesh intersection are decided later after the real asset exists.
Return strict JSON only.
Schema: {selection_status:{usable:true,confidence:0.0,reason:'non-empty'},placements:[{element_type,source_requirement_index,selected_candidate_id,placement_mode,height_offset,gameplay_purpose,placement_reason,surface_alignment_mode?,max_tilt_degrees?,surface_clearance_m?,vegetation_clearance_radius_m?,yaw_degrees?}],selection_notes:['layout summary']}
<INPUT_JSON>
\end{lstlisting}

\subsection*{Physically Constrained Candidate Selection}

\prompttitle{System Prompt}
\begin{lstlisting}[style=promptstyle]
Return only valid JSON. Never output coordinates, code, paths, or waypoints.
\end{lstlisting}

\prompttitle{User Prompt}
\begin{lstlisting}[style=promptstyle]
# Code2Games placement selector
You receive every candidate as objective facts. Python has not decided which candidate suits each gameplay purpose.
Supported element types: camera_anchor: first-person, third-person, vehicle, cinematic, or spectator camera anchor; capture_zone: logical or visibly marked defend, contest, or capture area; checkpoint: logical or visibly marked progression/race checkpoint; collectible: visible item collected for score, progression, or objectives; cover_point: tactical position for player or AI cover behavior; destructible: visible breakable object with a later damage response; enemy_spawn: hostile actor spawn or reinforcement anchor; event_trigger: logical trigger that starts a scripted gameplay event; goal_area: visible destination marker plus completion trigger; hazard_zone: visible dangerous surface or volume plus hazard trigger; interaction_point: logical or visibly represented use, dialogue, switch, door, or terminal point; item_pickup: weapon, ammunition, health, power-up, key, or inventory pickup; landmark: visible orientation or spectacle asset; moving_platform: visible platform or carrier whose motion is added by the director/runtime; navigation_anchor: unordered AI or gameplay navigation anchor for later path planning; npc_spawn: friendly, neutral, civilian, companion, or quest NPC anchor; objective_object: visible mission-critical object to protect, carry, activate, or destroy; obstacle: visible object that blocks, redirects, or tests movement; player_spawn: initial player or respawn anchor; puzzle_element: visible switch, mechanism, movable piece, or puzzle target; safe_zone: logical or visibly marked area where danger is reduced; traversal_point: climb, vault, jump, grapple, launch, landing, or transition anchor; vehicle_spawn: player, opponent, traffic, or ambient vehicle anchor.
Judge candidate purpose from the game concept, core loop, objective, scene understanding, visual notes, geometry labels, visibility, distance, vegetation context, terrain_region, scene_sector, elevation_band, support_class, grid cells, and the complete spatial layout.
Physical support is mandatory. Grounded visible assets must use candidates whose max_supported_footprint_radius_m and support_class can carry the intended object. Prefer level platforms and clearings; never place a rigid prop on an irregular hillside merely because its center ray is walkable. Large goals, landmarks, barriers, vehicles, and platforms need large or extra_large support whenever available.
Actor and vehicle spawns are logical anchors on reachable clear ground. Cover points need tactically useful nearby occlusion. Camera anchors and traversal points may be above ground only for an explicit gameplay purpose. Visible objects need enough clearance for their later asset.
Do not choose points independently. Spread visible assets across different scene_sector, terrain_region, elevation_band, rays, screen grid cells, and depth buckets wherever the game rules permit; do not form accidental prop piles. Use the whole playable scene rather than only one barren patch. Deterministic staging validates both minimum center distance, footprint-to-footprint clearance, and realized-asset platform support, and rejects violations instead of silently moving assets.
Vegetation is gameplay space, not merely background. For TPS prefer multiple vegetated clearings while preserving some open or mountain beats; for FPS mix vegetation, open ground, and only truly flat mountain shelves; for racing place route cues on drivable clear ground or supported cliff-edge shelves and avoid under-tree obstructions; for wingsuit place summit launch props on supported shelves and distribute later markers along the descent/landing progression.
Every selection must satisfy hard_constraints and use one of that candidate's physically_supported_modes. If the complete requirements cannot satisfy those constraints, return usable=false, an empty placement_selections list, and structured requested_rule_relaxations. Python will not accept relaxations automatically.
No element type has a predefined height. Choose placement_mode for the specific gameplay object or zone. Only above_ground requires a positive numeric height_offset; for every other placement_mode use 0. Explain the vertical choice in placement_reason.
height_offset means displacement from the placement anchor; it is not the visual asset's own height or a fog/water volume extent. Actual asset dimensions, interaction radius, collision size, and exact scene-mesh intersection are determined later after asset realization.
Each selection may optionally author reproducible surface staging intent: surface_alignment_mode is auto, upright, terrain_normal, or full_surface_normal; max_tilt_degrees is 0..180; surface_clearance_m is 0..10; vegetation_clearance_radius_m is 0..100; yaw_degrees is -180..180. Omit tuning values unless they express deliberate art direction; tuning cannot make an unsupported hillside physically valid.
Return strict JSON only. Do not output coordinates, locations, ordered paths, trajectories, code, new candidate ids, interaction radii, collision radii, or visibility priorities.
Schema: {selection_status:{usable:true,confidence:0.0,reason:'non-empty'},requested_rule_relaxations:[],placement_selections:[{placement_id,element_type,source_requirement_index,selected_candidate_id,placement_mode,height_offset,gameplay_purpose,placement_reason,surface_alignment_mode?,max_tilt_degrees?,surface_clearance_m?,vegetation_clearance_radius_m?,yaw_degrees?}],selection_notes:['final layout summary']}
<INPUT_JSON>
\end{lstlisting}

\subsection*{Asset Realization}

\prompttitle{System Prompt} 
\begin{lstlisting}[style=promptstyle]
Return one strict JSON asset plan. Fixed placements must not be changed.
\end{lstlisting}

\prompttitle{User Prompt}
\begin{lstlisting}[style=promptstyle]
# Code2Games direct asset generation plan
All placement locations are final. Do not choose, move, add, remove, or reorder placements.
For each fixed placement, decide whether it needs a visible GLB. Put placements that do not need GLB, such as purely logical zones, in non_glb_placements.
These types require a visible GLB: collectible, destructible, goal_area, hazard_zone, item_pickup, landmark, moving_platform, objective_object, obstacle, puzzle_element.
These types are logical anchors and MUST be non_glb_placements: camera_anchor, enemy_spawn, navigation_anchor, npc_spawn, player_spawn, vehicle_spawn. The later gameplay director supplies actors, vehicles, AI, rigs, animation, and cameras.
These types may be logical or visibly represented according to their purpose: capture_zone, checkpoint, cover_point, event_trigger, interaction_point, safe_zone, traversal_point.
Every obstacle must be a newly generated, visible, physically plausible gameplay object placed at its fixed location. Existing trees, bushes, rocks, terrain, or other scene geometry must never fulfill an obstacle placement.
An obstacle is a game obstacle with a clear player response such as jump, dodge, or route-around. The scene's existing vegetation and terrain are only the visual base. Do not generate ordinary replacement bushes or trees merely because the fixed point is near vegetation.
When several obstacles are requested, create meaningful visual and gameplay variety that remains plausible for each local ground context. Do not assign every obstacle placement to one identical asset; reuse only within a suitable subset.
Do not create freestanding pillars, columns, poles, narrow totems, detached gate pieces, or isolated wall fragments. These depend on a road, gate, wall, ruin complex, or other supporting structure and look arbitrary in a natural clearing. Every selected object must be visually and physically self-contained at its fixed point. Prefer independent natural gameplay forms such as fallen logs, tangled roots, thorny shrubs, spiked plants, bramble masses, grounded relic mounds, or rock formations when locally plausible.
Every goal area must have a newly generated visible marker in addition to its later logical trigger, and every landmark must have a newly generated visible orientation asset.
For placements that need visible objects, creatively choose a physically plausible asset from the actual local road and environment facts. Random variety is welcome only inside physical reality and gameplay intent.
An item_pickup is a visible weapon, ammunition, health, power-up, key, or inventory item; its exact role must follow gameplay_purpose. An objective_object, destructible, puzzle_element, or moving_platform must have a clear standalone silhouette and physical function.
A logical actor/vehicle spawn is not a request to generate a static person, enemy, or vehicle GLB. Record its later implementation in non_glb_placements.
A flat ground point cannot receive floating water, an unsupported hovering structure, or an object inconsistent with gravity. A slope-aware object must sit plausibly on that slope. An airborne asset must match the given above_ground mode and reachable height. Water-like assets require real water or suitable low ground evidence.
hit_object_context, nearest_tree, and nearest_bush only describe the scene base around the fixed ground point. They exist to improve composition and style matching. They are never the asset to reuse, never satisfy the placement, and must not appear as an existing-scene implementation. An upright tree is not a jump obstacle merely because it is nearby.
Every generated asset must match the current scene's shape language, natural materials, color palette, roughness, weathering, vegetation integration, realism, and visual age from scene_understanding and game_rule_visual_notes.
Do not merely repeat a broad biome or style label from scene text. Ground every material and palette decision in the actual visible evidence. Do not carry colors or art direction from a previous demo into the current scene.
Plan dimensions at gameplay scale, not tabletop-decoration scale. Judge every expected_size_meters against the player/vehicle scale and the requested first-person, third-person, vehicle, or cinematic camera: pickups must remain readable, obstacles and cover must communicate their function, hazards must occupy meaningful ground, and objectives/landmarks must read from the intended distance. Use local clearance facts as limits; do not blindly apply one multiplier to all assets.
expected_size_meters always uses Blender axis order [X width, Y depth, Z height] in meters. Never use [width,height,depth]. A flat ground patch must therefore have large X and Y and a small Z. State the intended physical orientation in generation_prompt_en so image-to-3D does not turn a ground patch into a vertical wall.
Every asset must declare flat_base_policy as allow or reject and explain it in flat_base_reason. Use allow only when a broad planar support is an intended physical part of the object, such as a crate/case, parked vehicle, grounded equipment shelter, platform, or deliberately flat floor marker. Use reject for organic rocks, roots, vegetation, irregular relics, pickups, and sculptural objects where a generated slab-like bottom would be an artifact. This policy controls validation; it does not ask the model to add a base.
Respect fixed surface_staging. A full_surface_normal placement is attached to a cliff or wall and needs a plausible contact side; terrain_normal follows ground or slope; upright must visually remain vertical. Do not redesign or move the placement.
Collectibles must have an unmistakable compact pickup silhouette, not resemble a chair, miniature shrine, building, cage, or pile of logs. Landmarks must be broad, asymmetric, self-supporting environmental masses; a renamed upright slab or stone marker still violates the no-pillar requirement.
You decide reuse: multiple placement_ids may share one asset only when the same self-contained object is physically suitable at every assigned placement. Split ground-supported and airborne variants when one model would look unsupported. Do not force either full reuse or one asset per placement.
Default to one unique generated asset per visible placement so the demo has real gameplay variety. Reuse is exceptional and must be justified by an intentionally identical repeated item. When a demo has ten or more visible or optionally visible placements, produce at least ten distinct generated assets.
Prefer unmistakably volumetric 3D objects with substantial height and depth, readable front/side/back surfaces, self-supporting construction, and a strong silhouette from a three-quarter view. Do not represent a hazard, goal, landmark, safe zone, or event cue as a thin ground patch, decal, painted mark, texture sheet, nearly flat tile, or shallow relief. The runtime owns the invisible trigger/zone; the generated GLB should be its three-dimensional visual marker.
generation_prompt_en must describe exactly one isolated standalone object for image-to-3D generation and must explicitly include the current scene style. Do not describe a full scene, terrain, character, camera, text, UI, or animation.
expected_glb_path must be a relative path under ./assets/generated_glb/ and end in .glb. This demo-specific prefix prevents assets from different games from overwriting each other.
Do not output world coordinates or Blender code.
Return strict JSON only.
Schema: {assets:[{asset_id,asset_name,asset_role,placement_ids,generation_prompt_en,negative_prompt_en,expected_glb_path,expected_size_meters:[x,y,z],flat_base_policy:'allow|reject',flat_base_reason:'non-empty'}],non_glb_placements:[{placement_id,implementation,reason}],notes:['summary']}
<INPUT_JSON>
\end{lstlisting}

\subsection*{Per-Asset Reference-Image Generation}

\prompttitle{User Prompt}
\begin{lstlisting}[style=promptstyle]
REFERENCE IMAGE COMPOSITION REQUIREMENTS: Create a single clean product reference for image-to-3D generation, not an environment concept image. Show exactly one standalone 3D game prop centered in the frame against a plain pure-white seamless studio background. Do not depict a forest, orchard, landscape, terrain, surrounding trees, surrounding plants, or a natural ground scene. Any environment or palette language in the object description is material guidance for the prop itself only. Use neutral shadowless studio lighting, full object visible, elevated three-quarter product view, with empty white margin around the complete silhouette. Show only the object's own natural lowest contact edge. Never add a support base, floor, board, platform, pedestal, plinth, tile, slab, disk, square, rectangle, or triangle beneath it. The background must remain uniformly white with no horizon, floor plane, cast shadow, contact shadow, reflection, or lighting gradient. For an airborne object, preserve clear empty background space underneath it and never invent feet or support geometry. The prop must have unmistakable three-dimensional volume: substantial height and depth, readable front, side, rear, top and underside surfaces, and a strong silhouette in the three-quarter view. Never turn a hazard, goal, landmark, safe-zone marker, or event cue into a flat ground patch, decal, painted mark, texture sheet, thin tile, mat, puddle, or shallow relief. The invisible gameplay zone is implemented separately at runtime. No character, no person, no text, no labels, no watermark, no UI, no scene background, no cropped parts.

OBJECT DESIGN: <OBJECT_DESIGN>

FINAL COMPOSITION CHECK: exactly one isolated prop, uniform pure-white background, no environment, no shadow, no floor or base, and no extra objects.
\end{lstlisting}
\vspace{-1em}
\prompttitle{Negative Prompt}
\begin{lstlisting}[style=promptstyle]
<ASSET_NEGATIVE_PROMPT>, people, character, human, hands, text, letters, watermark, logo, subtitle, UI, busy background, landscape scene, forest background, orchard background, trees in background, terrain, natural ground plane, floor, support board, base, pedestal, platform, plinth, tile, slab, disk, square, rectangle, triangle, cast shadow, contact shadow, reflection, surrounding vegetation, environmental scene, multiple objects, cropped object, blurry, low quality, distorted geometry
\end{lstlisting}

\end{document}